\documentclass{article}

\PassOptionsToPackage{numbers,compress}{natbib}

\usepackage[preprint]{neurips_2026}

\usepackage[utf8]{inputenc} 
\usepackage[T1]{fontenc}    
\usepackage{hyperref}       
\usepackage{url}            
\usepackage{booktabs}       
\usepackage{amsfonts}       
\usepackage{nicefrac}       
\usepackage{microtype}      
\usepackage[table]{xcolor}  

\usepackage{float}
\usepackage{graphicx}
\usepackage{amsmath}
\usepackage{array}
\usepackage{tikz}
\usetikzlibrary{arrows.meta,decorations.pathreplacing,calc,shapes.geometric,positioning}

\title{RIDE: An Open Dataset and Benchmark for Train Delay Prediction}

\author{%
  Clément Elliker$^{1}$ \quad
  Mathis Le Bail$^{1}$ \quad
  Clément Mantoux$^{2}$ \quad
  Jesse Read$^{1}$ \quad
  Sonia Vanier$^{1}$\\
  $^{1}$LIX, École Polytechnique, IP Paris, France \quad
  $^{2}$e.SNCF Solutions, France\\
  \texttt{clement.elliker@polytechnique.edu}
}

\begin{document}

\maketitle

\begin{abstract}
  Train delay prediction is an important problem for both passengers and railway operators, yet progress in the field remains difficult to assess due to the lack of standardized datasets, prediction targets, and evaluation protocols. To address this gap, we introduce RIDE, an open dataset and benchmark for train delay prediction built at nationwide scale over the Belgian railway network. RIDE covers 94.5M train events, 3.6M journeys, and 35.7M weather records from 2023 to 2025. It is organized as a layered data pipeline from raw railway and weather sources to two public releases: a reusable intermediate relational dataset and model-ready benchmark datasets. The benchmark standardizes the prediction task and the training and testing data. It also provides a unified evaluation protocol that supports direct comparison across models. Using this framework, we provide the first comprehensive comparative evaluation of non-learning, statistical learning, and deep learning models. We show that learning-based methods clearly outperform non-learning models, with graph neural networks achieving the best mean performance, while the strongest learning-based models remain relatively close to one another. Beyond aggregate mean absolute error (MAE) and root mean squared error (RMSE), the framework also provides breakdowns by prediction horizon and delay change, enabling more detailed analysis of model behavior across forecasting regimes.
\end{abstract}


\section{Introduction}

Railway networks are a central component of large-scale and sustainable mobility, supporting vast numbers of passenger journeys every day. Because rail operations are tightly scheduled and strongly interconnected, local disruptions can propagate across trains through the network, degrading reliability for both passengers and operators. Accurate delay prediction is therefore an important task: it can help railway operators anticipate how local disruptions propagate between trains, thereby improving passenger information, dispatching, and traffic management. These challenges are further amplified by the scale and complexity of modern railway systems, where operational, infrastructural, and exogenous factors such as weather jointly shape delay dynamics \cite{tiong2023review, spanninger2022review, spanninger2020approaches}.

Train delay prediction has been studied through a broad range of modeling approaches, from non-learning, rule-based methods to statistical approaches and more recently deep learning models \cite{tiong2023review, spanninger2022review, spanninger2020approaches, tiong2023quantitative}. Yet the field remains fragmented: delay prediction studies typically rely on different datasets, formulate different prediction targets, and evaluate under different protocols. Public datasets for train delay prediction have only recently begun to emerge \cite{zhang2022high, wu2025railway, dollmann2025bahn, borin2026integrating}, and they differ substantially in scope, coverage, and intended prediction task, without establishing a unified benchmark for downstream comparison. This makes it difficult to draw broad conclusions across studies, and motivates the need for a shared dataset and evaluation framework for train delay prediction.

To address these issues, we introduce RIDE (T\textbf{R}a\textbf{I}n \textbf{DE}lay Prediction Dataset and Benchmark), a comprehensive open resource for standardized train delay prediction nationwide. Unlike prior work, RIDE combines a reusable data release with a benchmark framework that standardizes prediction targets, temporal splits, and evaluation metrics. It covers three years of passenger railway operations over the Belgian network, spans diverse service types, and enables models to be compared on shared prediction instances under a common evaluation protocol. The main contributions are the following:
\begin{itemize}

\item[(i)] We release a layered data pipeline that transforms raw railway and weather sources into an intermediate relational data layer and then into model-ready benchmark datasets tailored to different model families; the intermediate layer enables users to construct alternative model-specific datasets without reproducing the core cleaning and integration steps.

\item[(ii)] We define a unified benchmark protocol with a common prediction target, common splits, and shared evaluation metrics, allowing diverse model families to be compared on identical prediction instances and metrics.

\item[(iii)] We provide the first comprehensive benchmark on a diverse set of approaches: non-learning methods, with a Translation baseline and Graph-event model; statistical learning methods, with XGBoost; and deep learning models, with MLP, LSTM, Transformer, and GNN. These experiments illustrate a clear advantage for learning-based methods, with graph neural networks achieving the best mean performance, while the strongest learning-based architectures remain close in absolute performance.

\item[(iv)] We analyze performance across prediction horizons and delay-change regimes, revealing regime-specific model strengths, such as stronger short-horizon performance for sequential models and stronger performance under large delay accumulation for models that explicitly capture interactions between trains.
\end{itemize}


\section{Related work}

\paragraph{Datasets.} Progress in train delay prediction is shaped not only by modeling choices, but also by the availability of shared datasets and standardized evaluation protocols. Public datasets for train delay prediction have only recently begun to emerge. Existing open resources include a dataset of high-speed rail operations on the Chinese railway network \cite{zhang2022high}, a multi-train-type operations dataset for Italy \cite{wu2025railway}, and more recent open archives for Germany \cite{dollmann2025bahn} and Finland \cite{borin2026integrating} that cover train operations over longer time spans, all of which include weather and scheduling information. While these resources are valuable, they remain limited in different ways, including temporal coverage, train diversity, or network scope, and do not by themselves establish a common benchmark framework combining event-level targets, fixed temporal splits, and a unified evaluation protocol for model comparison. As a result, papers in the literature rely on diverse public and proprietary datasets and use their own evaluation protocols, making it difficult to compare results across studies even when they address closely related problems.

\paragraph{Tasks.} Prediction targets are another area where standardization remains limited. Prior work considers a variety of formulations, including next-station delay prediction, multiple-stations-ahead delay prediction, delay after a fixed time interval, station-level aggregate quantities, and event-level outputs \cite{tiong2023review, spanninger2022review, spanninger2020approaches}. While these formulations can be well suited to specific operational settings, they further complicate comparison across studies. The common factor underlying these approaches is that train delays are measured at discrete scheduled events. In particular, scheduled arrivals, departures and passages of a train at an operational point are the finest-grained unit at which predictions can be validated against ground truth observations, and from which coarser quantities can be derived by aggregation. For this reason, we choose the delay at the next $n$ scheduled events of each train as the common prediction target for this benchmark.

\paragraph{Metrics.} Evaluation methods in the literature are similarly heterogeneous, although MAE and RMSE remain the dominant metrics across studies \cite{tiong2023review, spanninger2022review, yong2025ap}. Percentage-based measures such as mean absolute percentage error (MAPE) are also used, but are problematic in train delay prediction because delays are often close to zero \cite{yong2025ap}. Beyond aggregate accuracy, some works further stratify performance by prediction horizon or by delay ranges in order to study different operating regimes \cite{spanninger2022review, nair2019ensemble}. Our evaluation protocol follows this general direction, while extending it with a unified set of breakdowns by prediction horizon and delay change, designed to make regime-specific strengths and weaknesses easier to compare across models.

\paragraph{Network Scope.} Another important point of variation in the literature is the scope of prediction. Many studies focus on isolated train lines, single-track systems, or otherwise restricted operational settings \cite{liu2023prediction, meesit2025real}, which can be well suited to use cases where services can reasonably be treated in isolation. Other settings instead require modeling interactions at broader network scale \cite{chowdhury2025rstgcn, elliker2026simulation}. In this work, we adopt a full-network benchmark so that potential interactions and delay propagation across services remain part of the prediction problem. At the same time, the dataset remains flexible enough to support narrower benchmarks derived from specific lines or sub-networks, allowing future work to follow the same general framework under different assumptions about operational scope.

\paragraph{Input Features.} Across most modeling approaches, prior work relies on railway operational information, timetable and infrastructure context, and external variables such as weather \cite{tiong2023review, spanninger2022review, spanninger2020approaches}, suggesting that effective delay prediction depends not only on recent train operations, but also on network structure and exogenous conditions. RIDE follows this pattern by combining operational event data with timetable and infrastructure information published by Infrabel, the manager of the Belgian railway network, whose published data have previously supported delay propagation and prediction studies \cite{elliker2026simulation, dekker2022modelling}, and weather-derived features from Open-Meteo, which has recently been incorporated into train delay datasets \cite{wu2025railway}.

\paragraph{Models.}
Prior work has also explored a broad range of modeling approaches, which can be grouped into three families: non-learning approaches, statistical learning methods, and deep learning models. Non-learning approaches include rule-based formulations relying for example on graph models or Markov chains \cite{spanninger2022review, spanninger2020approaches, tiong2023quantitative}. Statistical learning methods include linear regression, support vector machines, random forests, and gradient-boosted trees \cite{tiong2023review, spanninger2022review, spanninger2020approaches}. Deep learning models include multilayer perceptrons, recurrent architectures such as LSTMs, and more recent transformer-based and graph-neural approaches \cite{tiong2023review, spanninger2022review, arthaud2024transformers, li2024railway, huang2024explainable}. Despite this diversity of modeling approaches, individual studies typically compare only a narrow subset of methods under study-specific datasets, prediction tasks, metrics, scopes, and input assumptions. As a result, the relative strengths of different model families remain only partially understood, motivating a benchmark that evaluates them on common prediction instances under a shared protocol.

Taken together, these considerations motivate RIDE as both a public dataset release organized into processing layers and a standardized benchmark framework. The processed release provides a reusable foundation from which downstream users can construct alternative datasets for their own modeling choices, while the benchmark defines a common prediction target and unified evaluation protocol used across a wide range of model families in our experiments.


\section{RIDE Dataset}

\begin{figure}[t]
  \centering
  \includegraphics[width=0.84\linewidth]{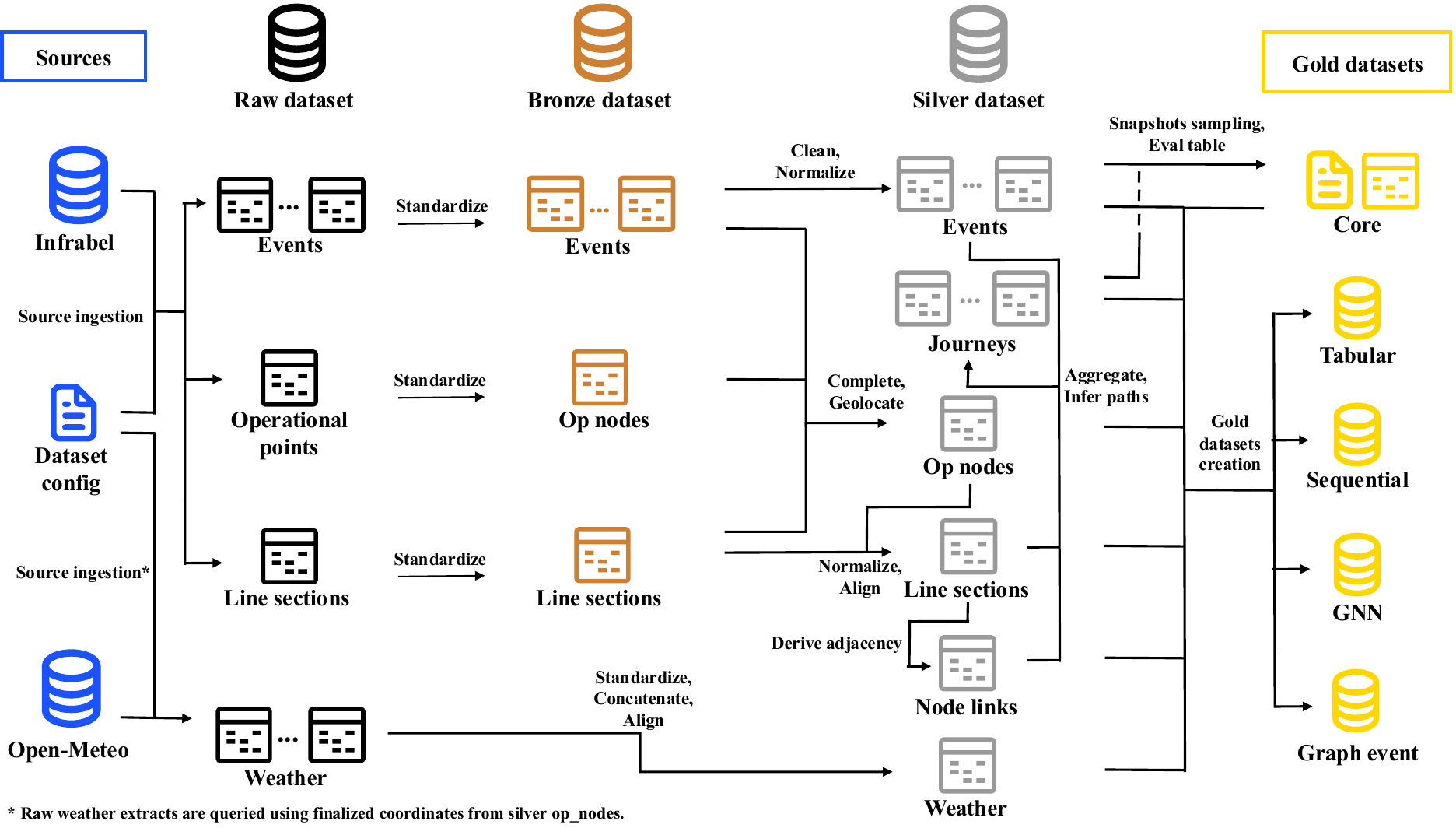}
  \caption{Overview of the RIDE data processing pipeline from sources to gold benchmark datasets.}
  \label{fig:data-processing}
\end{figure}

As a first major contribution, we introduce the RIDE dataset, organized as a layered data pipeline that transforms raw railway and weather data into both a reusable intermediate release and model-ready benchmark datasets. It covers three years of passenger railway operations from 2023 to 2025 over the Belgian railway network at nationwide scale, and includes a broad range of rail traffic, from local and intercity to high-speed trains. As illustrated in Figure~\ref{fig:data-processing}, RIDE is designed as a tiered release, organized into four stages: \textit{raw}, \textit{bronze}, \textit{silver}, and \textit{gold}. This tiered design is intended to support different levels of reuse: the raw and bronze stages support source acquisition and standardization, the silver stage provides a processed relational dataset that can be adapted to alternative modeling choices, and the gold stage provides concrete benchmark datasets tailored to different delay prediction approaches while sharing common splits and evaluation targets. We provide full details of the processing pipeline and resulting datasets in Appendix~\ref{app:dataset-details}.

\subsection{From Data Sources to the Bronze Stage}

RIDE is built from two complementary public data sources. The first is the Infrabel Open Data portal \cite{infrabel_opendata}, which provides the core railway data used throughout the pipeline, including train movement records with scheduled and observed timings, as well as infrastructure tables describing operational points and railway line sections. The second source is Open-Meteo \cite{Zippenfenig_Open-Meteo}, from which we obtain hourly historical weather observations aligned with operational points. The \textit{bronze} stage provides a standardized ingestion layer for the acquired data. Conceptually, it performs schema harmonization, identifier and type normalization, coordinate extraction, and lightweight integrity checks while preserving the source-level structure of the data.

\subsection{Silver Release: Relational Dataset}

The \textit{silver} release provides the reusable intermediate representation of RIDE: a relational dataset over events, journeys, infrastructure, and weather tables, as described in Table~\ref{tab:silver_components}. It spares users from reproducing the core data preparation steps required to build a usable train delay prediction dataset, including cleaning, standardization, completion of missing or inconsistent information, and enrichment. In particular, the silver tier completes missing operational point information, reconstructs the railway network topology, sanitizes event timelines to enforce journey-level consistency, and infers the exact path of trains throughout the railway network. It provides a standardized foundation from which downstream users can construct alternative model-specific datasets.

\begin{figure}
  \centering
  \includegraphics[width=8cm]{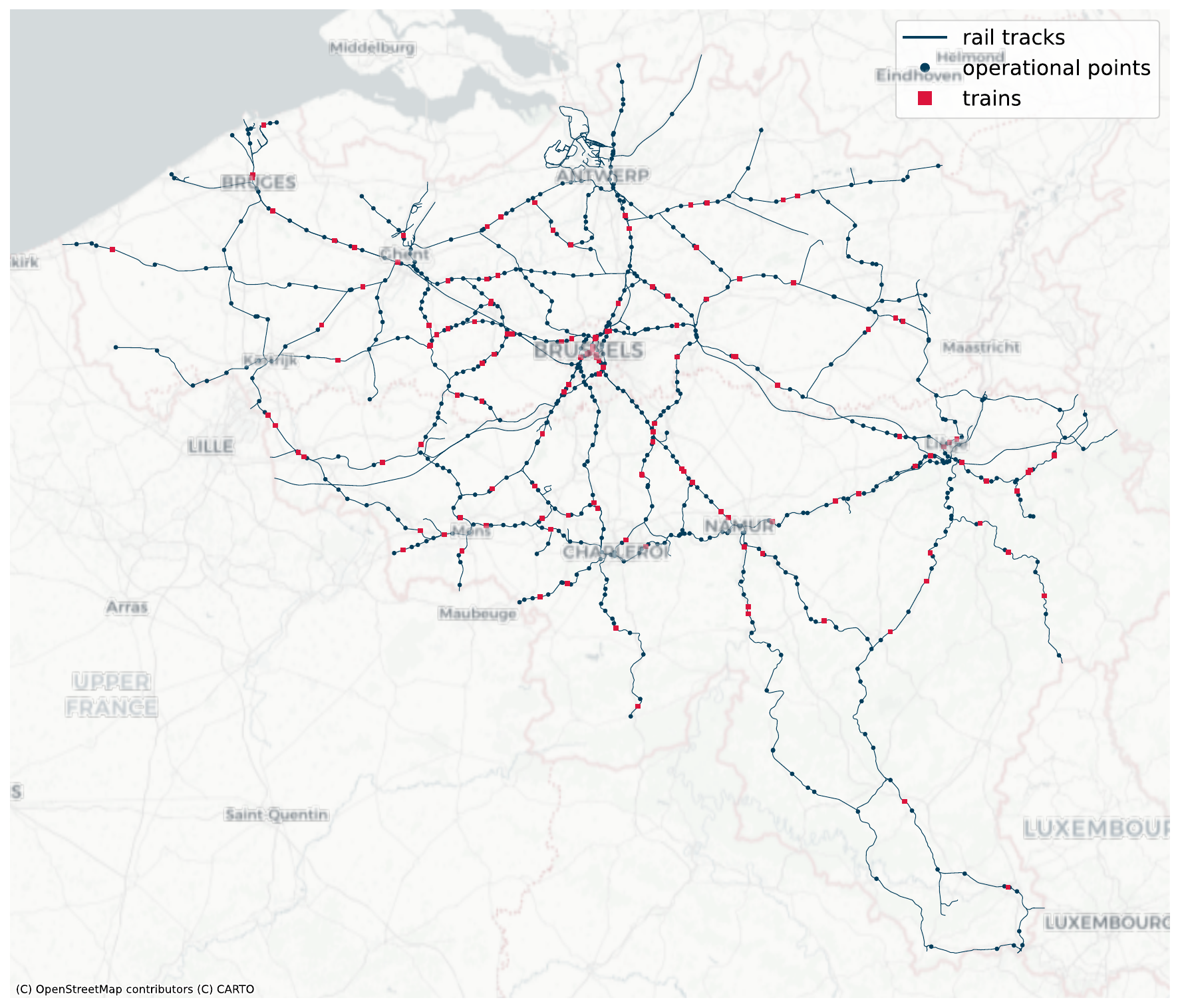}
  \caption{Visualization of the railway network for the snapshot at \texttt{2024-01-15 08:00:00}.}
  \label{fig:snapshot-fig}
\end{figure}

\renewcommand{\arraystretch}{1.15}
\begin{table}[t]
\centering
\small
\begin{tabular}{l p{10.2cm}}
\toprule
\texttt{Table} & Description \\
\midrule
\texttt{events} & train events, each corresponding to one train at one operational point with scheduled and observed timings, forming the core temporal table for delay prediction \\
\texttt{journeys} & journeys, each corresponding to one train itinerary on one day, with train-level metadata, journey statistics, and inferred exact paths through the railway network \\
\texttt{op\_nodes} & operational points with names, types, and coordinates, serving as the nodes of the railway network \\
\texttt{line\_sections} & fine-grained geospatial railway sections of the network whose geometry passes through multiple operational points \\
\texttt{node\_links} & links between consecutive operational points with distances, serving as the graph edges induced by physical rail connections \\
\texttt{weather} & hourly weather observations aligned to operational points, providing exogenous context for downstream models \\
\bottomrule
\end{tabular}
\caption{Description of the Silver release tables.}
\label{tab:silver_components}
\end{table}
\renewcommand{\arraystretch}{1.0}

The silver release contains 1,355 operational points, 1,212 line sections, and 1,797 node links. Across the 2023--2025 period, it includes 3.6 million journeys and 94.5 million train events, corresponding to averages of 100,301 journeys and 2.6 million events per month. The weather table covers this period with hourly observations aligned to operational points, for a total of 35.7 million weather records.

\subsection{Gold Release: Benchmark Datasets}

The \textit{gold} release is the model-ready tier of RIDE, materializing the silver data into two types of artifacts: (1) a shared benchmark \textit{core}, which defines the common train/test data splits, targets, and evaluation metadata, and (2) model-specific datasets tailored to different prediction approaches. Its central concept is the \textit{snapshot}, which serves as the fundamental unit of the benchmark. A snapshot represents the state of the railway network at a given time: it contains information on active trains (including schedule, past delays, \dots), which is then used as model inputs for predicting future delays (Figure~\ref{fig:snapshot-fig}). The gold datasets are defined over a common list of training snapshots and testing snapshots provided by the benchmark \textit{core}. The training and test sets are extracted from separate time periods, with training snapshots drawn from 2023--2024 and test snapshots drawn from 2025, so that evaluation reflects a realistic forward-in-time prediction setting. The benchmark \textit{core} also provides a dedicated table containing all information required to compute test-set metrics. It is designed to support new gold datasets provided by downstream users, with alternative feature designs or model-specific representations, while preserving the same snapshot splits and test evaluation table under a common evaluation protocol, so that all models are compared on the same prediction instances and target values. 

On top of this shared \textit{core}, the gold release provides four model-specific datasets. The \textit{tabular} dataset represents each train instance of each snapshot as a fixed-size feature vector and is used by the MLP, XGBoost, and Transformer models. The \textit{sequential} dataset arranges the tabular features into event sequences and static features for the LSTM model. The \textit{GNN} dataset represents each snapshot as a heterogeneous graph with node and edge features, used by the GNN model. The \textit{graph-event} dataset computes aggregate statistics and retains relevant information from the silver dataset for the Graph-event model. The feature families used in the \textit{tabular}, \textit{sequential}, and \textit{GNN} datasets are summarized in Table~\ref{tab:gold_feature_families}, while full schema and construction details are provided in Appendix~\ref{app:gold-dataset-details}.

The gold release is provided in two tiers, \textit{lite} and \textit{standard}, which cover the same 2023--2025 time span but differ in benchmark scale: \textit{lite} contains 15,000 training snapshots and 3,000 test snapshots, whereas \textit{standard} contains 50,000 training snapshots and 10,000 test snapshots, with an average of 210 active trains per snapshot.  The \textit{lite} tier provides a smaller benchmark configuration intended for faster experimentation and lower computational requirements, whereas the \textit{standard} tier provides the full benchmark setting used in Section~\ref{sec:experiments}.



\begin{table}[t]
\centering
\small
\renewcommand{\arraystretch}{1.15}
\begin{tabular}{l p{10.2cm}}
\toprule
Feature family & Description \\
\midrule
train information & train operator and train category indicators derived from journey metadata \\
snapshot time features & cyclical and calendar features derived from the prediction snapshot timestamp \\
past event context & planned timings, delays, event types, and node embeddings for 15 previous events \\
future event context & planned timings, event types, and node embeddings for 15 future events \\
local network context & context for each oriented rail segment and operational points, including distance, number of trains, and average delays \\
weather features & weather variables for each operational point at prediction time, including temperature, rain, snowfall, humidity, and wind speed \\
\bottomrule
\end{tabular}
\caption{Feature families in gold machine learning datasets.}
\label{tab:gold_feature_families}
\end{table}
\renewcommand{\arraystretch}{1.0}


\section{RIDE Benchmark}
\label{sec:experiments}

As a second major contribution, we introduce a unified benchmark built on top of the RIDE dataset, with a common prediction task, fixed temporal splits, and a shared evaluation protocol across model families, and we provide benchmark experiments across a diverse set of models.

\subsection{Delay Prediction Task}
\label{sec:prediction-task}

For a given snapshot $s$ at time $t^{s}$, the goal is to predict, for every active train, the delay at the next $n$ scheduled events, where each event corresponds to a scheduled arrival, departure, or passage at an operational point (see Table~\ref{tab:silver_components}). For a train $i$ in snapshot $s$, we define the delay at future scheduled event $j \in \{1,\ldots,n\}$ as $d_{s,i,j} = t^{\mathrm{obs}}_{s,i,j} - t^{\mathrm{sched}}_{s,i,j}$, where $t^{\mathrm{sched}}_{s,i,j}$ and $t^{\mathrm{obs}}_{s,i,j}$ denote the scheduled and observed times of that event, respectively. The model prediction is denoted by $\hat{d}_{s,i,j}$. In this benchmark, we set $n=15$, which corresponds to an average prediction horizon of approximately 40 minutes, a range relevant to operational use cases such as passenger information updates and short-term dispatching decisions.

For later use in this section, we define $d_{s,i}^{\mathrm{last}}$ as the last known delay of train $i$ at snapshot $s$. It is computed from the previous observed event delay $d_{s,i,-1}$ with an adjustment ensuring that the implied time of the next event does not fall before the snapshot: $d_{s,i}^{\mathrm{last}} = d_{s,i,-1} + \max\!\left(0,\; t^{s} - t_{s,i,1}^{\mathrm{sched}} - d_{s,i,-1}\right)$, where $t_{s,i,1}^{\mathrm{sched}}$ is the scheduled time of the next event.

\subsection{Evaluation Metrics}

We evaluate models using mean absolute error (MAE) and root mean squared error (RMSE), computed over the set $\mathcal{E}$ containing all prediction targets across all evaluation snapshots. Each element $(s,i,j) \in \mathcal{E}$ corresponds to a future event $j$ of a train instance $i$ in snapshot $s$. Then:
$$
\mathrm{MAE} = \frac{1}{|\mathcal{E}|} \sum_{(s,i,j)\in\mathcal{E}} |\hat{d}_{s,i,j} - d_{s,i,j}|,
\quad
\mathrm{RMSE} = \sqrt{\frac{1}{|\mathcal{E}|} \sum_{(s,i,j)\in\mathcal{E}} (\hat{d}_{s,i,j} - d_{s,i,j})^2}.
$$

Since delays are expressed in seconds, both metrics are also expressed in seconds and are therefore directly interpretable. While these aggregate metrics provide a global measure of predictive accuracy, they can hide important variations across prediction settings. We therefore complement MAE and RMSE with breakdowns along two dimensions:
\begin{itemize}
\item prediction horizon, defined by how far in the future the event occurs, i.e., $t^{\mathrm{obs}}_{s,i,j} - t^{s}$.
\item delay-delta bins, corresponding to delay change, defined from the difference between the target delay and the last known delay at prediction time, $d_{s,i,j} - d_{s,i}^{\mathrm{last}}$.
\end{itemize}
These complementary breakdowns make it possible to analyze the strengths and weaknesses of models across different forecasting regimes.

\subsection{Evaluation Protocol}

The benchmark is built from fixed training and test snapshot splits defined once in the shared gold benchmark \textit{core}. A single standardized \texttt{test\_eval\_table} is constructed from the test snapshots and used for final evaluation across all model families, ensuring that every method is evaluated on the same prediction instances and target values. For learning-based models, hyperparameters are selected on the training split using Optuna \cite{akiba2019optuna}, optimizing validation MAE on the last 10\% of training snapshots in temporal order. Final benchmark results are then obtained by retraining each model with the selected hyperparameters on 10 seeds and evaluating on the common test split. In the main paper, we report results on the \textit{standard} tier only; full details for training, hyperparameter search, and computation budgets are provided in Appendices~\ref{app:training_procedure}, \ref{app:hparamandconfigs} and~\ref{app:resourcesandbudget}.

\subsection{Models}

This subsection presents the models considered in our experiments. We compare three broad families of approaches under the common RIDE benchmark setting: non-learning models, namely the Translation and Graph-event models, which provide deterministic reference points for delay prediction; statistical learning models, represented here by XGBoost, which remains a strong tabular predictor for structured data; and deep learning models, namely MLP, LSTM, Transformer, and GNN, which can exploit richer sequential, relational, and interaction structure. Due to the diversity of prediction targets, operational scopes, and data representations used in the literature, several prior models cannot be adopted directly in a fully comparable form. We therefore implement benchmark-specific versions inspired by the literature but adapted to the common RIDE prediction setting and evaluation protocol. For all learning-based models, we predict delay relative to $d_{s,i}^{\mathrm{last}}$ at prediction time and recover absolute delay by adding $d_{s,i}^{\mathrm{last}}$ back at inference time. This reparameterization leaves the benchmark target unchanged but was found to improve predictive performance. These models are built from the same underlying feature families, which are structured differently depending on the architecture. Additional architectural and training details are provided in Appendix~\ref{app:modeldetails}.

\paragraph{Translation.} A commonly used baseline in operational railway settings is the Translation \cite{arthaud2024transformers, wang2015data}. For every target event, it predicts the last known delay $d_{s,i}^{\mathrm{last}}$. Despite its simplicity, this baseline is a strong reference point for evaluating the added value of more complex models.

\paragraph{Graph-event.}
The graph-event model is another non-learning rule-based approach for train delay prediction, inspired by prior graph- and event-based delay-propagation methods \cite{goverde2010delay, wei2015modeling}. Starting from each prediction snapshot, it simulates the future evolution of active trains over the railway network using the current train positions and schedules, graph connectivity, and simple precedence constraints between trains sharing infrastructure. Expected travel times between consecutive events are obtained from empirical travel-time statistics estimated from the training data, and delays are then propagated forward event by event to produce the future delay predictions. This model provides a deterministic graph-based reference point that contrasts with the learning-based benchmark models.

\paragraph{XGBoost.} We also include an XGBoost model serving as a strong statistical learning model, again operating on the same fixed-size tabular representation used for the MLP \cite{du2025xgboost, gao2024data}. In this setting, we train one independent XGBoost regressor for each future event slot, so that each model predicts the delay at a fixed position among the next $n$ scheduled events from the same tabular features.

\paragraph{MLP.} We include a standard multilayer perceptron model applied to the fixed-size tabular representation of each train at each snapshot \cite{peters2005prediction, oneto2018train}. The model maps the tabular input features directly to predicted delays at future events.

\paragraph{LSTM.} Sequential models are a natural fit for train delay prediction, and recurrent neural architectures, in particular LSTMs, have been explored for this task \cite{huang2020modeling, yu2023delay}. To match our multi-step prediction setting, we use an encoder-decoder LSTM that operates on one train at a time: the encoder processes the past event sequence, static non-sequential features are embedded with an MLP, and the decoder combines this context with known future event features to predict delays at future events.

\paragraph{Transformer.} Transformer-based models have recently been used for train delay prediction \cite{elliker2026simulation, arthaud2024transformers}. The model adopts an encoder-only architecture in which each train in the snapshot is represented as a token, allowing trains to attend to one another through self-attention mechanisms before predicting delays at future events.

\paragraph{GNN.} Graph-based models are also a natural fit for train delay prediction, and graph neural network (GNN) approaches have been explored in the literature \cite{li2024railway, huang2024explainable}. To match our prediction setting, we implement a heterogeneous GNN following this line of work: each snapshot is represented as a graph with train nodes and station nodes, edges between adjacent stations, and train-to-station edges for past and future stations encoding train itineraries. Message passing and edge updates are performed over this graph with heterogeneous GINE-style node aggregations \cite{hu2020gpt}, and delays at future events are predicted on the future train-to-station edges.

\subsection{Experimental Results and Analysis}

\begin{table}[t]
\centering
\begin{tabular}{lll}
\toprule
Model & MAE & RMSE \\
\midrule
Translation & 96.65 & 233.42 \\
Graph-event & 88.41 & 232.48 \\
MLP & 77.20 $\pm$ 0.04 & 203.21 $\pm$ 0.40 \\
 XGBoost & 76.58 $\pm$ 0.01 & 203.46 $\pm$ 0.02 \\
LSTM & 74.62 $\pm$ 0.27 & 202.63 $\pm$ 0.77 \\
Transformer & 74.54 $\pm$ 0.25 & 195.39 $\pm$ 0.59 \\
GNN & \textbf{73.62 $\pm$ 0.19} & \textbf{194.56 $\pm$ 0.88} \\
\bottomrule
\end{tabular}
\caption{Standard-tier test performance. MAE/RMSE in seconds; $\pm$: std. over 10 seeds.}
\label{tab:standard_main_results}
\end{table}

\paragraph{Aggregate Results.} Table~\ref{tab:standard_main_results} summarizes the main results on the standard benchmark tier. The GNN achieves the best mean performance, with 73.62 MAE and 194.56 RMSE, while the Transformer and LSTM follow very closely at 74.54 and 74.62 MAE, respectively. Among tabular learning-based models, XGBoost outperforms the MLP on MAE (76.58 vs.\ 77.20), while the MLP remains marginally better on RMSE (203.21 vs.\ 203.46). The graph-event model clearly improves over the simple translation baseline (88.41 vs.\ 96.65 MAE), yet still remains noticeably behind the learning-based approaches. Despite its simplicity, the translation baseline is surprisingly strong, providing a reference point that is difficult to beat by a large margin. Interestingly, the LSTM yields strong performance, close to the Transformer, despite operating on one train at a time and lacking an explicit mechanism to propagate information through the network; this suggests that effective single-train sequence modeling, combined with local network context features, already captures much of the signal needed for this benchmark. This may also reflect, in part, the relative scarcity of situations where knowing the precise position and itinerary of each train is required for accurate delay prediction. More broadly, the strongest learning-based models remain close to each other in absolute terms, so the benchmark does not point to a single overwhelmingly dominant architecture. Taken together, these results indicate that architectural changes alone do not lead to dramatic gains on this benchmark, and that further progress may also depend on better feature design and problem-specific modeling choices. Appendix~\ref{app:standard-tier-additional-figures} provides complementary visualizations of the standard-tier results.

\begin{table}[t]
\centering
\setlength{\tabcolsep}{3pt}
\begin{tabular}{lllllllllll}
\toprule
Model & 0:5 & 5:10 & 10:15 & 15:20 & 20:25 & 25:30 & 30:35 & 35:40 & 40:45 & 45+ \\
\midrule
Translation & 37.5 & 57.2 & 72.3 & 84.9 & 95.7 & 107.4 & 121.7 & 141.7 & 166.9 & 312.2 \\
Graph-event & 30.5 & 48.5 & 63.1 & 75.2 & 86.3 & 99.0 & 115.2 & 135.2 & 159.1 & 305.9 \\
MLP & 25.9 & 42.2 & 55.2 & 66.4 & 76.3 & 87.0 & 99.9 & 116.9 & 138.6 & 266.8 \\
XGBoost & \textbf{24.0} & 41.1 & 54.7 & 66.2 & 76.0 & 86.7 & 99.6 & 116.9 & 138.4 & 267.4 \\
LSTM & 24.1 & \textbf{39.7} & \textbf{52.6} & 63.6 & 73.3 & 84.1 & 97.1 & 114.3 & 136.0 & 264.3 \\
Transformer & 25.2 & 41.2 & 53.9 & 64.5 & 73.9 & 84.3 & 96.7 & 112.8 & 133.3 & 252.6 \\
GNN & 25.8 & 41.4 & 53.2 & \textbf{63.3} & \textbf{72.3} & \textbf{82.3} & \textbf{94.5} & \textbf{110.4} & \textbf{130.6} & \textbf{252.2} \\
\bottomrule
\end{tabular}
\caption{Test-set MAE in seconds by prediction horizon on the standard tier. Bins are in minutes.}
\label{tab:standard_horizon_mae}
\end{table}

\paragraph{Results by Prediction Horizon.} 
Table~\ref{tab:standard_horizon_mae} breaks down performance by prediction horizon. A clear pattern emerges: XGBoost performs best in the very shortest 0:5 minute bin, although the LSTM remains extremely close, and the LSTM is then strongest from 5 to 15 minutes. From 15 minutes onward, the GNN becomes the best-performing model across all remaining horizon bins, while the Transformer remains competitive throughout without clearly dominating either the short- or long-horizon regime. The strong 0:5 minute performance of XGBoost is also consistent with the delay-change analysis below, where it performs particularly well in the dominant regime of small delay changes, which is itself especially prevalent at very short horizons. One important caveat, however, is that the latest horizon bins are progressively biased toward trains that accumulate delay, with this effect being particularly pronounced in the final 45+ bin, since these bins become thinner by construction under the fixed future-event prediction window. This phenomenon is illustrated in Appendix~\ref{app:predtaskandmetricsdetails}. As a result, performance differences across horizon bins should not be interpreted as a pure horizon effect alone, but rather as reflecting a mixture of horizon length and delay regime.

\begin{table}[t]
\centering
\setlength{\tabcolsep}{3pt}
\begin{tabular}{llllllllllll}
\toprule
Model & <-5 & -5:-2 & -2:-1 & -1:-0.5 & -0.5:0 & 0:0.5 & 0.5:1 & 1:2 & 2:5 & 5:10 & 10+ \\
\midrule
Translation & 428.1 & 174.8 & 85.2 & 44.3 & \textbf{14.7} & \textbf{13.5} & 43.2 & 85.1 & 184.8 & 408.5 & 1260.2 \\
Graph-event & \textbf{169.6} & \textbf{81.4} & 53.2 & 39.3 & 31.2 & 32.2 & 45.7 & 75.9 & 169.5 & 398.2 & 1259.3 \\
MLP & 235.5 & 92.7 & 53.4 & 37.2 & 28.9 & 28.1 & 37.2 & 61.0 & 136.8 & 322.9 & 1053.6 \\
XGBoost & 265.3 & 100.7 & 56.5 & 37.2 & 24.9 & 23.9 & \textbf{35.4} & 59.8 & 135.7 & 324.0 & 1062.3 \\
LSTM & 200.3 & 81.5 & \textbf{49.1} & \textbf{35.5} & 29.0 & 28.7 & 37.2 & \textbf{59.2} & 131.3 & 314.3 & 1046.4 \\
Transformer & 197.1 & 81.6 & 50.6 & 36.9 & 27.8 & 27.8 & 39.5 & 63.7 & 136.6 & 312.3 & \textbf{968.5} \\
GNN & 197.8 & 82.0 & 50.2 & 36.9 & 29.2 & 28.6 & 38.4 & 61.2 & \textbf{131.0} & \textbf{299.5} & 970.5 \\
\bottomrule
\end{tabular}
\caption{Test-set MAE in seconds by delay-delta bin on the standard tier. Bins are in minutes; negative values correspond to delay recovery and positive values to delay accumulation.}
\label{tab:standard_delay_delta_mae}
\end{table}

\paragraph{Results by Delay Change.} Table~\ref{tab:standard_delay_delta_mae} reports performance by delay-delta bin. As expected, the translation baseline is strongest around zero delay change, where propagating the current delay provides a strong approximation. Excluding this baseline, XGBoost is the strongest model in the central delay-delta regime around zero, roughly from $-0.5$ to $1$ minute, corresponding to the most common regime in the data: small delay changes. This suggests that XGBoost tends to make comparatively conservative predictions. The LSTM performs particularly well on moderate negative delay-delta bins, indicating that a sequential prior is well suited for modeling how trains catch up on the delay. For moderate delay accumulation, between 0.5 and 2 minutes, tabular and sequential models tend to outperform the GNN and Transformer, suggesting that, in this regime, the fixed train-level feature representation already contains enough information to support accurate predictions, without requiring explicit modeling of network interactions. At the opposite end, the graph-event model performs best on large negative delay deltas, i.e.\ strong delay recovery, which is consistent with its tendency to predict travel times that are on average faster than the scheduled ones. In contrast, the GNN performs best across most large positive delay-delta bins, while in the most extreme accumulation regime it remains essentially tied with the Transformer. This suggests that models with learned interaction patterns between trains are better able to capture substantial delay increases, likely because these are tied to propagation effects in the network.

Together, these two evaluation breakdowns provide a more informative view of model behavior than aggregate MAE and RMSE alone. They make it possible to identify which models perform best in specific forecasting regimes, such as short- versus long-horizon prediction or delay recovery versus delay accumulation, and thereby support broader insights into the dynamics of train delay prediction.


\section{Conclusion}

In this work, we introduced RIDE, an open dataset and benchmark for train delay prediction built at nationwide scale over the Belgian railway network. RIDE is the first resource to provide both a reusable intermediate silver release, intended as a standardized foundation for downstream dataset construction, and model-ready gold benchmark datasets, together with a unified evaluation protocol that we use to compare the non-learning Translation baseline and Graph-event model, the statistical learning model XGBoost, and the deep learning models MLP, LSTM, Transformer, and GNN. Our benchmark results confirm a clear gap between learning-based and non-learning models, while differences among the strongest learning-based models remain comparatively small, with graph neural networks achieving the best mean performance. The competitiveness of recurrent, attention-based, and tabular models, despite their different inductive biases, suggests that further progress may depend as much on feature design as on architecture. More broadly, the proposed evaluation protocol goes beyond aggregate accuracy by enabling regime-specific analysis of model behavior, which helps reveal where different approaches succeed or fail and supports more nuanced insights into train delay prediction. While some findings may depend on the specific characteristics of the Belgian passenger railway network, we hope that RIDE will provide a useful foundation for future work and facilitate progress on delay prediction, delay propagation, and learning over large-scale transportation systems.

{
\bibliographystyle{unsrt}
\bibliography{refs}
}


\appendix

\section{Data Release, Ethics, and Reproducibility}

\subsection{Data and Code Availability}

All RIDE data releases and accompanying code are publicly available at the following links.

\textbf{Code} -- \href{https://github.com/orailix/ride}{GitHub repository}\\
\textbf{Silver} -- \href{https://huggingface.co/datasets/orailix/ride-silver}{Hugging Face dataset}\\
\textbf{Gold Standard} -- \href{https://huggingface.co/datasets/orailix/ride-gold-standard}{Hugging Face dataset}\\
\textbf{Gold Lite} -- \href{https://huggingface.co/datasets/orailix/ride-gold-lite}{Hugging Face dataset}

The full data-processing and benchmark pipeline can be reproduced end to end using the instructions provided in the GitHub repository.

\subsection{Licenses}

The source data used to construct RIDE are released under open-data licenses. The Infrabel punctuality, operational-point, and railway-line-section datasets are published through Infrabel Open Data under the CC0 Universal open license. The weather variables are obtained from the Open-Meteo Archive API, whose API data are provided under the Creative Commons Attribution 4.0 International license (CC-BY 4.0).

The released RIDE datasets are distributed under CC-BY 4.0. This license is chosen to preserve the attribution requirements of the Open-Meteo-derived weather variables while remaining compatible with the CC0 status of the Infrabel source data. Users of RIDE should attribute RIDE, Infrabel, and Open-Meteo, and should not imply endorsement by any source-data provider. The accompanying source code is released under the MIT license.

\subsection{Ethics Statement}

We use open railway operational data and weather data released under licenses that permit reuse. The dataset contains only non-personal operational information, such as train times, stations, delays, infrastructure metadata, and weather observations, and no data about individual passengers or staff. Our use of the data therefore complies with the providers' licenses and does not raise additional privacy concerns.

\subsection{Broader Impact}

RIDE is intended to support research on train delay prediction and delay propagation, with potential downstream benefits for passenger information, dispatching, traffic management, and the reliability of public transport systems. We do not identify direct negative societal impacts from releasing this benchmark: it is built from open operational and weather data, contains no personal information about passengers or staff, and is intended for offline research and evaluation rather than direct automated operational control.

\subsection{Acknowledgements}

This work received financial support from SNCF through the research chair ``AI and optimization for mobility'' with École Polytechnique. This work was granted access to the HPC resources of IDRIS under the allocation AD011015656R1 made by GENCI. Finally, we thank Alexi Canesse, Benoît Goupil and Mahammed El Sharkawy for helpful discussions and feedback on this work.

\section{Dataset Construction and Analysis} 
\label{app:dataset-details}

This section provides additional detail on the raw, bronze, silver, and gold stages of RIDE. It describes the original data sources, the main processing steps used to transform them into the released datasets and benchmark-ready artifacts, and additional analyses and visualizations of the resulting datasets.

\subsection{Sources}

\paragraph{Infrabel Open Data portal.}
The railway data used in RIDE are obtained from the Infrabel Open Data portal \cite{infrabel_opendata}. Infrabel is the Belgian railway infrastructure manager and publishes open datasets covering punctuality, infrastructure, safety, and other aspects of the Belgian rail network. RIDE relies on this portal as the source of both operational train movement data and infrastructure descriptions, which are later combined to construct event, journey, and railway-network representations. The data are made available by Infrabel under the portal's open-data terms, which publish the datasets under a CC0 Universal open licence.

\paragraph{Open-Meteo.}
Weather data are obtained from the Open-Meteo Historical Weather API \cite{Zippenfenig_Open-Meteo}. Open-Meteo provides open access to historical meteorological time series through an archive endpoint queried by geographic coordinates, date range, timezone, and selected weather variables. RIDE uses this source to add exogenous weather context to the railway data, using operational-point coordinates as the spatial interface between railway events and weather observations. Open-Meteo API data are provided under the Creative Commons Attribution 4.0 International licence (CC BY 4.0).

\subsection{Raw Data}

The raw data layer contains four source datasets: three railway datasets from the Infrabel Open Data portal and one weather dataset derived from the Open-Meteo Historical Weather API. The Infrabel sources provide complementary views of railway operations and infrastructure: monthly punctuality files describe train movements, while the operational-points and line-sections catalogues provide the spatial and topological reference needed to interpret those movements on the Belgian railway network.

\paragraph{Raw punctuality files.}
We download all 36 monthly raw punctuality files covering January 2023 through December 2025. These files contain arrival and departure records for domestic and international passenger trains at their stopping points, indexed by service day and train number. They provide the core operational information used in RIDE, including train identifiers, relations, operators, operational point identifiers, scheduled and observed arrival and departure times, line identifiers, and delays.

\paragraph{Operational points.}
The operational-points catalogue lists the railway network's operational points together with their identifiers, geographic positions, official names, and Infrabel internal classifications. This table provides the spatial anchor used to associate train events with locations in the Belgian railway network.

\paragraph{Line sections.}
The railway line-sections catalogue describes the segmentation of the rail network into line sections, including section geometries, endpoint operational points, line identifiers, and technical infrastructure attributes. This source provides the geometric and topological basis from which RIDE reconstructs railway links between operational points.

\paragraph{Weather.}
For each operational point in the finalized silver-stage \texttt{op\_nodes} table, we query hourly weather data from the Open-Meteo Historical Weather API. The queried period covers the full 2023--2025 railway observation window, extended by one day on both sides, from 2022-12-31 to 2026-01-01 to account for edge cases. Each query uses the operational point latitude and longitude, the \texttt{Europe/Brussels} timezone, and the following hourly variables: \texttt{temperature\_2m}, \texttt{rain}, \texttt{snowfall}, \texttt{relative\_humidity\_2m}, \texttt{wind\_speed\_10m}, and \texttt{weather\_code}. Because these coordinates are only finalized after the silver operational-node construction, weather acquisition is performed after the silver stage.

\subsection{Bronze Data}

The bronze layer provides a standardized version of the raw source files while preserving their original informational content. Its purpose is to make the heterogeneous raw inputs easier to process in later stages, rather than to perform dataset-level cleaning or modeling decisions. The bronze layer keeps the same table structure as the raw layer: no source table is added or removed, so we describe it globally rather than table by table. In this layer, source-specific column names are mapped to a consistent schema, unused fields are removed, basic data types are normalized, and the resulting CSV inputs are stored as Parquet tables. Temporal fields from the punctuality data are still kept close to their source representation, with full timestamp reconstruction, consistency checks, and journey-level filtering deferred to the silver layer. The exact column mappings and bronze-stage transformations are specified in the bronze manifests released with the code.

The bronze layer therefore acts as a stable interface between raw data acquisition and the relational construction performed in the silver stage. It makes the pipeline reproducible and easier to inspect, while ensuring that substantive transformations, such as operational-node correction, railway-link reconstruction, event cleaning, journey construction, path inference, and weather alignment, remain explicit in the later stages.

\subsection{Silver Release}

The silver release is the main reusable relational data layer of RIDE. It converts the standardized bronze railway inputs and raw weather extracts into a coherent set of railway, weather, and infrastructure tables that can be used independently of the benchmark representations defined in the gold layer. Its purpose is to provide a cleaned and structurally consistent view of Belgian passenger railway operations over the full 2023--2025 period, while preserving enough event-level detail for users to construct alternative feature sets, temporal windows, or model-specific datasets.

\subsubsection{Silver Relational Data}

This subsection documents the structure of the silver release as a relational dataset. In the following paragraphs, we describe the schemas of these tables, their relational structure, and provide a small illustrative example showing how they can be joined.

\paragraph{Table schemas.}
The silver release contains six Parquet tables. The \texttt{events} table (Table~\ref{tab:silver_events_schema}) contains event-level train records with timestamps, event types, delays, and arrival/departure line context. The \texttt{journeys} table (Table~\ref{tab:silver_journeys_schema}) contains one record per operated train service, with journey-level metadata, timing summaries, delay summaries, event counts, and inferred \texttt{deduced\_paths}. The \texttt{op\_nodes} table (Table~\ref{tab:silver_op_nodes_schema}) contains the operational-point catalogue with identifiers, names, types, and coordinates. The \texttt{line\_sections} table (Table~\ref{tab:silver_line_sections_schema}) describes railway line sections through line identifiers, endpoint operational points, geometries, track counts, and ordered matched operational nodes. The \texttt{node\_links} table (Table~\ref{tab:silver_node_links_schema}) contains graph links between consecutive operational nodes along line-section geometries. Finally, the \texttt{weather} table (Table~\ref{tab:silver_weather_schema}) contains hourly weather observations associated with operational points. These tables report the field name, data type, and role of each column for each table, while short notes below the tables clarify non-obvious keys and references.

\begin{table}[H]
\centering
\small
\renewcommand{\arraystretch}{1.2}
\begin{tabular}{
p{0.22\linewidth}
p{0.14\linewidth}
p{0.52\linewidth}
}
\toprule
\textbf{Field} & \textbf{Type} & \textbf{Description} \\
\midrule
\texttt{train\_id}
& string
& Train identifier for the operated service. \\

\texttt{service\_date}
& string
& Scheduled service date associated with the train. \\

\texttt{op\_id}
& integer
& Operational point at which the event occurs. \\

\texttt{event\_type}
& string
& Event type: arrival, departure, or passage. \\

\texttt{arr\_line\_id}
& string/null
& Railway line identifier on the arrival side of the event. \\

\texttt{dep\_line\_id}
& string/null
& Railway line identifier on the departure side of the event. \\

\texttt{planned\_ts}
& timestamp
& Scheduled timestamp of the event. \\

\texttt{observed\_ts}
& timestamp
& Observed timestamp of the event. \\

\texttt{delay\_sec}
& integer
& Event delay in seconds. \\
\bottomrule
\end{tabular}
\vspace{0.5em}
\parbox{0.95\linewidth}{
The pair \texttt{train\_id}, \texttt{service\_date} identifies the events belonging to the same operated train service. The fields \texttt{arr\_line\_id} and \texttt{dep\_line\_id} refer to railway line identifiers, i.e. \texttt{line\_id}, not to \texttt{line\_section\_id}.
}
\caption{Schema of the silver \texttt{events} table.}
\label{tab:silver_events_schema}
\end{table}

\begin{table}[H]
\centering
\small
\renewcommand{\arraystretch}{1.2}
\begin{tabular}{
p{0.22\linewidth}
p{0.11\linewidth}
p{0.57\linewidth}
}
\toprule
\textbf{Field} & \textbf{Type} & \textbf{Description} \\
\midrule
\texttt{train\_id}
& string
& Train identifier for the operated service. \\

\texttt{service\_date}
& string
& Scheduled service date associated with the train. \\

\texttt{train\_relation}
& string
& Train relation code. \\

\texttt{operator}
& string
& Railway operator. \\

\texttt{relation\_direction}
& string
& Route description of the operated service. \\

\texttt{start\_op\_id}
& integer
& Operational point of the first event. \\

\texttt{end\_op\_id}
& integer
& Operational point of the last event. \\

\texttt{start\_planned\_ts}
& timestamp
& Scheduled timestamp of the first event. \\

\texttt{end\_planned\_ts}
& timestamp
& Scheduled timestamp of the last event. \\

\texttt{start\_observed\_ts}
& timestamp
& Observed timestamp of the first event. \\

\texttt{end\_observed\_ts}
& timestamp
& Observed timestamp of the last event. \\

\texttt{events\_count}
& integer
& Number of events in the journey. \\

\texttt{max\_delay\_sec}
& integer
& Maximum observed delay along the journey. \\

\texttt{min\_delay\_sec}
& integer
& Minimum observed delay along the journey. \\

\texttt{deduced\_paths}
& array
& List of inferred \texttt{link\_id} sequences between consecutive events. \\
\bottomrule
\end{tabular}
\vspace{0.5em}
\parbox{0.95\linewidth}{
The field \texttt{deduced\_paths} is a list of paths, with one path per transition between consecutive events; each path is itself a list of \texttt{link\_id}s, and transitions that remain at the same operational point are represented by empty lists.
}
\caption{Schema of the silver \texttt{journeys} table.}
\label{tab:silver_journeys_schema}
\end{table}

\begin{table}[H]
\centering
\small
\renewcommand{\arraystretch}{1.2}
\begin{tabular}{
p{0.12\linewidth}
p{0.08\linewidth}
p{0.37\linewidth}
}
\toprule
\textbf{Field} & \textbf{Type} & \textbf{Description} \\
\midrule
\texttt{op\_id}
& integer
& Operational point identifier. \\

\texttt{op\_name}
& string
& Operational point name. \\

\texttt{op\_type}
& string
& Operational point type or category. \\

\texttt{lat}
& float
& Latitude in WGS84 coordinates. \\

\texttt{lon}
& float
& Longitude in WGS84 coordinates. \\
\bottomrule
\end{tabular}
\caption{Schema of the silver \texttt{op\_nodes} table.}
\label{tab:silver_op_nodes_schema}
\end{table}

\begin{table}[H]
\centering
\small
\renewcommand{\arraystretch}{1.2}
\begin{tabular}{
p{0.26\linewidth}
p{0.10\linewidth}
p{0.54\linewidth}
}
\toprule
\textbf{Field} & \textbf{Type} & \textbf{Description} \\
\midrule
\texttt{line\_section\_id}
& string
& Line-section identifier. \\

\texttt{line\_id}
& string
& Railway line identifier. \\

\texttt{op\_begin\_id}
& integer
& Operational point at the beginning of the line section. \\

\texttt{op\_end\_id}
& integer
& Operational point at the end of the line section. \\

\texttt{nb\_tracks}
& integer
& Number of tracks in the line section. \\

\texttt{geoshape}
& string
& LineString geometry of the line section. \\

\texttt{matched\_op\_node\_ids}
& array
& List of \texttt{op\_id}s aligned with the geometry vertices. \\
\bottomrule
\end{tabular}
\vspace{0.5em}
\parbox{0.95\linewidth}{
The field \texttt{matched\_op\_node\_ids} has the same order as the coordinates in \texttt{geoshape}; entries are null for geometry vertices that are not matched to an operational node. A railway line identified by \texttt{line\_id} may contain multiple line sections.
}
\caption{Schema of the silver \texttt{line\_sections} table.}
\label{tab:silver_line_sections_schema}
\end{table}

\begin{table}[H]
\centering
\small
\renewcommand{\arraystretch}{1.2}
\begin{tabular}{
p{0.22\linewidth}
p{0.11\linewidth}
p{0.57\linewidth}
}
\toprule
\textbf{Field} & \textbf{Type} & \textbf{Description} \\
\midrule
\texttt{link\_id}
& integer
& Node-link identifier. \\

\texttt{u\_node\_id}
& integer
& First endpoint operational node. \\

\texttt{v\_node\_id}
& integer
& Second endpoint operational node. \\

\texttt{line\_section\_id}
& string
& Line section from which the link is derived. \\

\texttt{distance\_m}
& float
& Link distance in meters. \\
\bottomrule
\end{tabular}
\vspace{0.5em}
\parbox{0.95\linewidth}{
Node links are treated as undirected graph edges. The pair \texttt{u\_node\_id}, \texttt{v\_node\_id} is not unique: multiple rows may connect the same endpoints when they are derived from different line sections, so \texttt{link\_id} is the unique link identifier.
}
\caption{Schema of the silver \texttt{node\_links} table.}
\label{tab:silver_node_links_schema}
\end{table}

\begin{table}[H]
\centering
\small
\renewcommand{\arraystretch}{1.2}
\begin{tabular}{
p{0.25\linewidth}
p{0.12\linewidth}
p{0.53\linewidth}
}
\toprule
\textbf{Field} & \textbf{Type} & \textbf{Description} \\
\midrule
\texttt{op\_id}
& integer
& Operational point identifier. \\

\texttt{time}
& timestamp
& Hourly weather timestamp. \\

\texttt{temperature\_2m}
& float
& Air temperature at 2 m above ground, in degrees Celsius. \\

\texttt{rain}
& float
& Rain amount, in millimeters. \\

\texttt{snowfall}
& float
& Snowfall amount. \\

\texttt{relative\_humidity\_2m}
& float
& Relative humidity at 2 m above ground, in percent. \\

\texttt{wind\_speed\_10m}
& float
& Wind speed at 10 m above ground. \\

\texttt{weather\_code}
& integer
& WMO weather code. \\
\bottomrule
\end{tabular}
\vspace{0.5em}
\parbox{0.95\linewidth}{
The field \texttt{time} is stored at hourly resolution and represents the \texttt{Europe/Brussels} local time used when querying Open-Meteo.
}
\caption{Schema of the silver \texttt{weather} table.}
\label{tab:silver_weather_schema}
\end{table}

\begin{figure}[H]
  \centering
  \includegraphics[width=0.8\textwidth]{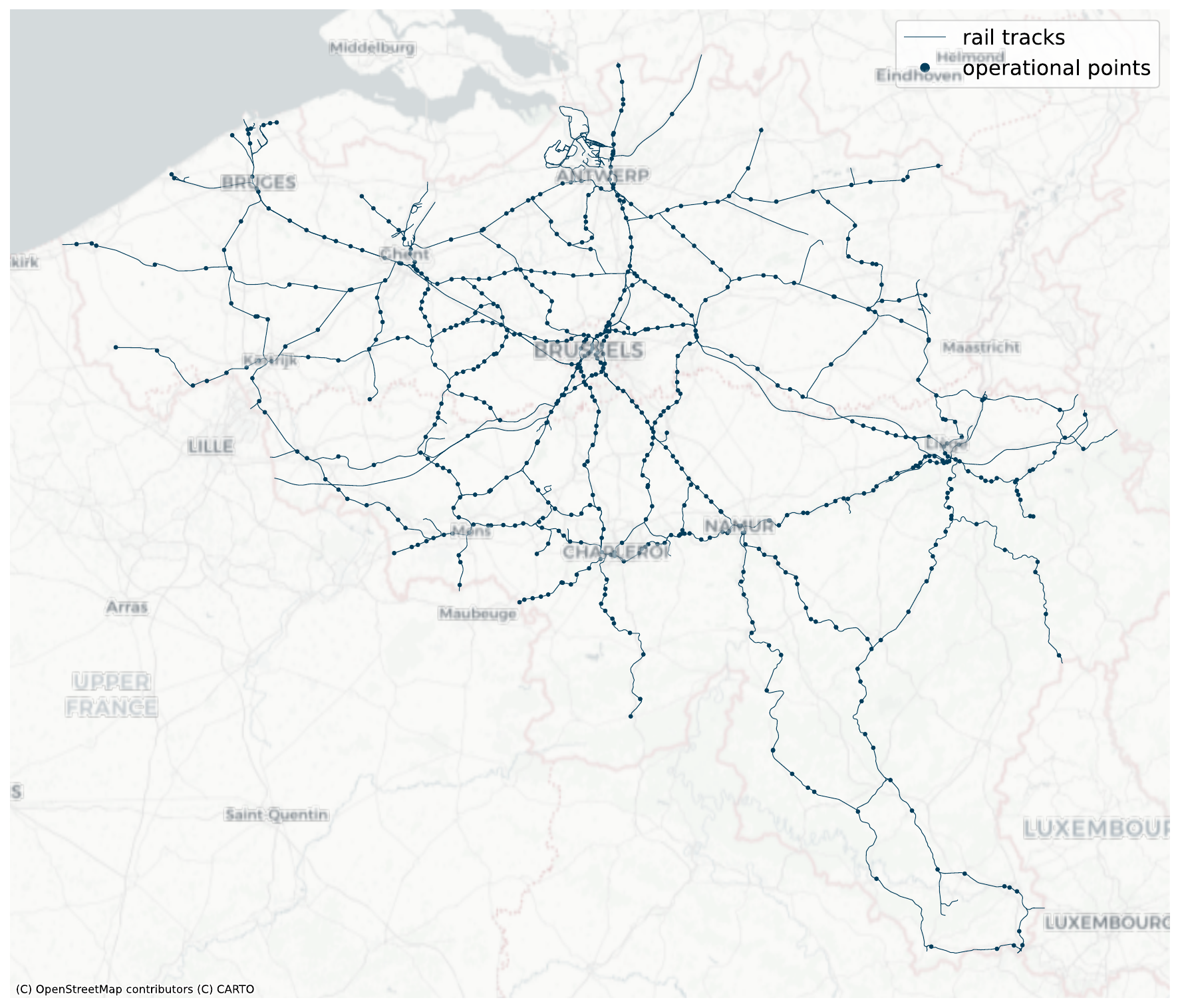}
  \caption{Visualization of the railway network. Some track segments appear to terminate away from a plotted operational point because the published line-section geometries can extend to infrastructure endpoints that are not represented in the published operational-point catalogue.}
  \label{fig:full-network-visualization}
\end{figure}

\begin{figure}[H]
\centering
\small
\resizebox{0.72\linewidth}{!}{
\begin{tikzpicture}[
    tableblock/.style={
        inner sep=0pt,
        font=\small,
        align=center
    }
]
\node[tableblock] (journeys) at (0,0) {
\textbf{\texttt{journeys}}\\[1mm]
\begin{tabular}{|c|c|c|c|c|}
\hline
\rowcolor{gray!15}
\textbf{\texttt{train\_id}} & \textbf{\texttt{service\_date}} & \textbf{\texttt{train\_relation}} & \textbf{\texttt{deduced\_paths}} & \textbf{\texttt{...}} \\
\hline
\texttt{8622} & \texttt{08JAN2023} & \texttt{P} & \texttt{[[354],[],[353,352]]} & \texttt{...} \\
\hline
\end{tabular}
};

\node[tableblock, below=6mm of journeys] (events) {
\textbf{\texttt{events}}\\[1mm]
\begin{tabular}{|c|c|c|c|c|c|c|}
\hline
\rowcolor{gray!15}
\textbf{\texttt{train\_id}} & \textbf{\texttt{service\_date}} & \textbf{\texttt{op\_id}} & \textbf{\texttt{type}} & \textbf{\texttt{planned\_ts}} & \textbf{\texttt{observed\_ts}} & \textbf{\texttt{...}} \\
\hline
\texttt{8622} & \texttt{08JAN2023} & \texttt{801} & \texttt{D} & \texttt{20:15:00} & \texttt{20:15:58} & \texttt{...} \\
\texttt{8622} & \texttt{08JAN2023} & \texttt{788} & \texttt{A} & \texttt{20:19:00} & \texttt{20:19:23} & \texttt{...} \\
\texttt{8622} & \texttt{08JAN2023} & \texttt{788} & \texttt{D} & \texttt{20:20:00} & \texttt{20:20:38} & \texttt{...} \\
\texttt{8622} & \texttt{08JAN2023} & \texttt{818} & \texttt{P} & \texttt{20:26:00} & \texttt{20:26:47} & \texttt{...} \\
\hline
\end{tabular}
};

\node[tableblock, below=6mm of events] (opnodes) {
\textbf{\texttt{op\_nodes}}\\[1mm]
\begin{tabular}{|c|c|c|c|c|}
\hline
\rowcolor{gray!15}
\textbf{\texttt{op\_id}} & \textbf{\texttt{op\_name}} & \textbf{\texttt{lat}} & \textbf{\texttt{lon}} & \textbf{\texttt{...}} \\
\hline
\texttt{801} & \texttt{MARLOIE} & \texttt{50.203216} & \texttt{5.314306} & \texttt{...} \\
\texttt{788} & \texttt{MARCHE-EN-FAMENNE} & \texttt{50.222535} & \texttt{5.346178} & \texttt{...} \\
\texttt{1413} & \texttt{RACC.MARCHE-DEFENSE NATIONALE} & \texttt{50.244133} & \texttt{5.390791} & \texttt{...} \\
\texttt{818} & \texttt{MELREUX-HOTTON} & \texttt{50.283600} & \texttt{5.440353} & \texttt{...} \\
\hline
\end{tabular}
};

\node[tableblock, below=6mm of opnodes] (sections) {
\textbf{\texttt{line\_sections}}\\[1mm]
\begin{tabular}{|c|c|c|c|}
\hline
\rowcolor{gray!15}
\textbf{\texttt{line\_section\_id}} & \textbf{\texttt{line\_id}} & \textbf{\texttt{matched\_op\_node\_ids}} & \textbf{\texttt{...}} \\
\hline
\texttt{1297} & \texttt{43} & \texttt{[818, 1413, 788, 801]} & \texttt{...} \\
\hline
\end{tabular}
};

\node[tableblock, below=6mm of sections] (links) {
\textbf{\texttt{node\_links}}\\[1mm]
\begin{tabular}{|c|c|c|c|c|}
\hline
\rowcolor{gray!15}
\textbf{\texttt{link\_id}} & \textbf{\texttt{u\_node\_id}} & \textbf{\texttt{v\_node\_id}} & \textbf{\texttt{line\_section\_id}} & \textbf{\texttt{distance\_m}} \\
\hline
\texttt{354} & \texttt{788} & \texttt{801} & \texttt{1297} & \texttt{3144.4} \\
\texttt{353} & \texttt{1413} & \texttt{788} & \texttt{1297} & \texttt{4053.0} \\
\texttt{352} & \texttt{818} & \texttt{1413} & \texttt{1297} & \texttt{5684.8} \\
\hline
\end{tabular}
};

\node[tableblock, below=6mm of links] (weather) {
\textbf{\texttt{weather}}\\[1mm]
\begin{tabular}{|c|c|c|c|c|}
\hline
\rowcolor{gray!15}
\textbf{\texttt{op\_id}} & \textbf{\texttt{time}} & \textbf{\texttt{temperature\_2m}} & \textbf{\texttt{rain}} & \textbf{\texttt{...}} \\
\hline
\texttt{801} & \texttt{20:00} & \texttt{6.50} & \texttt{0.60} & \texttt{...} \\
\texttt{788} & \texttt{20:00} & \texttt{6.50} & \texttt{0.60} & \texttt{...} \\
\texttt{818} & \texttt{20:00} & \texttt{6.75} & \texttt{0.20} & \texttt{...} \\
\hline
\end{tabular}
};

\node[tableblock, below=8mm of weather] (map) {
\includegraphics[width=0.7\linewidth]{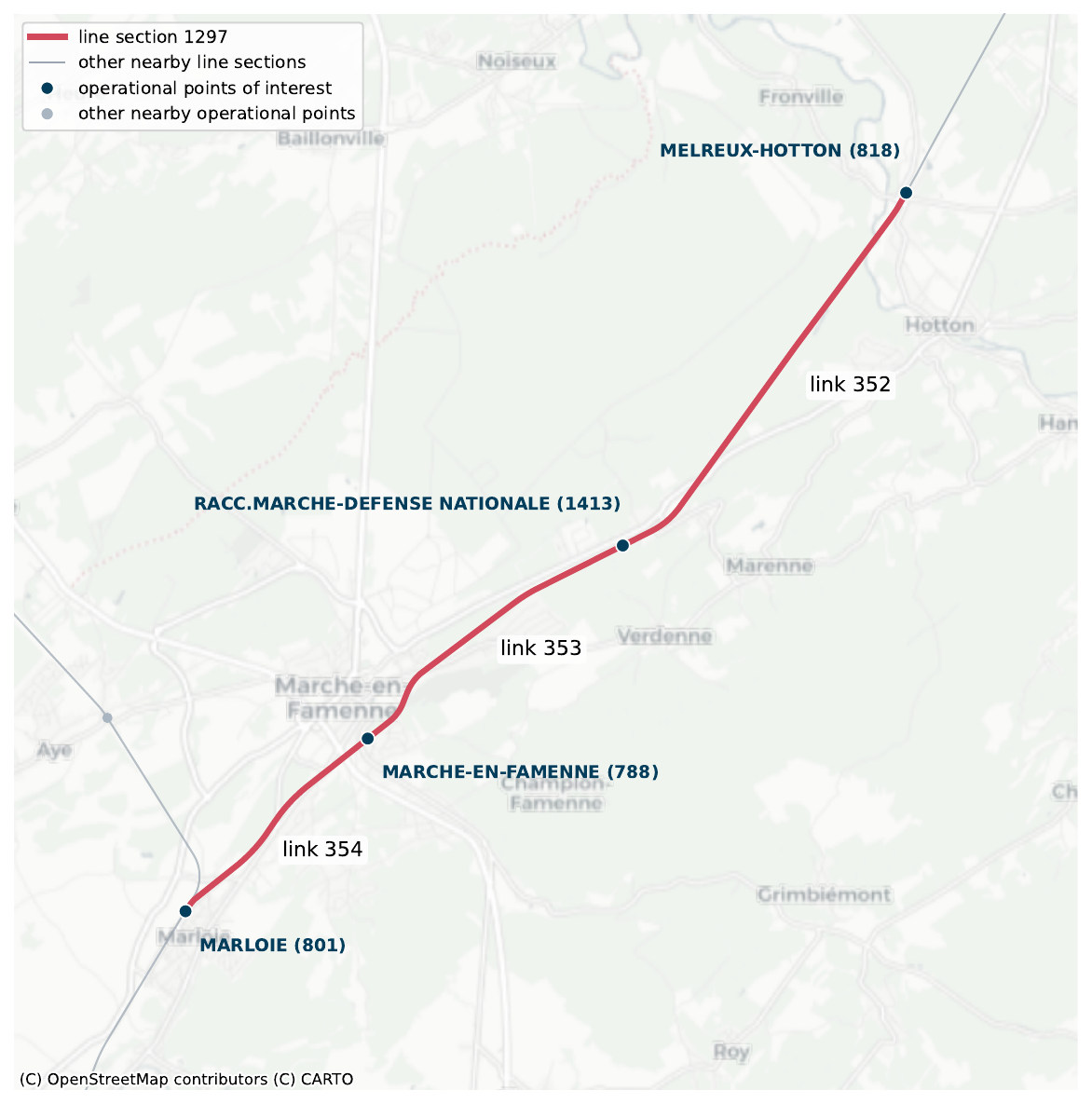}
};
\end{tikzpicture}
}
\caption{Illustrative example of silver tables for a single journey, together with a map view of the corresponding sub-network.}
\label{fig:silver_illus_example}
\end{figure}

\paragraph{Relational structure.} Table~\ref{tab:silver_relational_structure} summarizes the relational structure of the silver release. At the train-operation level, \texttt{events} and \texttt{journeys} are linked by \texttt{train\_id} and \texttt{service\_date}: events provide ordered event-level observations, while journeys aggregate them into one record per operated train service. Railway infrastructure is represented through \texttt{op\_nodes}, \texttt{line\_sections}, and \texttt{node\_links}. \texttt{op\_nodes} define the operational points referenced by \texttt{events}, \texttt{line\_sections}, \texttt{node\_links}, and \texttt{weather}. \texttt{line\_sections} store section geometries, endpoint identifiers, and the ordered operational nodes matched along each section; Figure~\ref{fig:full-network-visualization} shows the corresponding full-network visualization of \texttt{op\_nodes} and \texttt{line\_sections}. \texttt{node\_links} expand these ordered node sequences into graph edges. Journey paths are stored as ordered lists of \texttt{link\_id}s, linking journey-level records back to the railway graph. Finally, \texttt{weather} provides exogenous context that can be aligned with train-operation records through operational points and timestamps. Figure~\ref{fig:silver_illus_example} provides an illustrative example of how these tables connect around a single train operation and its associated infrastructure and weather context.

\begin{table}[H]
\centering
\small
\renewcommand{\arraystretch}{1.35}
\begin{tabular}{
p{0.17\linewidth}
>{\raggedright\arraybackslash}p{0.23\linewidth}
p{0.50\linewidth}
}
\toprule
\textbf{Table} & \textbf{Key fields} & \textbf{Description} \\
\midrule
\texttt{events}
& \texttt{train\_id}, \texttt{service\_date}, \texttt{op\_id}
& One train event at one operational point; stores event-level train movements, timestamps, event types, line context, and delays. \\

\texttt{journeys}
& \texttt{train\_id}, \texttt{service\_date}, \texttt{deduced\_paths}
& One train journey on one service date; stores journey-level metadata, numeric summaries, event counts, and inferred paths describing the exact route taken by the train through the railway graph. \\

\texttt{op\_nodes}
& \texttt{op\_id}
& One operational point; stores node identifiers, names, coordinates, and classifications. \\

\texttt{line\_sections}
& \texttt{line\_section\_id}, \texttt{op\_begin\_id}, \texttt{op\_end\_id}, \texttt{matched\_op\_node\_ids}
& One railway line section; stores infrastructure geometry, line identifiers, endpoint operational points, and matched operational nodes along the geometry. \\

\texttt{node\_links}
& \texttt{link\_id}, \texttt{u\_node\_id}, \texttt{v\_node\_id}, \texttt{line\_section\_id}
& One graph edge between consecutive operational nodes; represents railway graph connectivity and link distances. \\

\texttt{weather}
& \texttt{op\_id}, \texttt{time}
& One hourly operational-point record with temperature, rain, snowfall, humidity, wind speed, and weather code. \\
\bottomrule
\end{tabular}
\caption{Relational structure of the silver release.}
\label{tab:silver_relational_structure}
\end{table}

\paragraph{Illustrative example.}
Figure~\ref{fig:silver_illus_example} shows a small illustrative example of how the silver tables connect around one train journey with four consecutive events, together with the corresponding infrastructure and weather context.

\begin{figure}[H]
  \centering
  \includegraphics[width=0.65\textwidth]{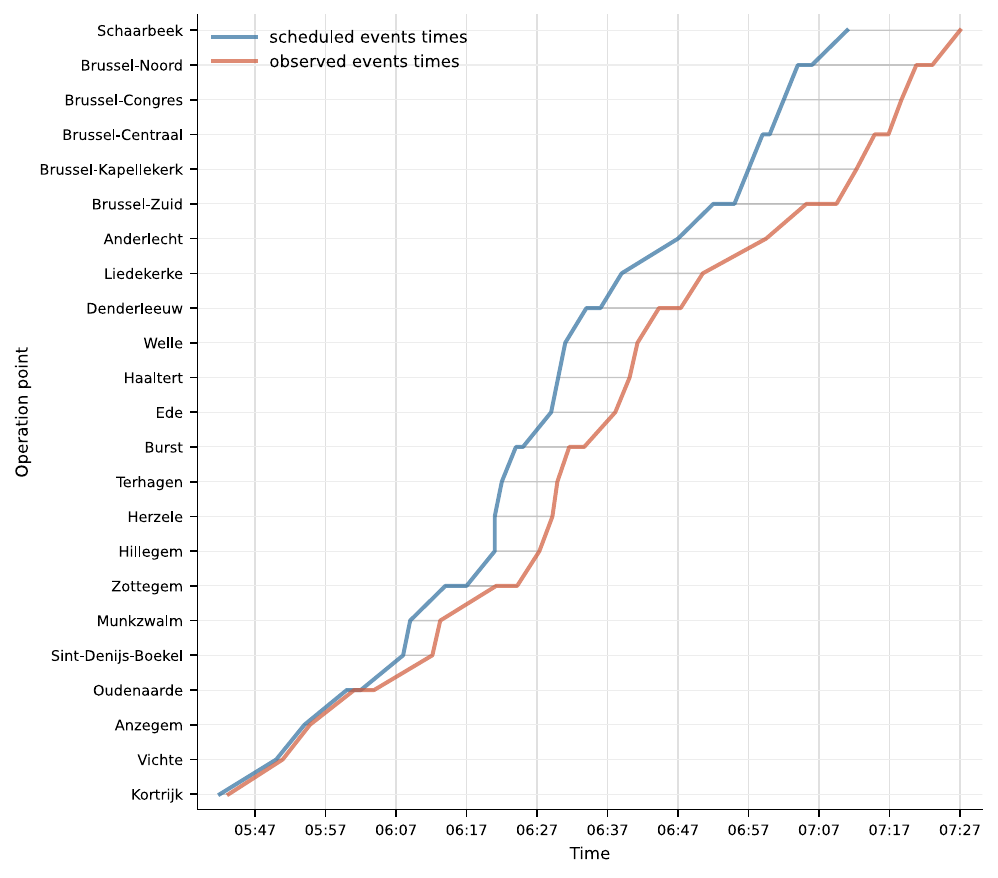}
  \caption{Visualization of the event sequence for one train on a given service date. The x-axis shows time and the y-axis lists operational points in journey order. Scheduled and observed event trajectories are plotted together. Horizontal trajectory segments at the same operational point correspond to dwell time between arrival and departure events.}
  \label{fig:journey-time-station-plot}
\end{figure}

Figure~\ref{fig:journey-time-station-plot} provides a complementary temporal view of a single silver journey, illustrating how scheduled and observed event times are represented along an ordered sequence of operational points.

\subsubsection{Silver Processing and Descriptive Statistics}

\paragraph{Processing overview.}

The silver release is constructed from the raw and bronze stages through a sequence of processing steps that transform heterogeneous source data into a coherent relational dataset. These steps combine cleaning, standardization, completion of missing or inconsistent information, and enrichment across train events, journeys, infrastructure, and weather data. In broad terms, the pipeline first consolidates operational points and infrastructure references, then reconstructs railway topology and graph connectivity, then sanitizes event timelines and aggregates them into journey-level records, and finally aligns hourly weather observations with operational points. The following paragraphs describe these components in turn.

\paragraph{Operational-node construction.}

\begin{figure}[H]
  \centering
  \includegraphics[width=0.9\textwidth]{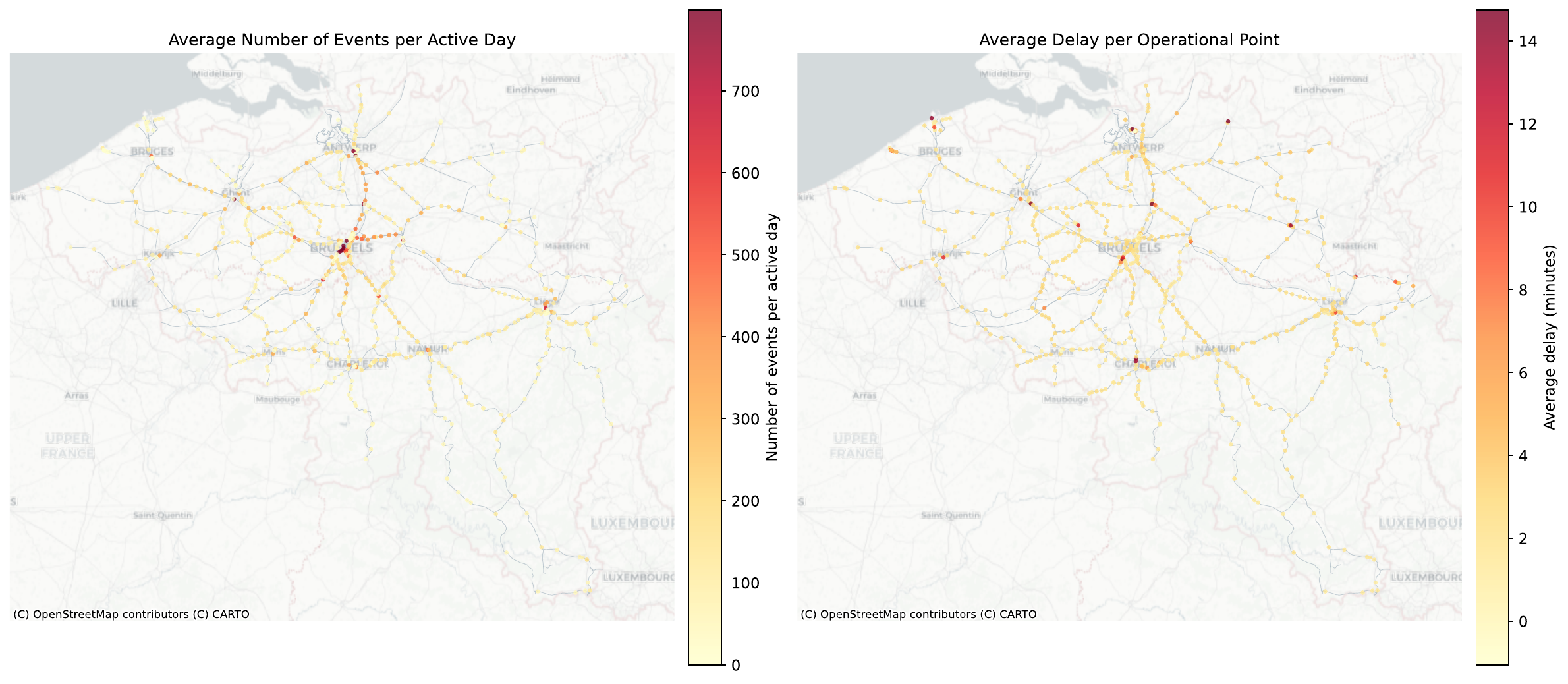}
  \caption{Descriptive statistics for the silver \texttt{op\_nodes} table. The figure maps the average number of events per active day and the average delay per operational point over the full release period.}
  \label{fig:op-nodes-descriptive-stats}
\end{figure}

The purpose of the \texttt{op\_nodes} construction step is to obtain a node catalogue that covers all operational points referenced in the bronze stage, except a small number of clearly problematic cases that do not appear in event data, and assigns them coordinates consistent with railway line sections so that a graph matching the real network can be constructed. The \texttt{op\_nodes} table is built by starting from the 1,337 operational points available in the bronze stage, among which 17 initially lacked coordinates, and then consolidating additional references required by infrastructure and event data. In particular, line-section information was used in two ways: to complete missing coordinates for 17 bronze operational points, and to add 18 additional operational points that appeared as line-section endpoints but were absent from the bronze operational-point catalogue. In addition, 4 event-only placeholder nodes were added from event files; their coordinates were then supplied manually for \texttt{op\_id} 125, 864, 2291, and 2300 by cross-referencing external web sources with the surrounding event context and line-section geometries. At the same time, 4 problematic identifiers (\texttt{1375}, \texttt{1679}, \texttt{1296}, and \texttt{1297}) were removed because they were not attached to an existing line section or the main network and did not appear in event records. No duplicate \texttt{op\_id} entries were present in the bronze input. The final silver \texttt{op\_nodes} table contains 1,355 rows, covers all operational-point references required by downstream infrastructure and event processing, and has complete coordinate information. Figure~\ref{fig:op-nodes-descriptive-stats} summarizes descriptive statistics of the final silver \texttt{op\_nodes} table through two spatial diagnostics, namely average event activity per active day and average delay per operational point.

\paragraph{Line-section construction.}

\begin{figure}[H]
  \centering
  \includegraphics[width=0.9\textwidth]{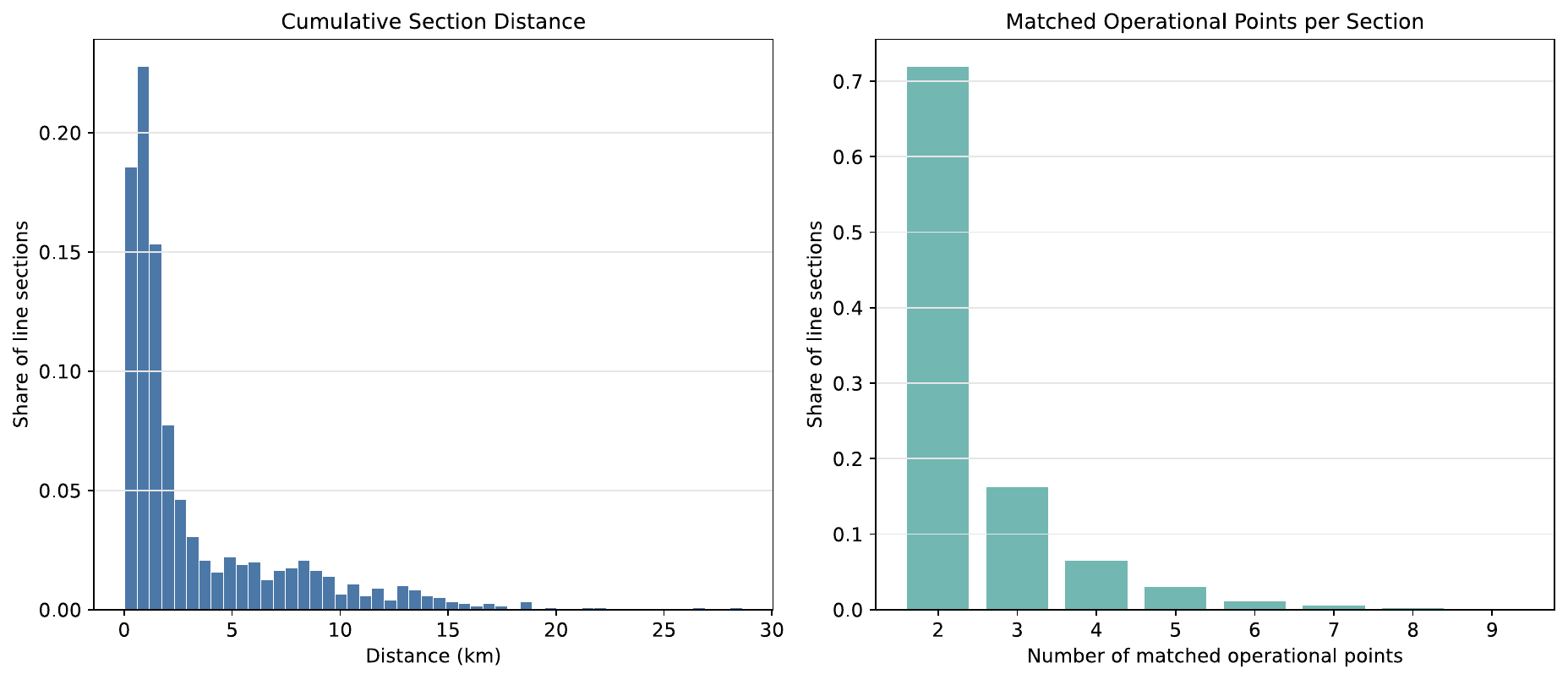}
  \caption{Descriptive statistics for the silver \texttt{line\_sections} table. The figure shows the distribution of cumulative line-section distance and the distribution of the number of matched operational points per section.}
  \label{fig:line-sections-descriptive-stats}
\end{figure}

The purpose of the \texttt{line\_sections} construction step is to transform the bronze line-section data catalogue into an operational-point-aligned representation of railway geometry that can support exact graph construction. In the bronze stage, each line section is described by its begin and end operational-point identifiers together with a \texttt{geoshape} polyline, but intermediate geometry vertices are not yet linked to operational nodes. The silver construction step enriches these geometries by aligning them to the finalized \texttt{op\_nodes} catalogue and recording the resulting ordered sequence of matched operational points in \texttt{matched\_op\_node\_ids}. Concretely, line-section endpoints are anchored to \texttt{op\_begin\_id} and \texttt{op\_end\_id}, exact coordinate matches to geometry vertices are reused when available, and additional nearby operational points are projected onto the line geometry when they fall within a 1\,m matching tolerance. For example, in the map view of Figure~\ref{fig:silver_illus_example}, the illustrated line section contains four matched operational points; in the bronze representation, only the two extreme endpoints are explicitly attached to the geometry, whereas this step recovers the two intermediate operational points and inserts them into the ordered sequence. One line section (\texttt{1165}) was removed because it is disconnected from the main network and has no linked event records. The final silver \texttt{line\_sections} table contains 1,212 rows; all line sections have at least two matched operational points, 340 sections contain at least one interior matched operational point beyond their endpoints, for a total of 585 such interior assignments, every operational point appears in at least one line section, and all operational-point references in \texttt{line\_sections} are valid. Figure~\ref{fig:full-network-visualization} provides a visualization of the resulting network, while Figure~\ref{fig:line-sections-descriptive-stats} summarizes the distributions of cumulative section distance and matched operational-point counts.

\paragraph{Node-link construction.}

\begin{figure}[H]
  \centering
  \includegraphics[width=0.9\textwidth]{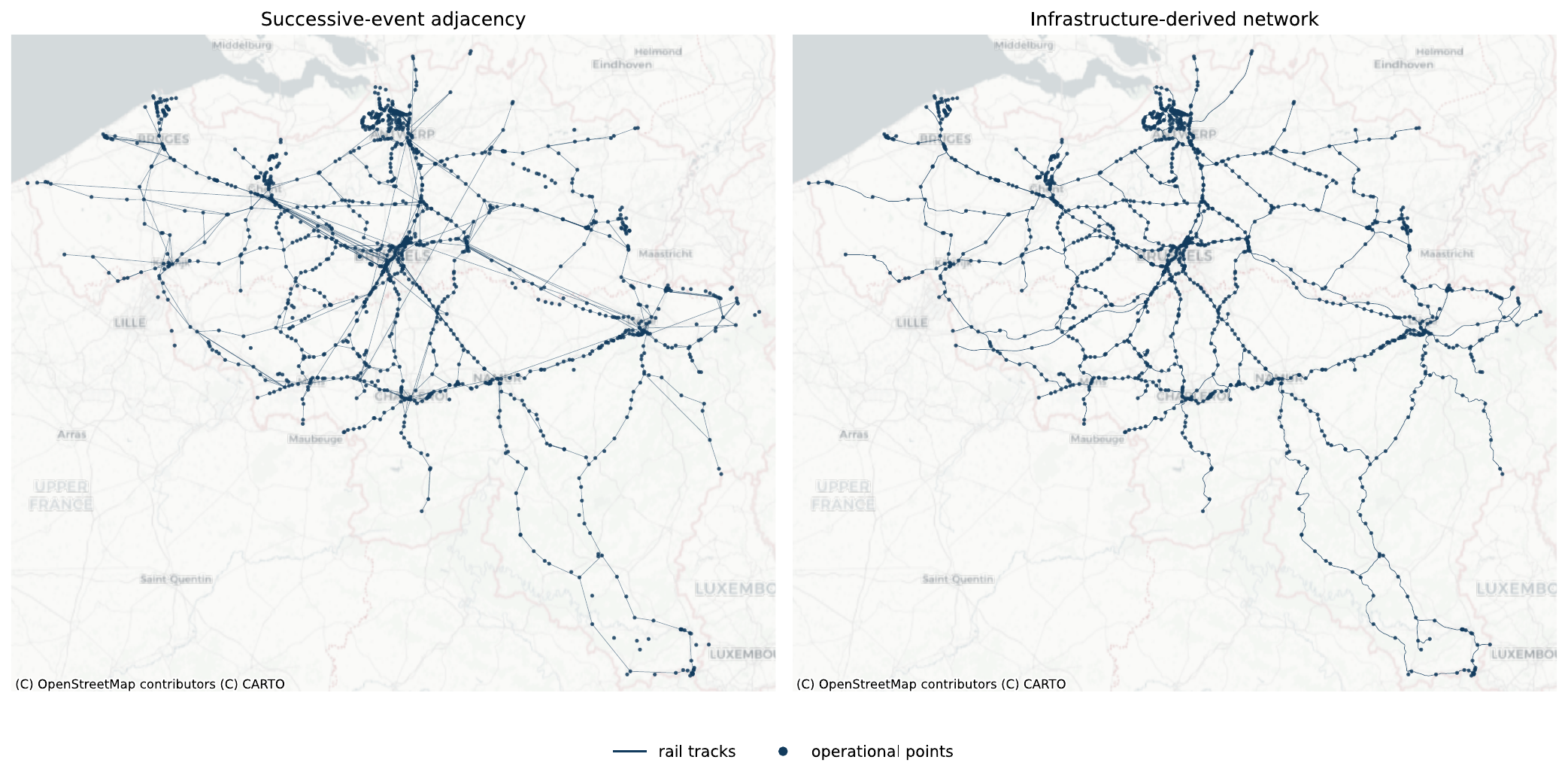}
  \caption{Comparison between an event-derived railway graph (left) and the infrastructure-derived graph used in RIDE (right).}
  \label{fig:compare-event-vs-infra-graph}
\end{figure}

\begin{figure}[H]
  \centering
  \includegraphics[width=0.9\textwidth]{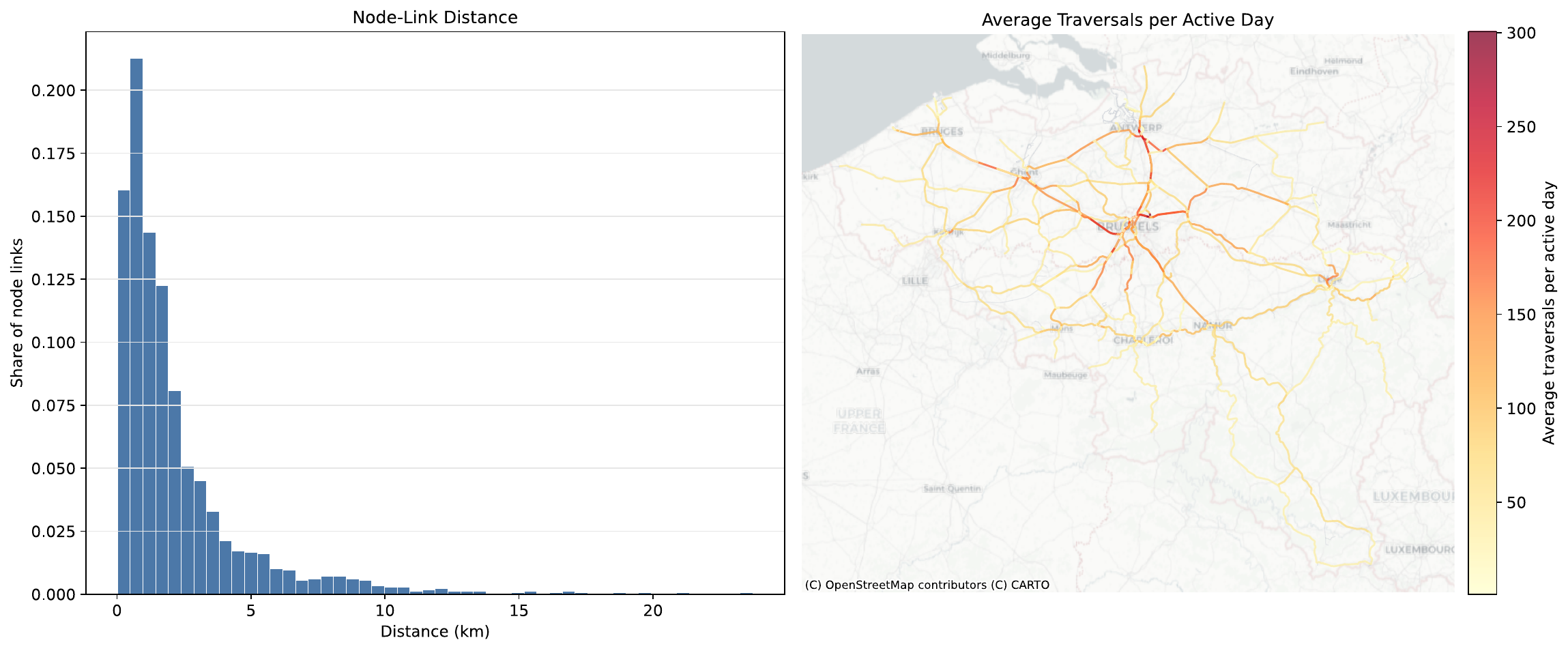}
  \caption{Descriptive statistics for the silver \texttt{node\_links} table. The figure shows the distribution of link distance and a map of average traversals per active day inferred from \texttt{deduced\_paths}. Because distinct node links can overlap visually on the same railway geometry, some map segments may appear superposed.}
  \label{fig:node-links-descriptive-stats}
\end{figure}

\begin{figure}[H]
  \centering
  \includegraphics[width=0.62\textwidth]{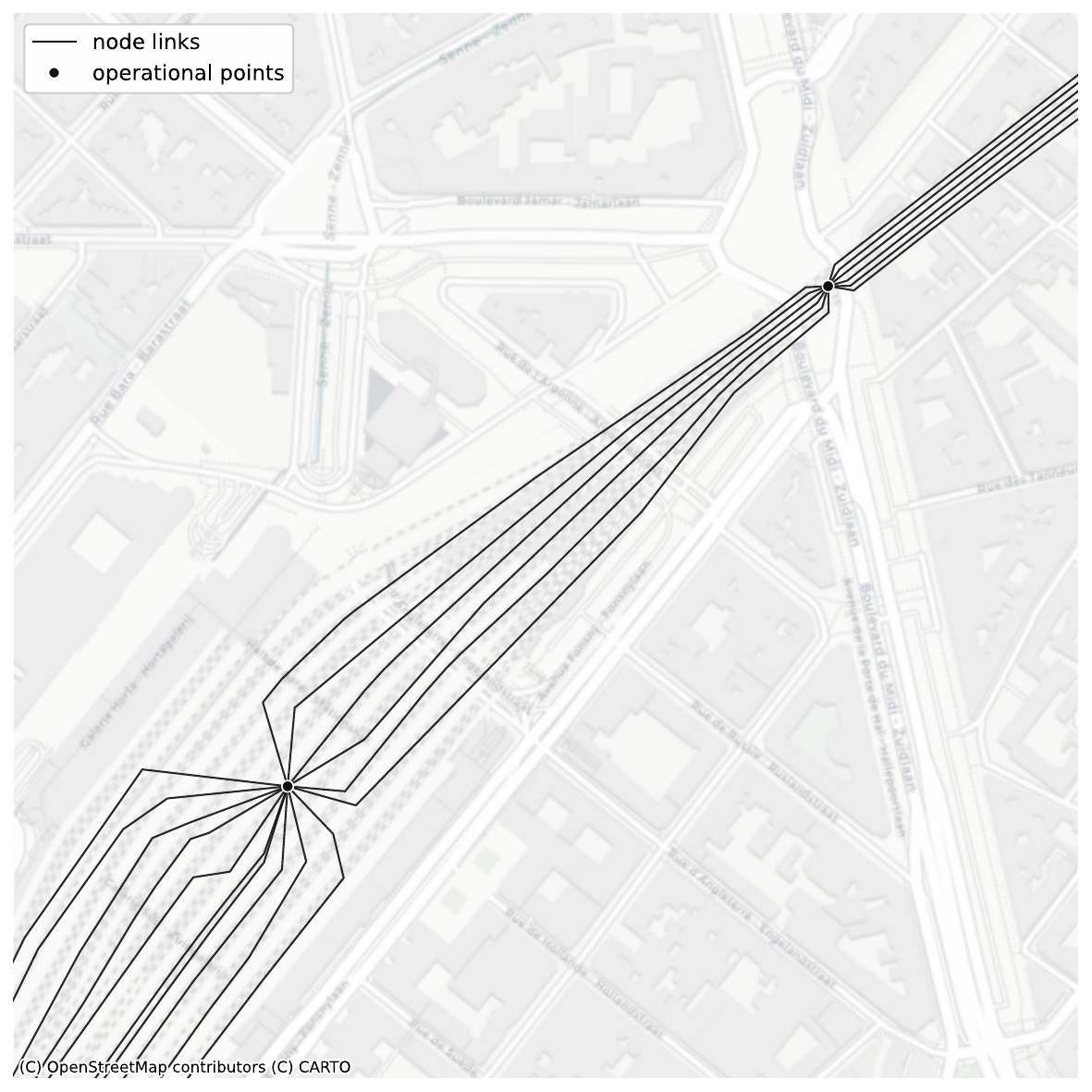}
  \caption{Local zoom on the operational-point pair with the highest node-link multiplicity. The figure illustrates how several distinct silver \texttt{node\_links} can share the same two endpoint nodes while following overlapping local railway geometries.}
  \label{fig:node-links-overlap-zoom}
\end{figure}

The purpose of the \texttt{node\_links} construction step is to derive the exact railway graph induced by the finalized \texttt{line\_sections}. For each line section, consecutive matched operational points in \texttt{matched\_op\_node\_ids} are expanded into graph edges, and the corresponding link distance is computed by summing haversine distances along the geometry segment between the two matched nodes. This produces one undirected graph edge per consecutive pair of matched operational points along each line section. This construction is important because it recovers the physical railway topology directly from infrastructure geometry, rather than approximating it from successive train events: some operational points have no event records at all, and even when an event-based adjacency can be inferred, it does not provide the actual rail geometry and the true along-track distance between nodes. Figure~\ref{fig:compare-event-vs-infra-graph} illustrates this distinction. The final silver \texttt{node\_links} table contains 1,797 rows; no link connects an operational point to itself, and all link distances are strictly positive. Because multiple line sections may connect the same pair of operational points, 121 undirected endpoint pairs appear more than once, with a maximum multiplicity of 6; Figure~\ref{fig:node-links-overlap-zoom} shows a local zoom on the most extreme such case. All 1,355 operational points appear in at least one node link, and the resulting railway graph forms a single connected component. Figure~\ref{fig:node-links-descriptive-stats} summarizes descriptive statistics of the resulting table through the link-distance distribution and a map of average traversals per active day inferred from journey paths.

\paragraph{Event cleaning.}

\begin{figure}[H]
  \centering
  \resizebox{0.85\linewidth}{!}{
  \begin{tikzpicture}[
    every node/.style={font=\small},
    casebox/.style={draw, rounded corners=2pt, fill=gray!8, inner sep=5pt, align=center},
    eventbox/.style={draw, rounded corners=2pt, fill=blue!6, inner sep=4pt, align=center, minimum width=34mm},
    cond/.style={font=\small\itshape, align=center},
    flow/.style={-{Latex[length=2mm]}, semithick}
  ]
    \node[casebox, minimum width=48mm] (bronze) at (0,0) {
      \textbf{Bronze stop row}\\[1mm]
      \begin{tabular}{@{}l@{\hspace{2mm}}l@{}}
      \texttt{planned\_arr\_ts} & \texttt{08:14:00} \\
      \texttt{planned\_dep\_ts} & \texttt{08:16:00} \\
      \texttt{observed\_arr\_ts} & \texttt{08:15:10} \\
      \texttt{observed\_dep\_ts} & \texttt{08:17:05} \\
      \end{tabular}
    };

    \node[cond] (condsplit) at (4.5,1.2) {\texttt{planned\_arr\_ts}\\$\neq$\\\texttt{planned\_dep\_ts}};
    \node[eventbox] (arr) at (8.9,2.25) {
      \textbf{Silver event row}\\[1mm]
      \begin{tabular}{@{}l@{\hspace{2mm}}l@{}}
      \texttt{event\_type} & \texttt{A} \\
      \texttt{planned\_ts} & \texttt{08:14:00} \\
      \texttt{observed\_ts} & \texttt{08:15:10} \\
      \end{tabular}
    };
    \node[eventbox] (dep) at (8.9,0.15) {
      \textbf{Silver event row}\\[1mm]
      \begin{tabular}{@{}l@{\hspace{2mm}}l@{}}
      \texttt{event\_type} & \texttt{D} \\
      \texttt{planned\_ts} & \texttt{08:16:00} \\
      \texttt{observed\_ts} & \texttt{08:17:05} \\
      \end{tabular}
    };

    \node[cond] (condpass) at (4.5,-1.7) {\texttt{planned\_arr\_ts}\\$=$\\\texttt{planned\_dep\_ts}};
    \node[casebox, minimum width=48mm] (bronzeeq) at (0,-3.15) {
      \textbf{Bronze stop row}\\[1mm]
      \begin{tabular}{@{}l@{\hspace{2mm}}l@{}}
      \texttt{planned\_arr\_ts} & \texttt{09:02:00} \\
      \texttt{planned\_dep\_ts} & \texttt{09:02:00} \\
      \texttt{observed\_arr\_ts} & \texttt{09:03:12} \\
      \texttt{observed\_dep\_ts} & \texttt{09:03:12} \\
      \end{tabular}
    };
    \node[eventbox] (pass) at (8.9,-3.15) {
      \textbf{Silver event row}\\[1mm]
      \begin{tabular}{@{}l@{\hspace{2mm}}l@{}}
      \texttt{event\_type} & \texttt{P} \\
      \texttt{planned\_ts} & \texttt{09:02:00} \\
      \texttt{observed\_ts} & \texttt{09:03:12} \\
      \end{tabular}
    };

    \draw[flow] (bronze.east) to[out=8,in=180] (condsplit.west);
    \draw[flow] (condsplit.east) to[out=0,in=180] (arr.west);
    \draw[flow] (condsplit.east) to[out=0,in=180] (dep.west);

    \draw[flow] (bronzeeq.east) to[out=0,in=180] (condpass.west);
    \draw[flow] (condpass.east) to[out=0,in=180] (pass.west);

    \node[cond, above=2mm of arr.north] {split into two silver event rows};
    \node[cond, above=2mm of pass.north] {collapse into one silver event row};
  \end{tikzpicture}
  }
  \caption{Illustration of event-type separation from bronze stop records into silver event rows.}
  \label{fig:event-type-separation}
\end{figure}
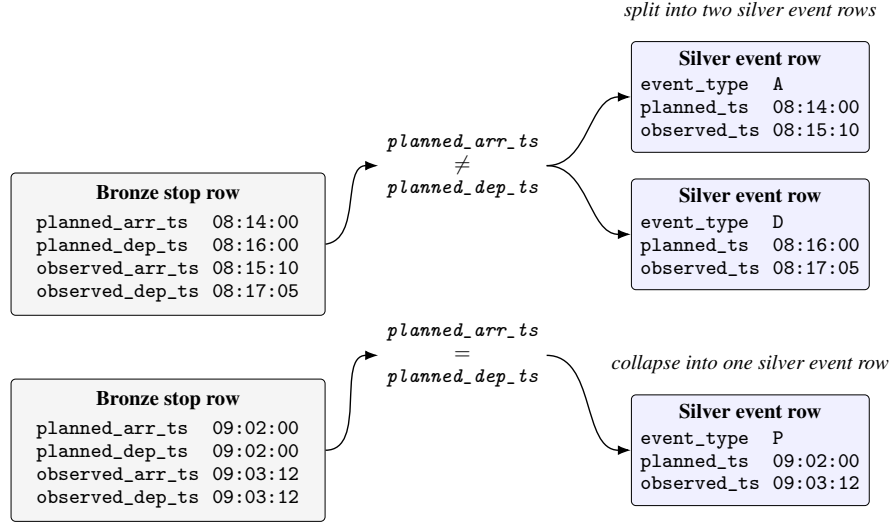

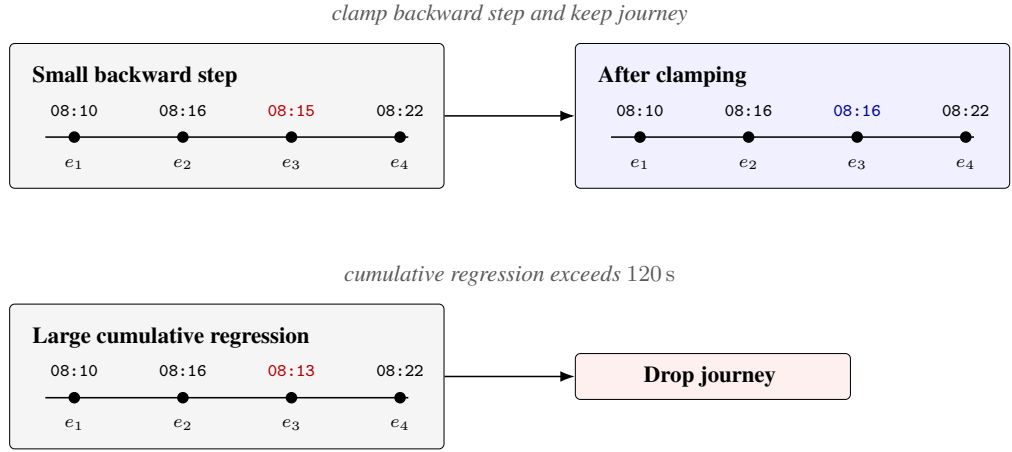
\begin{figure}[H]
  \centering
  \resizebox{0.96\linewidth}{!}{
  \begin{tikzpicture}[
    every node/.style={font=\small},
    timeline/.style={draw, rounded corners=2pt, fill=gray!8, inner sep=5pt},
    goodtimeline/.style={draw, rounded corners=2pt, fill=blue!6, inner sep=5pt},
    badbox/.style={draw, rounded corners=2pt, fill=red!6, inner sep=5pt, align=center},
    ann/.style={font=\small\itshape, text=black!65, align=center},
    flow/.style={-{Latex[length=2mm]}, semithick},
    eventdot/.style={circle, fill=black, inner sep=1.6pt}
  ]
    \node[timeline, minimum width=60mm, minimum height=20mm, anchor=north west] (smallcase) at (0,0) {};
    \node[anchor=north west, font=\small\bfseries] at ($(smallcase.north west)+(2mm,-2mm)$) {Small backward step};
    \draw[semithick] ($(smallcase.north west)+(5mm,-13mm)$) -- ($(smallcase.north east)+(-5mm,-13mm)$);
    \foreach \x/\lab in {9/1,24/2,39/3,54/4} {
      \node[eventdot] at ($(smallcase.north west)+(\x mm,-13mm)$) {};
      \node[font=\scriptsize, anchor=north] at ($(smallcase.north west)+(\x mm,-14.8mm)$) {$e_{\lab}$};
    }
    \node[font=\scriptsize, anchor=south] at ($(smallcase.north west)+(9mm,-11.3mm)$) {\texttt{08:10}};
    \node[font=\scriptsize, anchor=south] at ($(smallcase.north west)+(24mm,-11.3mm)$) {\texttt{08:16}};
    \node[font=\scriptsize, anchor=south, text=red!70!black] at ($(smallcase.north west)+(39mm,-11.3mm)$) {\texttt{08:15}};
    \node[font=\scriptsize, anchor=south] at ($(smallcase.north west)+(54mm,-11.3mm)$) {\texttt{08:22}};

    \node[goodtimeline, minimum width=60mm, minimum height=20mm, anchor=west] (smallfixed) at ($(smallcase.east)+(18mm,0)$) {};
    \node[anchor=north west, font=\small\bfseries] at ($(smallfixed.north west)+(2mm,-2mm)$) {After clamping};
    \draw[semithick] ($(smallfixed.north west)+(5mm,-13mm)$) -- ($(smallfixed.north east)+(-5mm,-13mm)$);
    \foreach \x/\lab in {9/1,24/2,39/3,54/4} {
      \node[eventdot] at ($(smallfixed.north west)+(\x mm,-13mm)$) {};
      \node[font=\scriptsize, anchor=north] at ($(smallfixed.north west)+(\x mm,-14.8mm)$) {$e_{\lab}$};
    }
    \node[font=\scriptsize, anchor=south] at ($(smallfixed.north west)+(9mm,-11.3mm)$) {\texttt{08:10}};
    \node[font=\scriptsize, anchor=south] at ($(smallfixed.north west)+(24mm,-11.3mm)$) {\texttt{08:16}};
    \node[font=\scriptsize, anchor=south, text=blue!60!black] at ($(smallfixed.north west)+(39mm,-11.3mm)$) {\texttt{08:16}};
    \node[font=\scriptsize, anchor=south] at ($(smallfixed.north west)+(54mm,-11.3mm)$) {\texttt{08:22}};

    \draw[flow] (smallcase.east) -- (smallfixed.west);
    \node[ann] at ($(smallcase.east)!0.5!(smallfixed.west)+(0,14mm)$) {clamp backward step and keep journey};

    \node[timeline, minimum width=60mm, minimum height=20mm, anchor=north west] (bigcase) at (0,-3.6) {};
    \node[anchor=north west, font=\small\bfseries] at ($(bigcase.north west)+(2mm,-2mm)$) {Large cumulative regression};
    \draw[semithick] ($(bigcase.north west)+(5mm,-13mm)$) -- ($(bigcase.north east)+(-5mm,-13mm)$);
    \foreach \x/\lab in {9/1,24/2,39/3,54/4} {
      \node[eventdot] at ($(bigcase.north west)+(\x mm,-13mm)$) {};
      \node[font=\scriptsize, anchor=north] at ($(bigcase.north west)+(\x mm,-14.8mm)$) {$e_{\lab}$};
    }
    \node[font=\scriptsize, anchor=south] at ($(bigcase.north west)+(9mm,-11.3mm)$) {\texttt{08:10}};
    \node[font=\scriptsize, anchor=south] at ($(bigcase.north west)+(24mm,-11.3mm)$) {\texttt{08:16}};
    \node[font=\scriptsize, anchor=south, text=red!70!black] at ($(bigcase.north west)+(39mm,-11.3mm)$) {\texttt{08:13}};
    \node[font=\scriptsize, anchor=south] at ($(bigcase.north west)+(54mm,-11.3mm)$) {\texttt{08:22}};

    \node[badbox, minimum width=38mm, anchor=west] (dropjourney) at ($(bigcase.east)+(18mm,0)$) {
      \textbf{Drop journey}
    };
    \draw[flow] (bigcase.east) -- (dropjourney.west|-bigcase.east);
    \node[ann] at ($(bigcase.east)!0.5!(dropjourney.west|-bigcase.east)+(0,14mm)$) {cumulative regression exceeds $120\,\mathrm{s}$};
  \end{tikzpicture}
  }
  \caption{Illustration of local consistency enforcement during event cleaning.}
  \label{fig:monotonicity-enforcement}
\end{figure}

\begin{figure}[H]
  \centering
  \includegraphics[width=0.9\textwidth]{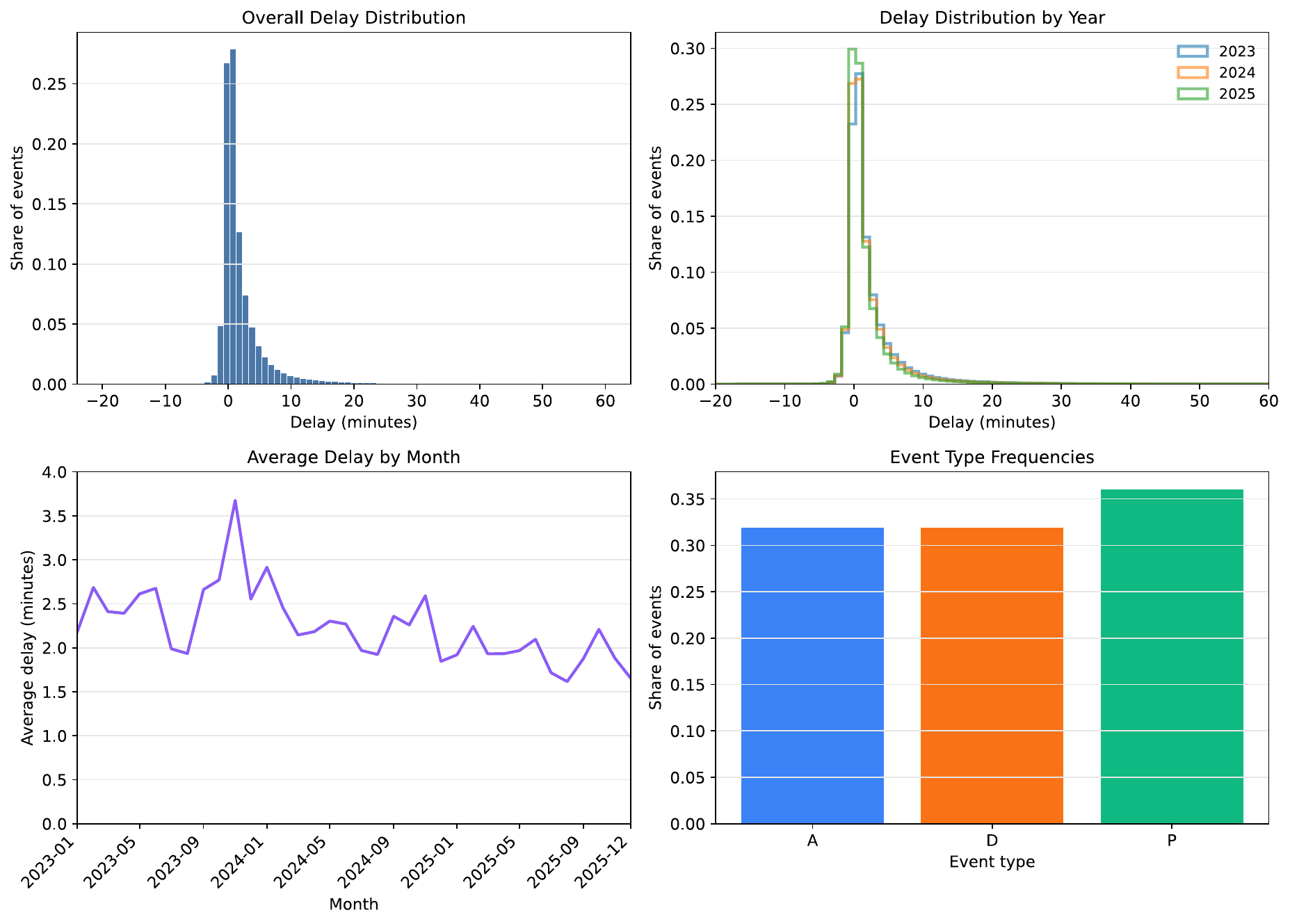}
  \caption{Descriptive statistics for the silver \texttt{events} table. The figure shows the overall delay distribution, the delay distribution by year, the average delay by month, and event-type frequencies.}
  \label{fig:events-descriptive-stats}
\end{figure}

The purpose of the event-cleaning step is to transform the bronze train movement records into journey-level event sequences that remain faithful to the original event order while satisfying the temporal consistency expected from a clean journey representation. A central assumption of this step is that the original row order in the raw files already reflects the correct physical progression of the train along its journey; we therefore preserve this order throughout cleaning and apply all repairs in place rather than reordering events. In our inspection of the raw data, this assumption was found to hold overwhelmingly well, including for journeys whose timestamps exhibit non-monotonicity. In the bronze data, each row corresponds to one train-at-operational-point record and may contain both arrival and departure information. Event-type separation converts such records into silver event rows: when scheduled arrival and departure times differ, the row is split into one arrival event and one departure event, whereas when they are equal, it is represented as a single passage event. Figure~\ref{fig:event-type-separation} illustrates these two cases. The resulting silver table therefore contains one row per arrival, departure, or passage event, with one scheduled timestamp, one observed timestamp, and event timelines that are non-decreasing in both scheduled and observed time within each journey. The cleaning pipeline first reconstructs timestamps from the raw date and time fields, then resolves local arrival/departure conflicts within the same operational point. Concretely, a conflict occurs when a scheduled departure precedes a scheduled arrival within the same bronze row. Using a tolerance of 120\,s, such cases are repaired in place when the discrepancy remains small by setting the arrival timestamp equal to the departure timestamp, which subsequently yields a single passage event after event-type separation; otherwise the whole journey is discarded. Planned and observed event timelines are then enforced separately with the same 120\,s tolerance: small backward steps are clamped in place, whereas journeys whose cumulative regressions exceed the threshold are removed. Figure~\ref{fig:monotonicity-enforcement} illustrates this repair-versus-drop logic on a simplified event sequence; the same illustration is also valid for local arrival/departure conflicts within a bronze stop record. Delays are then computed explicitly as the difference in seconds between observed and scheduled timestamps, and journeys are filtered if any event delay exceeds 4\,h late or 1\,h early, so as to exclude rare extreme outliers that are unlikely to be informative for the benchmark task. Finally, duplicate events with the same \texttt{train\_id}, \texttt{service\_date}, \texttt{event\_type}, and \texttt{op\_id} are removed. Table~\ref{tab:event-cleaning-stats} summarizes the main journey-level removals and row-level adjustments introduced by this event-cleaning step. Figure~\ref{fig:events-descriptive-stats} summarizes descriptive statistics of the final silver \texttt{events} table, including delay distributions, temporal variation in average delay, and event-type frequencies.

\begin{table}[H]
\centering
\normalsize
\renewcommand{\arraystretch}{1.15}
\begin{tabular}{llr}
\toprule
\textbf{Operation} & \textbf{Count} & \textbf{Share} \\
\midrule
\multicolumn{3}{l}{\textit{Journey-level removals (relative to 3,625,094 bronze journeys)}} \\
Dropped due to local arrival/departure conflicts & 207 & 0.006\% \\
Dropped after planned monotonicity checks & 1,821 & 0.050\% \\
Dropped after observed monotonicity checks & 12,063 & 0.333\% \\
Dropped by delay bounds & 130 & 0.004\% \\
Total removed journeys & 14,221 & 0.392\% \\
\midrule
\multicolumn{3}{l}{\textit{Event-level adjustments/removals (relative to 94,492,813 final silver events)}} \\
Arrival/departure pairs collapsed into passage events & 6,826 & 0.007\% \\
Planned timestamps modified for monotonicity & 6,767 & 0.007\% \\
Observed timestamps modified for monotonicity & 32,115 & 0.034\% \\
Duplicate event rows removed & 442 & 0.0005\% \\
Total event-level adjustments/removals & 46,150 & 0.049\% \\
\bottomrule
\end{tabular}
\caption{Summary statistics for the event-cleaning step.}
\label{tab:event-cleaning-stats}
\end{table}

\begin{figure}[H]
  \centering
  \resizebox{0.96\linewidth}{!}{
  \begin{tikzpicture}[
    every node/.style={font=\small},
    eventbox/.style={draw, rounded corners=2pt, fill=gray!8, inner sep=5pt, align=left},
    netnode/.style={circle, draw, fill=white, minimum size=7mm, inner sep=0pt},
    chosen/.style={draw=blue!70!black, line width=1.1pt},
    alt/.style={draw=black!55, line width=0.9pt},
    faint/.style={draw=black!25, line width=0.8pt},
    ann/.style={font=\small\itshape, text=black!65, align=center},
    flow/.style={-{Latex[length=2mm]}, semithick}
  ]
    \node[netnode] (s) at (1.6,0) {\texttt{100}};
    \node[netnode] (a) at (4.5,1.1) {\texttt{212}};
    \node[netnode] (b) at (4.5,-1.1) {\texttt{320}};
    \node[netnode] (t) at (7.7,0) {\texttt{400}};
    \coordinate (graphcenter) at ($(s)!0.5!(t)$);

    \node[eventbox, anchor=south east] (evsrc) at ($(s)+(-10mm,12mm)$) {
      \textbf{Event $i$}\\
      \texttt{op\_id = 100}\\
      \texttt{dep\_line\_id = L1}
    };
    \node[eventbox, anchor=south west] (evdst) at ($(t)+(10mm,12mm)$) {
      \textbf{Event $i+1$}\\
      \texttt{op\_id = 400}\\
      \texttt{arr\_line\_id = L2}
    };

    \draw[faint] (a) -- node[pos=0.6, above,  font=\scriptsize] {\texttt{L1}, $500\,\mathrm{m}$, $\times 0.65$} (b);
    \draw[chosen] (s) -- node[pos=0.56, above, xshift=-20pt, font=\scriptsize, text=blue!70!black] {\texttt{L1}, $900\,\mathrm{m}$, $\times 0.55$} (a);
    \draw[chosen] (a) -- node[pos=0.44, above, xshift=20pt, font=\scriptsize, text=blue!70!black] {\texttt{L2}, $900\,\mathrm{m}$, $\times 0.55$} (t);
    \draw[alt] (s) -- node[pos=0.56, below, xshift=-20pt, font=\scriptsize] {\texttt{L3}, $700\,\mathrm{m}$, $\times 1.0$} (b);
    \draw[alt] (b) -- node[pos=0.44, below, xshift=20pt, font=\scriptsize] {\texttt{L1}, $700\,\mathrm{m}$, $\times 0.75$} (t);

    \draw[flow] (evsrc.south east) to[out=-28,in=140] ($(s)+(0,0.42)$);
    \draw[flow] (evdst.south west) to[out=-152,in=40] ($(t)+(0,0.42)$);

    \node[ann, anchor=south] at ($(graphcenter)+(0,16mm)$) {biased Dijkstra prefers paths consistent with\\the departure and arrival line identifiers};
    \node[ann, text=blue!70!black, anchor=north] at ($(graphcenter)+(0,-16mm)$) {chosen path (100-212-400): physical length $= 1800\,\mathrm{m}$, effective cost $= 900\times0.55 + 900\times0.55 = 990$};
    \node[ann, anchor=north] at ($(graphcenter)+(0,-24mm)$) {alternative path (100-320-400): physical length $= 1400\,\mathrm{m}$, effective cost $= 700\times1.0 + 700\times0.75 = 1225$};
    \node[ann, anchor=north] at ($(graphcenter)+(0,-32mm)$) {\texttt{deduced\_paths} therefore stores the blue route, despite a shorter physical alternative};
  \end{tikzpicture}
  }
\caption{Illustrative example of path inference between two consecutive events. Edge labels indicate the line identifier, physical distance, and multiplicative line-preference coefficient used in path inference.}
\label{fig:journey-path-inference}
\end{figure}
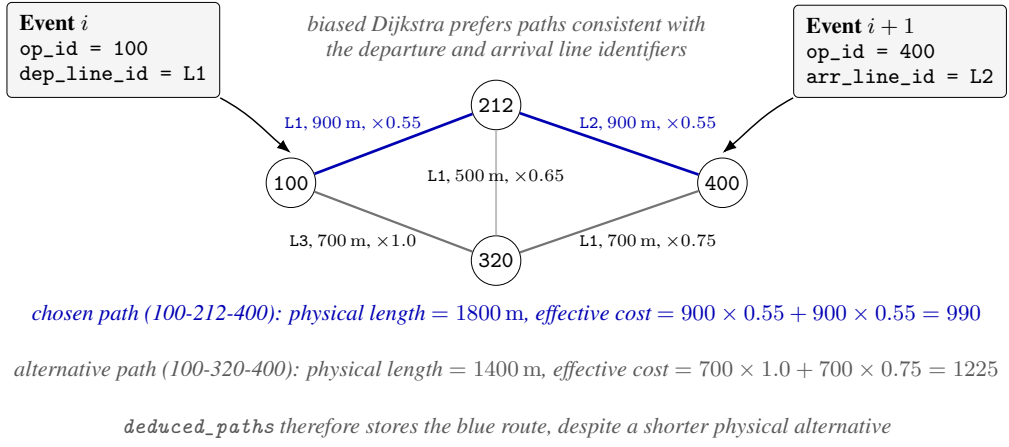

\begin{figure}[H]
  \centering
  \includegraphics[width=0.9\textwidth]{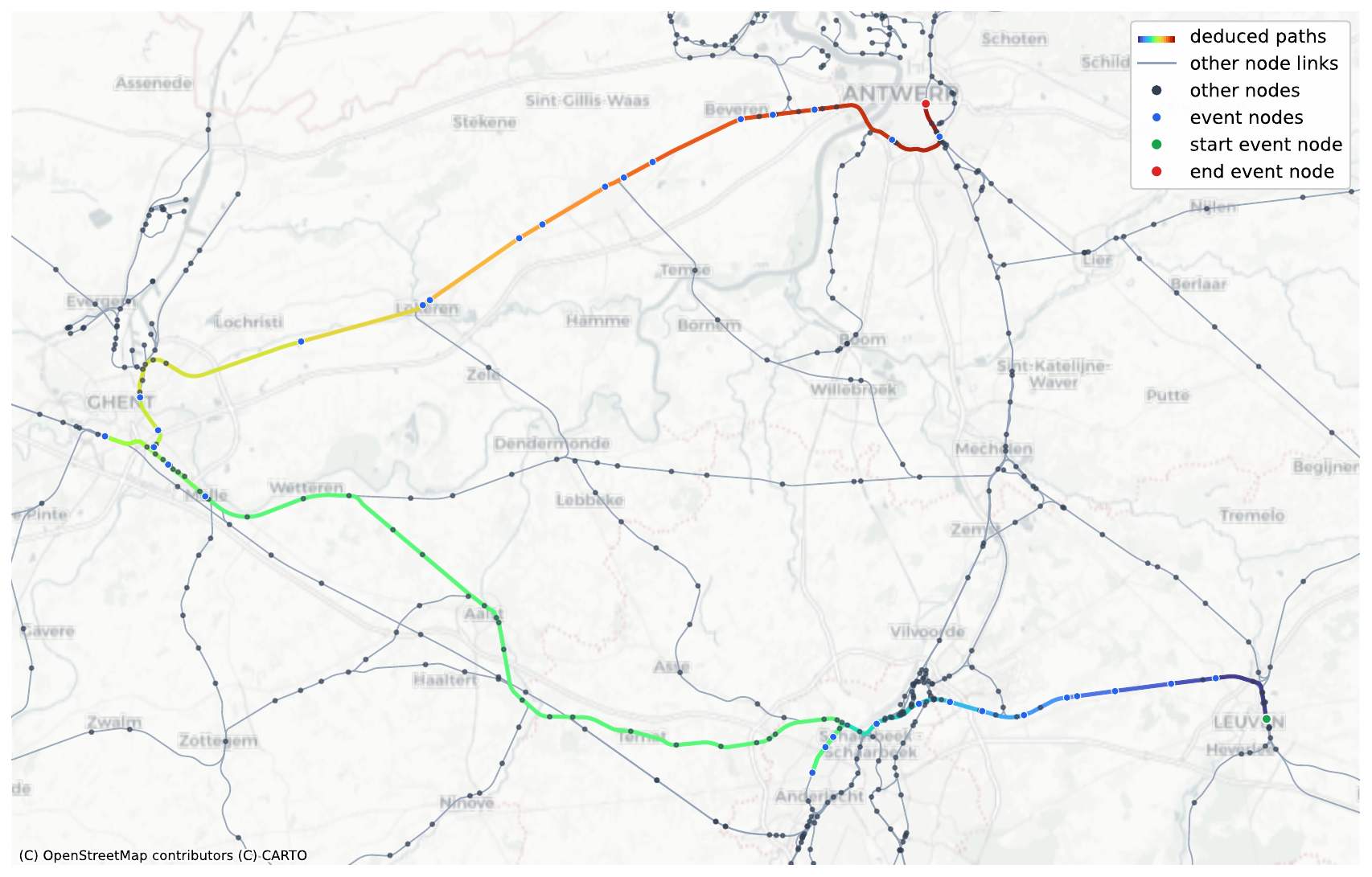}
\caption{Illustrative example of a journey-level inferred path. Each color corresponds to one inferred subpath between two consecutive event locations. The color therefore remains unchanged when the train passes through intermediate operational points without generating an event, and changes only when the next event location is reached. Blue nodes denote event locations from the silver \texttt{events} table, while darker nodes denote other operational points in the surrounding network. Gray lines show the remaining local railway network; the green and red nodes mark the first and last events of the journey.}
  \label{fig:journey-illustration}
\end{figure}

\begin{figure}[H]
  \centering
  \includegraphics[width=\textwidth]{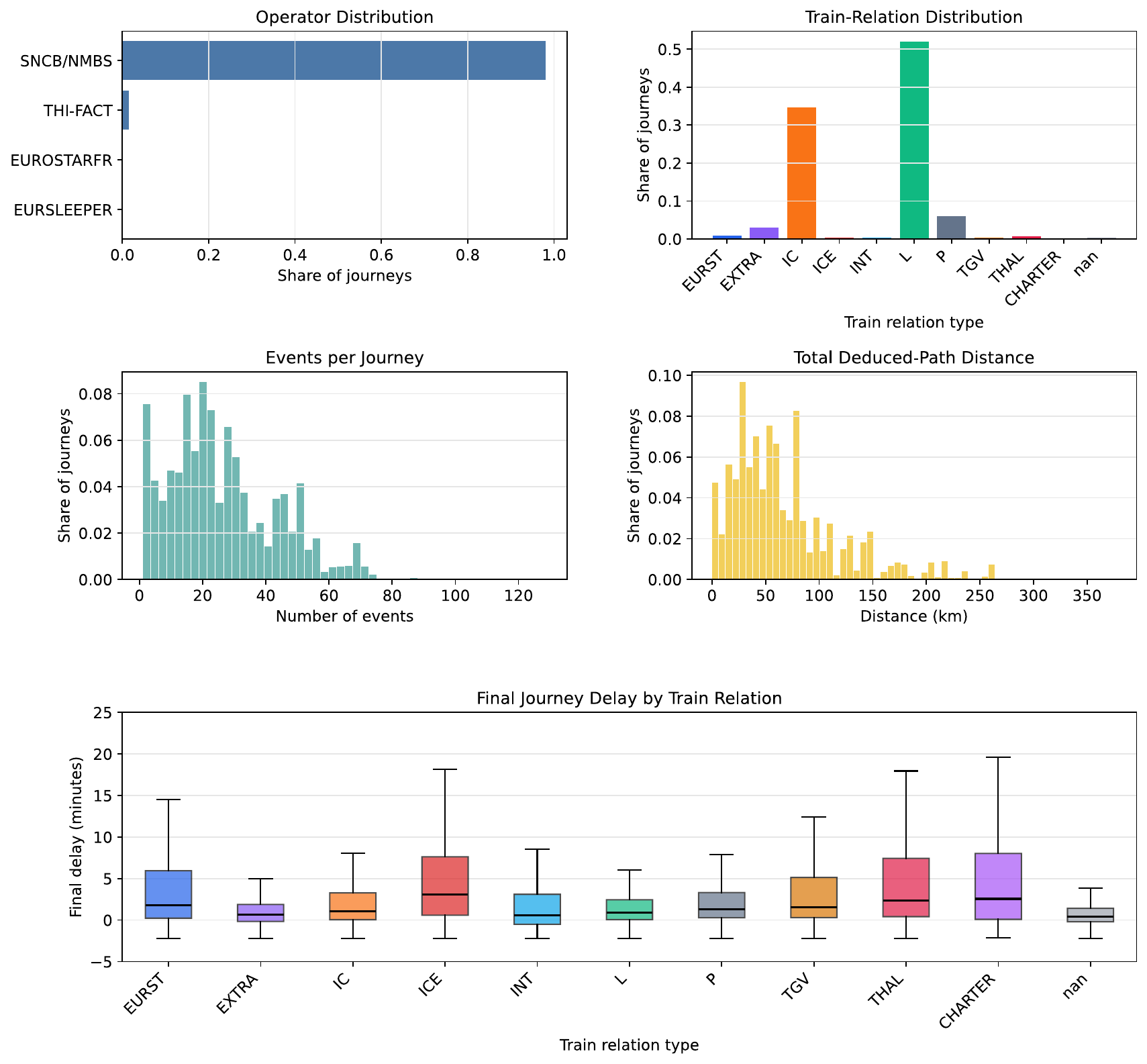}
  \caption{Descriptive statistics for the silver \texttt{journeys} table. The figure shows operator frequencies, train-relation frequencies, the distribution of event counts, the distribution of total deduced-path distance, and final journey delay by train-relation category.}
  \label{fig:journeys-descriptive-stats}
\end{figure}

\paragraph{Journey construction.}
The purpose of the \texttt{journeys} construction step is to provide one high-level descriptor per train and service date, aggregating information that would otherwise be repeated across the event table. Starting from the cleaned silver \texttt{events} table, the pipeline constructs one row per \texttt{(train\_id, service\_date)} pair and restores journey-level metadata from the corresponding bronze records, including \texttt{train\_relation}, \texttt{operator}, and \texttt{relation\_direction}. It then aggregates journey-level summary fields from the event sequence, namely the start and end operational points, the start and end planned and observed timestamps, the total number of events, and the maximum and minimum delay observed along the journey. Because many downstream methods may benefit from, or directly require, the exact edge sequence used by a train, the \texttt{journeys} table also stores \texttt{deduced\_paths}, which records for each consecutive pair of events the sequence of \texttt{node\_links} traversed between their operational points. Figure~\ref{fig:journey-illustration} shows an illustrative example of one such journey, where the inferred subpaths between consecutive events may pass through additional operational points without generating new event records. Inferring these paths is not trivial. Consecutive events in the cleaned event sequence provide only an ordered list of operational points, and successive operational points are not necessarily adjacent in the railway graph. Recovering the traversed segment therefore requires resolving a path through the network between each consecutive pair of event locations. This is further complicated by the fact that multiple feasible routes may exist between the same two operational points, especially when parallel or branching line segments are available locally. To resolve this ambiguity, path inference uses two additional sources of information. First, each split silver event retains its arrival and departure line identifiers, so that for a consecutive pair of events we know the line context at both ends of the segment. Second, each \texttt{node\_link} is associated with a \texttt{line\_section}, and therefore with a railway \texttt{line\_id}, as well as a physical along-track distance. We therefore construct the full silver railway graph from \texttt{node\_links} and resolve each consecutive event pair with a biased Dijkstra search \cite{dijkstra2022note} whose edge costs are the physical link distances multiplied by line-preference coefficients. For a consecutive event pair, let $\ell_{\mathrm{src}}$ denote the departure line identifier of the first event and $\ell_{\mathrm{dst}}$ the arrival line identifier of the second event. The biased Dijkstra search then multiplies each edge length by a line-preference coefficient. Edges incident to the source node prefer $\ell_{\mathrm{src}}$ first and $\ell_{\mathrm{dst}}$ second; edges incident to the destination node prefer $\ell_{\mathrm{dst}}$ first and $\ell_{\mathrm{src}}$ second; and intermediate edges receive a preference when their line identifier matches either $\ell_{\mathrm{src}}$ or $\ell_{\mathrm{dst}}$. Concretely, the multiplicative coefficients are 0.55 for the strongest source/destination preference, 0.75 for the secondary source/destination preference, 0.65 for intermediate matches, and 1.0 otherwise. The coefficient values were selected empirically through manual inspection of resolved paths on representative cases, with the goal of favoring line-consistent routes without overwhelming the underlying physical-distance criterion. Figure~\ref{fig:journey-path-inference} illustrates how this bias can favor a slightly longer physical route when it is more consistent with the line identifiers observed at the two ends of the segment. Figure~\ref{fig:journeys-descriptive-stats} summarizes descriptive statistics of the final silver \texttt{journeys} table, including operator and train-relation frequencies, journey size, deduced-path distance, and final-delay behavior across train-relation categories.

\begin{figure}[H]
  \centering
  \includegraphics[width=\textwidth]{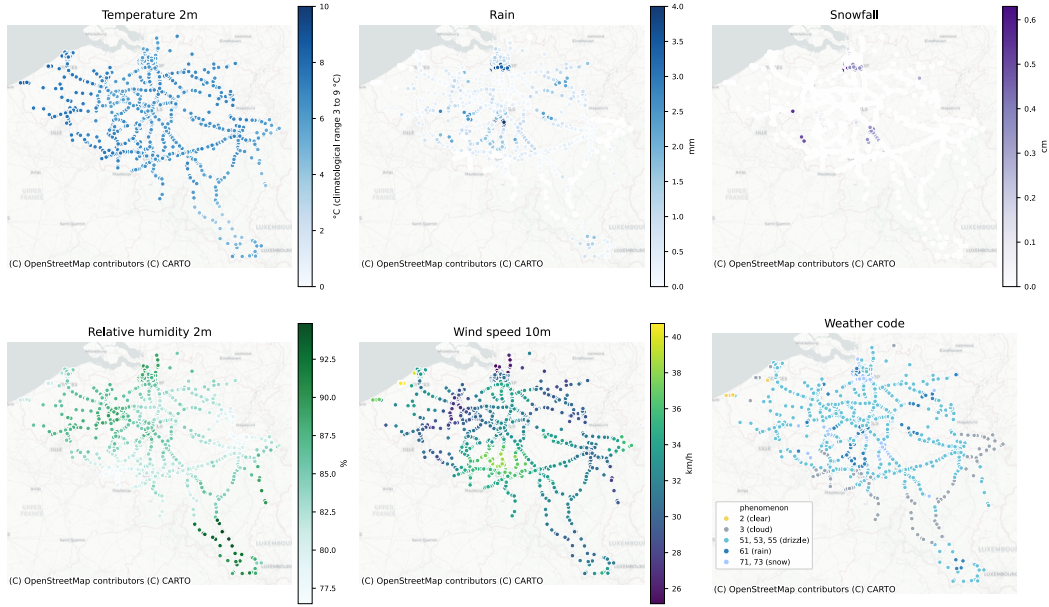}
  \caption{Illustrative weather snapshot from the silver \texttt{weather} table, showing the six aligned weather variables for all operational points at a single timestamp.}
  \label{fig:silver-weather-snapshot}
\end{figure}

\begin{figure}[H]
  \centering
  \includegraphics[width=\textwidth]{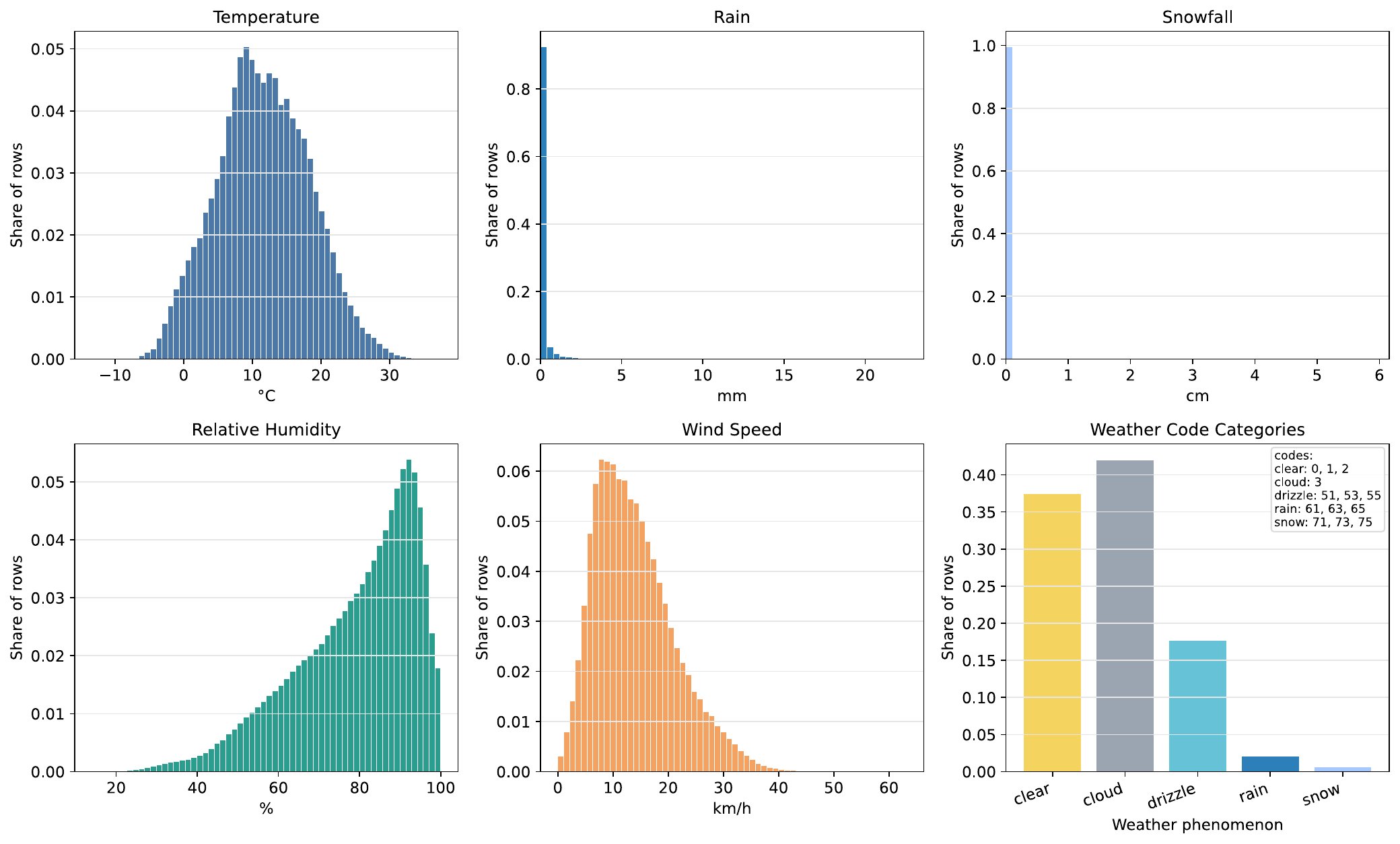}
  \caption{Descriptive statistics for the silver \texttt{weather} table. The figure shows the distributions of temperature, rain, snowfall, relative humidity, and wind speed, together with grouped frequencies of the observed \texttt{weather\_code} categories.}
  \label{fig:weather-descriptive-stats}
\end{figure}

\paragraph{Weather construction.}
The purpose of the \texttt{weather} construction step is to provide, for every operational point and every hour from 2022-12-31 to 2026-01-01, covering the 2023--2025 period with one-day padding on both sides, a synchronized set of local weather features that can be aligned with train events and prediction snapshots. Unlike the other silver tables, this step is not executed directly within the main silver pipeline: the raw weather data must first be downloaded from the Open-Meteo archive API after the silver \texttt{op\_nodes} table has been finalized, since weather extraction requires the final operational-point coordinates. The raw extracts are then concatenated into a single table containing the six retained variables, namely \texttt{temperature\_2m}, \texttt{rain}, \texttt{snowfall}, \texttt{relative\_humidity\_2m}, \texttt{wind\_speed\_10m}, and \texttt{weather\_code}.
To obtain a temporally regular representation, the concatenated records are sorted by \texttt{op\_id} and time, converted to hourly timestamps in Europe/Brussels time, and reindexed independently for each operational point onto a complete hourly grid, with forward filling applied if gaps are present. In the final dataset, this yields 35,706,960 rows, corresponding to 1,355 operational points with a complete hourly series of 26,352 timestamps each. In practice, the downloaded batches already form a complete hourly grid, so reindexing and forward filling act here as validation-preserving safeguards rather than substantive repairs: no rows are added, no duplicate \texttt{(op\_id, time)} pairs are present, all 1,355 operational points are covered, all six weather variables are fully observed, and every operational point has a strictly hourly time series. Figure~\ref{fig:silver-weather-snapshot} shows an illustrative snapshot from the resulting table, while Figure~\ref{fig:weather-descriptive-stats} summarizes the distributions of the five scalar weather variables and grouped \texttt{weather\_code} frequencies.

\subsection{Gold Datasets Release}
\label{app:gold-dataset-details}

The gold release is the model-ready tier of RIDE. It contains two types of artifacts: a shared benchmark core, which defines the common snapshots, splits, targets, and evaluation metadata, and four representation-specific datasets tailored to different prediction approaches. In the current release, these downstream datasets correspond to the tabular, sequential, GNN, and graph-event benchmark variants. In addition, the gold release provides lite and standard versions, which instantiate the same benchmark design at different scales. The following paragraphs first describe the shared core artifact and the lite/standard variants, then the representation-specific datasets built on top of them.

\paragraph{Core dataset.} 
The core gold dataset is the authoritative benchmark artifact for RIDE. It defines the prediction problem through a fixed set of snapshots, train/test splits, active-train instances, and future delay targets, so that downstream users can build new gold datasets, feature constructions, or model families while evaluating on exactly the same prediction instances and target values. This design makes comparisons across papers possible without requiring prior work to be retrained or adapted to a different snapshot definition, split, or target construction. As explained in the main paper, the benchmark is organized around the notion of a snapshot, that is, a timestamped view of the railway network from which active trains are identified, model inputs are constructed, and future delay targets are defined; see Figure~\ref{fig:snapshot-fig} for an illustration.

A core dataset is defined by the following components:
\begin{itemize}
\item the training and test period bounds $T^{\mathrm{train}}_{\mathrm{start}}$, $T^{\mathrm{train}}_{\mathrm{end}}$, $T^{\mathrm{test}}_{\mathrm{start}}$, and $T^{\mathrm{test}}_{\mathrm{end}}$. These correspond to \texttt{start\_train\_day}, \texttt{end\_train\_day}, \texttt{start\_test\_day}, and \texttt{end\_test\_day};
\item the numbers of sampled training and test snapshots, $N_{\mathrm{train}}$ and $N_{\mathrm{test}}$, given by \texttt{n\_train} and \texttt{n\_test};
\item the horizon parameter $n$ (\texttt{n\_future}), giving the number of future events used to define the prediction targets for each active train;
\item the parameters $\Delta_{\mathrm{beg}}$ (\texttt{idle\_time\_beg}) and $\Delta_{\mathrm{end}}$ (\texttt{idle\_time\_end}), which specify how much time before its first event and after its last event a train is still considered present in a snapshot;
\item the additional tail-safety buffer $\Delta_{\mathrm{tail}}$, stored as \texttt{tail\_safety\_buffer\_min} and used to separate the training and test periods;
\item the training snapshot set $\mathcal{S}_{\mathrm{train}}$, consisting of sampled snapshot timestamps used for training;
\item the test snapshot set $\mathcal{S}_{\mathrm{test}}$, consisting of sampled snapshot timestamps used for testing.
\end{itemize}

For a journey with planned start time $t^{\mathrm{planned}}_{\mathrm{start}}$, observed start time $t^{\mathrm{obs}}_{\mathrm{start}}$, and observed end time $t^{\mathrm{obs}}_{\mathrm{end}}$, the core benchmark categorically defines its activity window as
\[
t^{\mathrm{appear}}
=
\min\!\bigl(t^{\mathrm{planned}}_{\mathrm{start}} - \Delta_{\mathrm{beg}},\; t^{\mathrm{obs}}_{\mathrm{start}}\bigr),
\qquad
t^{\mathrm{disappear}}
=
t^{\mathrm{obs}}_{\mathrm{end}} + \Delta_{\mathrm{end}}.
\]
Given any snapshot time $t^{\mathrm{snap}}$, this definition uniquely determines the active-train set through the following half-open interval condition:
\[
t^{\mathrm{appear}} \le t^{\mathrm{snap}} < t^{\mathrm{disappear}}.
\]
The release code exposes reusable implementations of both steps. Activity windows are computed by \texttt{compute\_journey\_activity\_windows}, and active trains at a snapshot are selected by \texttt{get\_active\_mask\_for\_snapshot}.

The snapshot sets are sampled from the seconds at which at least one journey is active. Training snapshots are sampled uniformly without replacement from the active seconds in the training period, yielding $\mathcal{S}_{\mathrm{train}}$ with $|\mathcal{S}_{\mathrm{train}}| = N_{\mathrm{train}}$. For the test split, the nominal start time is shifted forward when needed to avoid overlap with trains from the last training service day. Let $t^{\mathrm{last}}_{\mathrm{end}}$ denote the latest observed end time among journeys from the final training day. The effective test start time is
\[
\widetilde{T}^{\mathrm{test}}_{\mathrm{start}}
=
\max\!\left(
T^{\mathrm{test}}_{\mathrm{start}},
t^{\mathrm{last}}_{\mathrm{end}} + \Delta_{\mathrm{end}} + \Delta_{\mathrm{tail}}
\right).
\]
Test snapshots are then sampled uniformly without replacement from the active seconds in the resulting test period, yielding $\mathcal{S}_{\mathrm{test}}$ with $|\mathcal{S}_{\mathrm{test}}| = N_{\mathrm{test}}$.
The benchmark-defining parameters and the realized snapshot sets are stored in the core specification file, \texttt{dataset\_core\_spec.yaml}. Release metadata and output paths are stored² separately in \texttt{metadata.yaml}.

The \texttt{test\_eval\_table.parquet} file is the central evaluation artifact of the core dataset. Its purpose is to make the test prediction instances and target values fully explicit: downstream users can build any representation or model they want, but evaluation is performed by matching predictions against this fixed table. Consequently, users only need to provide predicted delays in seconds for the valid target slots, and the evaluation code automatically applies the target mask and computes all aggregate and breakdown metrics. The table contains one row per \texttt{(snapshot, active train)} pair. Each row records:
\begin{itemize}
\item the prediction instance identifiers: the snapshot time \texttt{ts}, \texttt{train\_id}, and \texttt{service\_date};
\item the current delay state: \texttt{last\_known\_delay}, defined as $d_{s,i}^{\mathrm{last}}$ in Section~\ref{sec:prediction-task};
\item the future observation times: \texttt{future\_obs\_ts\_1}, \ldots, \texttt{future\_obs\_ts\_n};
\item the future operational-point identifiers: \texttt{future\_op\_id\_1}, \ldots, \texttt{future\_op\_id\_n};
\item the future event types: \texttt{future\_event\_type\_1}, \ldots, \texttt{future\_event\_type\_n};
\item the future delay targets: \texttt{future\_delay\_1}, \ldots, \texttt{future\_delay\_n}.
\end{itemize}
The future-delay columns define the common target values against which all downstream gold datasets are evaluated. If fewer than $n$ valid future events remain for an active train, the invalid target slots are encoded with placeholder values: \texttt{future\_op\_id} is set to $-1$, \texttt{future\_event\_type} is set to $-1$, \texttt{future\_delay} is set to \texttt{NaN}, and \texttt{future\_obs\_ts} is set to \texttt{NaT}.

\paragraph{Lite and standard variants.}
The gold release is provided in two scales. The lite variant is designed for fast iteration and lower-resource experimentation, and can also serve as the primary benchmark tier for users with limited computational resources. The standard variant is the main benchmark scale used for the main model comparison in this paper. Each variant has its own core artifact, including a separate \texttt{dataset\_core\_spec.yaml}, realized training and test snapshot sets, \texttt{test\_eval\_table.parquet}, and \texttt{metadata.yaml}. The core parameters and snapshot sets are stored in \texttt{dataset\_core\_spec.yaml}, while \texttt{metadata.yaml} stores release metadata and output paths. The representation-specific tabular, sequential, GNN, and graph-event datasets are then built separately from the corresponding lite or standard core, so that all datasets within a given tier share exactly the same prediction instances and targets.

The same construction code can instantiate additional cores for custom studies with different parameters, but the released lite and standard cores are the fixed reference tiers for comparable reporting across papers.

Both variants use the same training period, test period, horizon, activity-window definition, and tail-safety buffer. They differ only in the number of sampled training and test snapshots. Table~\ref{tab:gold_core_variant_params} reports the core parameters, while Table~\ref{tab:gold_core_variant_stats} reports realized scale statistics for each variant.

\begin{table}[H]
\centering
\begin{tabular}{lcc}
\toprule
Parameter & Lite & Standard \\
\midrule
$T^{\mathrm{train}}_{\mathrm{start}}$ & 2023-01-01 & 2023-01-01 \\
$T^{\mathrm{train}}_{\mathrm{end}}$ & 2024-12-31 & 2024-12-31 \\
$T^{\mathrm{test}}_{\mathrm{start}}$ & 2025-01-01 & 2025-01-01 \\
$T^{\mathrm{test}}_{\mathrm{end}}$ & 2025-12-31 & 2025-12-31 \\
$N_{\mathrm{train}}$ & 15{,}000 & 50{,}000 \\
$N_{\mathrm{test}}$ & 3{,}000 & 10{,}000 \\
$n$ & 15 & 15 \\
$\Delta_{\mathrm{beg}}$ & 5 min & 5 min \\
$\Delta_{\mathrm{end}}$ & 5 min & 5 min \\
$\Delta_{\mathrm{tail}}$ & 1 min & 1 min \\
\bottomrule
\end{tabular}
\caption{Core parameters for the lite and standard gold variants.}
\label{tab:gold_core_variant_params}
\end{table}

\begin{table}[H]
\centering
\begin{tabular}{lcc}
\toprule
Statistic & Lite & Standard \\
\midrule
training active-train rows & 3{,}153{,}365 & 10{,}554{,}576 \\
test active-train rows & 623{,}314 & 2{,}093{,}550 \\
valid training targets & 32{,}638{,}929 & 109{,}200{,}414 \\
valid test targets & 6{,}528{,}133 & 21{,}941{,}499 \\
mean active trains per training snapshot & 210.22 & 211.09 \\
mean active trains per test snapshot & 207.77 & 209.36 \\
mean valid targets per training row & 10.35 & 10.35 \\
mean valid targets per test row & 10.47 & 10.48 \\
\bottomrule
\end{tabular}
\caption{Realized scale statistics for the lite and standard gold variants.}
\label{tab:gold_core_variant_stats}
\end{table}

\paragraph{Tabular dataset.}
\label{app:gold_tabular_dataset}
The tabular gold dataset is the fixed-vector materialization of the core benchmark. It contains one row per active train at one snapshot and is used by the MLP, XGBoost, and Transformer models in our benchmark. For each split, the dataset stores input features in \texttt{x.npy}, target values in \texttt{y.npy}, target-validity indicators in \texttt{y\_mask.npy}, and row metadata in \texttt{md.npy}. The row metadata columns are \texttt{snapshot\_ts}, \texttt{train\_id}, and \texttt{service\_date}. The exported column contract is stored in \texttt{scheme.yaml}, while the fitted preprocessing statistics are stored in \texttt{normalization.yaml}. In the released lite and standard variants, \texttt{x.npy} has 452 input features, \texttt{y.npy} has 15 future-delay targets, and \texttt{y\_mask.npy} has 15 corresponding validity indicators.

Table~\ref{tab:gold_tabular_features} summarizes the input feature groups in \texttt{x.npy}. Train metadata come from the silver \texttt{journeys} table and include an indicator for SNCB/NMBS-operated services and a one-hot encoding of a coarse category derived from the journey \texttt{train\_relation} field. Snapshot-time features are derived from the core snapshot timestamp and encode day of week together with cyclical hour-of-day and day-of-year information. Event-window features come from the silver \texttt{events} table. Link features use the inferred \texttt{deduced\_paths} stored in \texttt{journeys} together with silver \texttt{node\_links}. For each active train, the pipeline estimates its current or next oriented \texttt{node\_link} along the inferred path, then records up to 10 oriented node links ahead of the train. For each such slot, the tabular vector stores the link distance, a traffic count, an average delay, and a placeholder flag. Weather features come from the silver \texttt{weather} table and are computed at the snapshot hour from the weather observations at the previous and next event operational points; the two values are averaged when both sides are available, and the available side is used when one side is padded. Operational-point embeddings are computed from the silver \texttt{node\_links} graph.

The remaining input features describe event windows around the snapshot. The past window contains 15 previously observed event slots and records, for each slot, the planned-time offset from the snapshot, the observed delay, the event type, and an operational-point embedding. The future window also contains 15 slots, matching the core horizon, but only uses information that is known at prediction time: the planned-time offset, the scheduled event type, and the operational-point embedding. When an event window extends beyond the available journey events, event identifiers are padded with the missing-event value $-1$; embeddings for missing event identifiers are zero vectors, with padded past event types set to departure and padded future event types set to arrival. Missing link slots are marked by the link placeholder flag, and invalid future target slots are excluded through \texttt{y\_mask.npy}. The future delays themselves are not included in \texttt{x.npy}; they are stored as residual targets in \texttt{y.npy}, and row metadata is stored in \texttt{md.npy}, as summarized in Table~\ref{tab:gold_tabular_others}.

\begin{figure}[H]
\centering
\includegraphics[width=\textwidth]{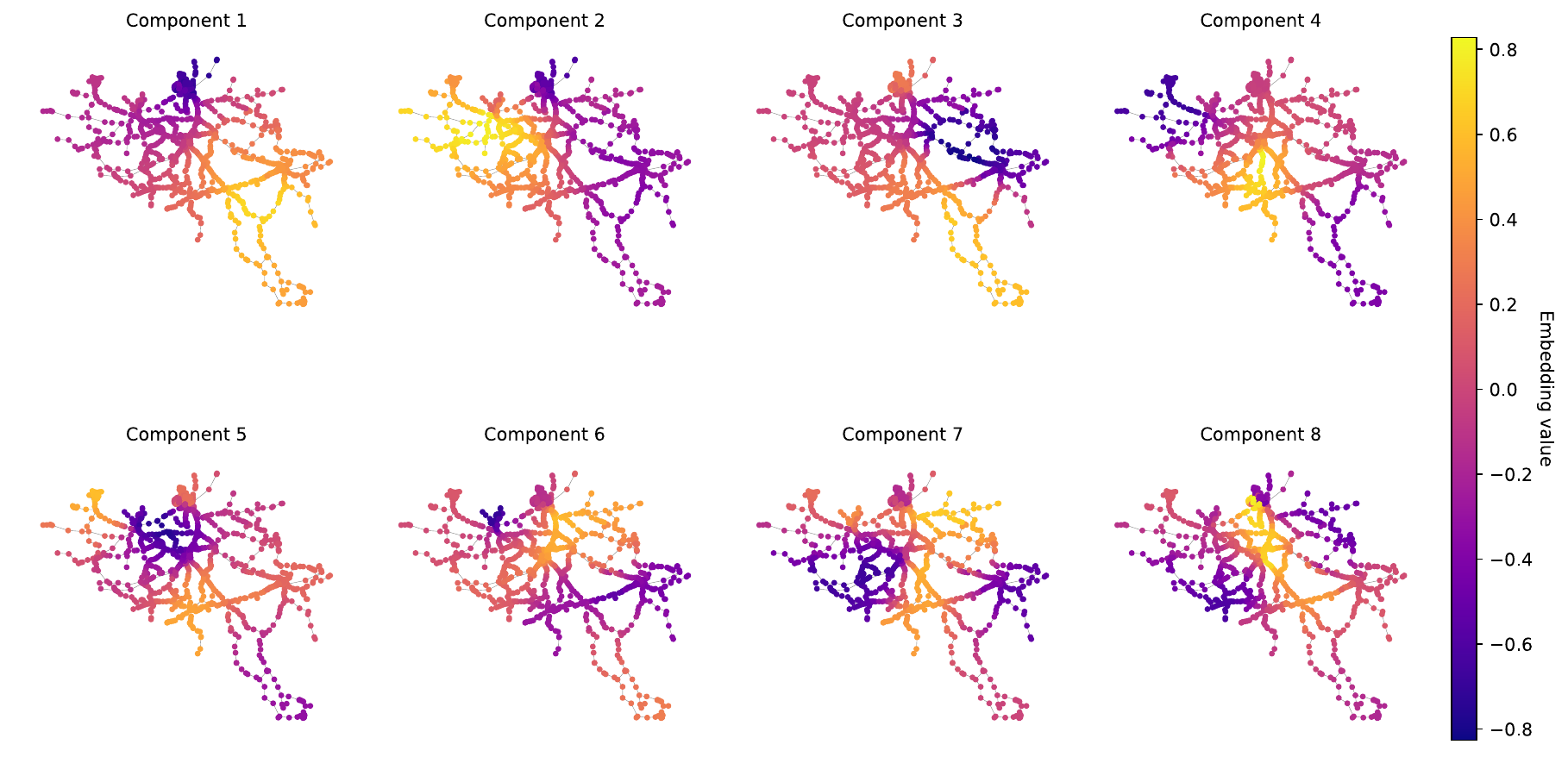}
\caption{The 8 normalized-Laplacian embedding components used as operational-point features in gold datasets. Each panel colors the silver railway graph by one embedding component.}
\label{fig:gold_station_embeddings}
\end{figure}

Operational-point embeddings are computed from the silver railway graph. The code builds an undirected adjacency from \texttt{node\_links}, computes normalized-Laplacian eigenmap coordinates, keeps the first 8 nonzero components, and normalizes each node embedding to unit length. These components provide a compact topological encoding of where an event occurs in the railway network; Figure~\ref{fig:gold_station_embeddings} visualizes the embedding components. Because the railway graph is reconstructed from source files downloaded after the 2025 test period, these embeddings are not a perfect historical snapshot of the infrastructure available at the end of the training period in 2024. In practice, this leakage is limited: only 11 operational points appearing in test events are absent from the training period, out of 1355 operational points overall, and they represent 0.00015\% of test events. Its effect is also narrow, since it can only slightly improve the embedding coordinates assigned to these new stations.

All preprocessing statistics are fitted on training samples only and then reused for both training and test arrays, mimicking the information available in a real forward-in-time evaluation. The target array stores preprocessed residual future delays, with \texttt{future\_delay\_delta\_i} first defined as the future delay at horizon $i$ minus \texttt{last\_known\_delay}. For each horizon, these target residuals are transformed with signed sqrt scaling and then z-scored using training-split statistics recorded in \texttt{normalization.yaml}. At evaluation time, model outputs are transformed back to residual-delay seconds with the same stored statistics and then shifted by \texttt{last\_known\_delay} before being matched against \texttt{test\_eval\_table.parquet}. The corresponding \texttt{y\_mask.npy} array marks which future event slots are valid and should be included in training losses and evaluation metrics.

\begin{table}[H]
\centering
\small
\renewcommand{\arraystretch}{1.3}
\begin{tabular}{p{3.0cm} p{5.5cm} p{1.4cm} p{2.4cm}}
\toprule
Feature group & Content & Encoding & Normalization \\
\midrule
train operator & indicator equal to one for SNCB/NMBS-operated services & binary & none \\
train relation type & coarse category derived from the journey \texttt{train\_relation} field & one-hot & none \\
day of week & snapshot day-of-week indicators & one-hot & none \\
hour of day & sine/cosine hour-of-day features at frequencies 1, 2, and 4 & 6 scalars & none \\
day of year & sine/cosine day-of-year features at frequencies 1, 2, and 4 & 6 scalars & none \\
link \# distance & distance of oriented node link \# ahead of the train & scalar & min--max \\
link \# traffic count & number of active trains on oriented node link \# ahead of the train & scalar & min--max \\
link \# average delay & average delay of active trains on oriented node link \# ahead of the train & scalar & signed sqrt, z-score \\
link \# placeholder flag & indicator that oriented node link \# ahead of the train is missing or padded & binary & none \\
temperature & average temperature at the previous and next event operational points at the snapshot hour & scalar & z-score \\
rain & average rain at the previous and next event operational points at the snapshot hour & scalar & signed sqrt, z-score \\
snowfall & average snowfall at the previous and next event operational points at the snapshot hour & scalar & signed sqrt, z-score \\
relative humidity & average relative humidity at the previous and next event operational points at the snapshot hour & scalar & z-score \\
wind speed & average wind speed at the previous and next event operational points at the snapshot hour & scalar & z-score \\
past event \# timing & planned-time offset of past event \# & scalar & signed sqrt, z-score \\
past event \# delay & observed delay of past event \# & scalar & signed sqrt, z-score \\
past event \# type & event type of past event \# & one-hot & none \\
past event \# embedding & 8-dimensional operational-point embedding of past event \# & 8 scalars & none \\
future event \# timing & planned-time offset of future event \# & scalar & signed sqrt, z-score \\
future event \# type & event type of future event \# & one-hot & none \\
future event \# embedding & 8-dimensional operational-point embedding of future event \# & 8 scalars & none \\
\bottomrule
\end{tabular}
\vspace{0.3em}

\parbox{\linewidth}{\footnotesize\raggedright
\# indexes repeated slots: $\#=1,\ldots,10$ for node links ahead of the train, and $\#=1,\ldots,15$ for past and future event slots. Weather features fall back to the available side when one of the two event operational points is padded.
}
\caption{Input feature groups in the tabular gold dataset's \texttt{x.npy} array.}
\label{tab:gold_tabular_features}
\end{table}
\renewcommand{\arraystretch}{1.0}

\begin{table}[H]
\centering
\small
\begin{tabular}{p{2.4cm} p{3.7cm} p{6.5cm}}
\toprule
Array & Columns & Description \\
\midrule
\begin{tabular}[t]{@{}l@{}}\texttt{y.npy}\end{tabular} & \begin{tabular}[t]{@{}l@{}}
\texttt{future\_delay\_delta\_1}, \\
\ldots, \\
\texttt{future\_delay\_delta\_15}
\end{tabular} & normalized residual future-delay targets, defined from future delays minus \texttt{last\_known\_delay} \\
\begin{tabular}[t]{@{}l@{}}\texttt{y\_mask.npy}\end{tabular} & \begin{tabular}[t]{@{}l@{}}
\texttt{is\_target\_valid\_1}, \\
\ldots, \\
\texttt{is\_target\_valid\_15}
\end{tabular} & boolean indicators specifying which future event slots are valid and should be included in training losses and evaluation metrics \\
\begin{tabular}[t]{@{}l@{}}\texttt{md.npy}\end{tabular} & \begin{tabular}[t]{@{}l@{}}
\texttt{snapshot\_ts}, \\
\texttt{train\_id}, \\
\texttt{service\_date}
\end{tabular} & row metadata identifying the snapshot and active train associated with each sample \\
\bottomrule
\end{tabular}
\caption{Target, mask arrays and metadata files in the tabular gold dataset.}
\label{tab:gold_tabular_others}
\end{table}

\paragraph{Sequential dataset.}
\label{app:gold_sequential_dataset}
The sequential gold dataset uses the same sample set as the tabular dataset: each sample corresponds to one active train at one snapshot, with the same metadata keys, target column names, target-validity mask, and residual target definition. It uses the same underlying 452 input columns listed in \texttt{scheme.yaml}. It reorganizes the tabular input vector into static features, past-event sequences, and future-known-event sequences, and is used by the LSTM model in our benchmark. The resulting arrays are summarized in Table~\ref{tab:gold_sequential_arrays}. The static array contains train metadata, snapshot-time features, link features, and weather features. The past-event sequence contains the 15 past event slots in chronological order, from older events toward the most recent event before the snapshot. Each past step contains 13 features: planned-time offset, observed delay, three event-type indicators, and eight operational-point embedding components. The future-known sequence contains the 15 future event slots in horizon order. Each future step contains 12 features: planned-time offset, three event-type indicators, and eight operational-point embedding components. Future delays are not included in the input sequence; they remain targets in \texttt{y.npy}.

As in the tabular dataset, preprocessing statistics are fitted on training samples only and stored in \texttt{normalization.yaml}. The sequential dataset differs in one normalization detail: for repeated event-window quantities, the statistics are shared across slots for \texttt{past\_planned\_delta\_*}, \texttt{future\_planned\_delta\_*}, \texttt{past\_delay\_sec\_*}, and \texttt{future\_delay\_delta\_*}, so that the same transformation is applied at each time step of the corresponding sequence.

\begin{table}[H]
\centering
\small
\begin{tabular}{p{3.7cm} p{1.4cm} p{7.5cm}}
\toprule
Array & Shape/row & Description \\
\midrule
\texttt{x\_static.npy} & 77 & static input features: train metadata, snapshot-time features, link features, and weather features \\
\texttt{x\_past\_seq.npy} & $15 \times 13$ & past-event input sequence; each step contains planned-time offset, observed delay, event type, and operational-point embedding features \\
\texttt{x\_future\_known\_seq.npy} & $15 \times 12$ & future-known input sequence; each step contains planned-time offset, event type, and operational-point embedding features \\
\texttt{y.npy} & 15 & normalized residual future-delay targets \\
\texttt{y\_mask.npy} & 15 & Boolean indicators specifying which future target slots are valid \\
\texttt{md.npy} & 3 & row metadata: snapshot timestamp, train identifier, and service date \\
\bottomrule
\end{tabular}
\caption{Arrays in the sequential gold dataset.}
\label{tab:gold_sequential_arrays}
\end{table}

\begin{figure}[H]
  \centering
  \begin{tikzpicture}[
    >=Latex,
    x=1.42cm,
    y=1.42cm,
    station/.style={circle, draw=black, fill=white, minimum size=7mm, inner sep=0pt, font=\small},
    train/.style={rectangle, draw=black, fill=black!10, minimum width=6mm, minimum height=4mm, inner sep=0pt, font=\scriptsize},
    rail/.style={draw=black!70, line width=1pt},
    past/.style={draw=teal!70!black, line width=0.9pt},
    future/.style={draw=orange!85!black, line width=0.9pt}
  ]
    \node[draw=black!30, rounded corners=2pt, fill=white, align=left, anchor=north west, inner sep=2.5pt, font=\scriptsize] (legend) at (-1.3,5.35) {
      \begin{tabular}{@{}ll@{}}
        \tikz{\node[station, minimum size=2.2mm] {}; } & station node \\
        \tikz{\node[train, minimum width=2.9mm, minimum height=1.6mm] {}; } & train node \\
        \tikz{\draw[rail] (0,0) -- (0.34,0); } & rail link \\
        \tikz{\draw[past] (0,0) -- (0.34,0); } & past event edge \\
        \tikz{\draw[future] (0,0) -- (0.34,0); } & future event edge \\
      \end{tabular}
    };

    \node[station] (s1) at (0.9,0) {$s_1$};
    \node[station] (s2) at (4.1,0) {$s_2$};
    \node[station] (s3) at (6.4,2.3) {$s_3$};

    \node[draw=black!30, rounded corners=2pt, fill=white, align=left, anchor=east, font=\scriptsize, inner sep=3pt] (s1box) at (0.1,-1) {
      \textbf{$s_1$} \\
      lat, lon: 50.85, 4.36 \\
      weather: 18$^\circ$C, \ldots \\
      embedding: [...] \\
      nb. trains: 1 \\
      avg. delay: 120 s
    };
    \draw[black!50, dashed] (s1box.east) -- (s1.west);

    \node[draw=black!30, rounded corners=2pt, fill=white, align=left, anchor=north, font=\scriptsize, inner sep=3pt] (s2box) at (4.1,-1.35) {
      \textbf{$s_2$} \\
      lat, lon: 50.85, 4.44 \\
      weather: 19$^\circ$C, \ldots \\
      embedding: [...] \\
      nb. trains: 0 \\
      avg. delay: 0 s
    };
    \draw[black!50, dashed] (s2box.north) -- (s2.south);

    \node[draw=black!30, rounded corners=2pt, fill=white, align=left, anchor=west, font=\scriptsize, inner sep=3pt] (s3box) at (6.5,3.35) {
      \textbf{$s_3$} \\
      lat, lon: 50.88, 4.50 \\
      weather: 17$^\circ$C, \ldots \\
      embedding: [...] \\
      nb. trains: 0 \\
      avg. delay: 0 s
    };
    \draw[black!50, dashed] (s3box.south) -- (s3.north east);

    \node[draw=black!30, rounded corners=2pt, fill=white, align=left, anchor=south, font=\scriptsize, inner sep=3pt] (e12box) at (2.5,0.45) {
      \textbf{$s_1$--$s_2$} \\
      dist: 2.3 km \\
      angle: uv $0^\circ$, vu $180^\circ$ \\
      nb. trains: uv 0, vu 0 \\
      avg. delay: uv 0 s, vu 0 s
    };
    \draw[black!50, dashed] (e12box.south) -- (2.5,0);

    \node[draw=black!30, rounded corners=2pt, fill=white, align=left, anchor=west, font=\scriptsize, inner sep=3pt] (e23box) at (3.5,1.85) {
      \textbf{$s_2$--$s_3$} \\
      dist: 2.8 km \\
      angle: uv $45^\circ$, vu $225^\circ$ \\
      nb. trains: uv 0, vu 1 \\
      avg. delay: uv 0 s, vu 60 s
    };
    \draw[black!50, dashed] (e23box.south) -- (5.25,1.15);

    \draw[rail] (s1) -- (s2);
    \draw[rail] (s2) -- (s3);

    \node[train] (t1) at (0.9,1.45) {$t_1$};
    \node[train] (t2) at (5.2,-1.05) {$t_2$};

    \node[draw=black!30, rounded corners=2pt, fill=white, align=left, anchor=south, font=\scriptsize, inner sep=3pt] (t1box) at (0.9,2.25) {
      \textbf{$t_1$} \\
      operator: SNCB \\
      relation: IC \\
      weekday: Monday \\
      date: 22 Apr \\
      hour: 08:15
    };
    \draw[black!50, dashed] (t1box.south) -- (t1.north);

    \node[draw=black!30, rounded corners=2pt, fill=white, align=left, anchor=west, font=\scriptsize, inner sep=3pt] (t2box) at (5.9,-1.8) {
      \textbf{$t_2$} \\
      operator: SNCB \\
      relation: THAL \\
      weekday: Monday \\
      date: 22 Apr \\
      hour: 08:15
    };
    \draw[black!50, dashed] (t2box.west) -- (t2.east);

    \node[draw=teal!40!black, rounded corners=2pt, fill=white, align=left, anchor=east, font=\scriptsize, inner sep=3pt] (p11box) at (0.2,1.55) {
      \textbf{$t_1$--$s_1$ (past)} \\
      past idx: 1 \\
      type: A \\
      planned $\Delta$: -180 s \\
      delay: 120 s
    };

    \node[draw=orange!55!black, rounded corners=2pt, fill=white, align=left, anchor=east, font=\scriptsize, inner sep=3pt] (f11box) at (0.2,0.3) {
	      \textbf{$t_1$--$s_1$ (future)} \\
	      future idx: 1 \\
	      type: D \\
	      planned $\Delta$: -90 s \\
	      $y$ (delay $\Delta$): -30 s
	    };

    \node[draw=orange!55!black, rounded corners=2pt, fill=white, align=left, anchor=west, font=\scriptsize, inner sep=3pt] (f13box) at (3.25,3.55) {
	      \textbf{$t_1$--$s_3$ (future)} \\
	      future idx: 2 \\
	      type: P \\
	      planned $\Delta$: 420 s \\
	      $y$ (delay $\Delta$): 90 s
	    };

    \node[draw=teal!40!black, rounded corners=2pt, fill=white, align=left, anchor=west, font=\scriptsize, inner sep=3pt] (p23box) at (6.85,-0.35) {
      \textbf{$t_2$--$s_3$ (past)} \\
      past idx: 1 \\
      type: D \\
      planned $\Delta$: -240 s \\
      delay: 60 s
    };

    \node[draw=orange!55!black, rounded corners=2pt, fill=white, align=left, anchor=west, font=\scriptsize, inner sep=3pt] (f21box) at (0.85,-1.9) {
	      \textbf{$t_2$--$s_1$ (future)} \\
	      future idx: 2 \\
	      type: P \\
	      planned $\Delta$: 300 s \\
	      $y$ (delay $\Delta$): 60 s
	    };

    \node[draw=orange!55!black, rounded corners=2pt, fill=white, align=left, anchor=north, font=\scriptsize, inner sep=3pt] (f22box) at (5.58,0.75) {
	      \textbf{$t_2$--$s_2$ (future)} \\
	      future idx: 1 \\
	      type: P \\
	      planned $\Delta$: 60 s \\
	      $y$ (delay $\Delta$): 0 s
	    };

    \draw[past] (t1) to[bend right=8] coordinate[pos=0.52] (p11anchor) (s1);
    \draw[future] (t1) to[bend left=8] coordinate[pos=0.52] (f11anchor) (s1);
    \draw[future] (t1) to[bend left=25] coordinate[pos=0.54] (f13anchor) (s3);

    \draw[past] (t2) to[bend right=55] coordinate[pos=0.50] (p23anchor) (s3);
    \draw[future] (t2) -- coordinate[pos=0.50] (f22anchor) (s2);
    \draw[future] (t2) to[bend left=15] coordinate[pos=0.52] (f21anchor) (s1);
    \draw[black!50, dashed] (p11box.east) -- (p11anchor);
    \draw[black!50, dashed] (f11box.east) -- (f11anchor);
    \draw[black!50, dashed] (f13box.south) -- (f13anchor);
    \draw[black!50, dashed] (p23box.west) -- (p23anchor);
    \draw[black!50, dashed] (f22box.south) -- (f22anchor);
    \draw[black!50, dashed] (f21box.north) -- (f21anchor);
  \end{tikzpicture}
  \caption{Toy railway layout used to illustrate the heterogeneous GNN snapshot structure and features. The graph contains three consecutive station nodes together with two active trains: one stopped at station $s_1$ and one moving from station $s_3$ toward station $s_2$.}
  \label{fig:gnn_toy}
\end{figure}
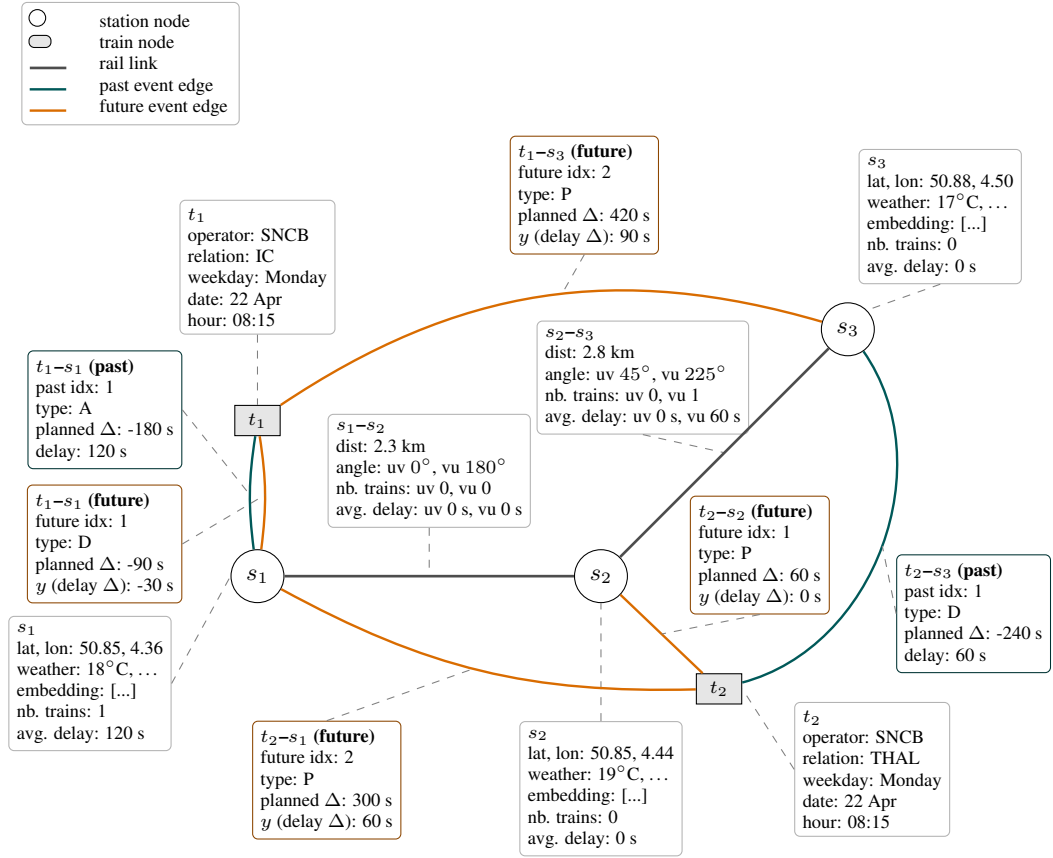

\begin{figure}[H]
  \centering
  \includegraphics[width=\linewidth]{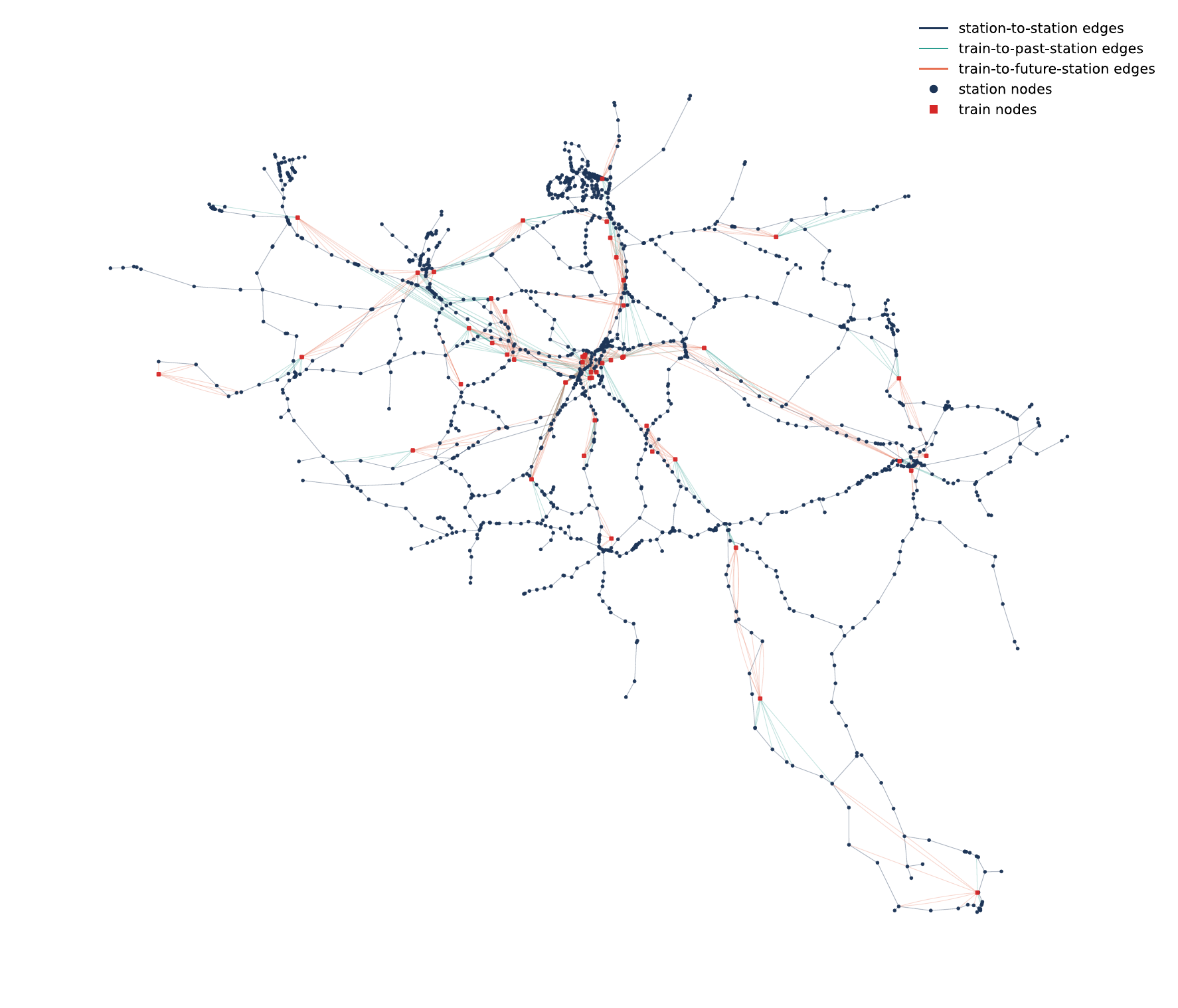}
  \caption{Visualization of a full-network GNN snapshot. Station nodes are placed at their geographic locations, rail links define the infrastructure graph, and train nodes are connected to the operational points associated with the 7 closest past and future events in their itinerary. We thin the window for readability, the released GNN dataset uses 15 past and 15 future event slots.}
  \label{fig:gnn_snapshot_country}
\end{figure}

\paragraph{GNN dataset.}
\label{app:gold_gnn_dataset}
The GNN gold dataset represents each benchmark snapshot as a heterogeneous graph rather than as one fixed-size vector per active train. This graph view keeps the main entities of the prediction problem explicit. Active trains are represented as train nodes carrying train metadata and snapshot-time features (Table~\ref{tab:gold_gnn_train_node_features}). Operational points are represented as station nodes carrying position, local weather, operational-point embeddings, and stopped-train traffic features (Table~\ref{tab:gold_gnn_station_node_features}). Railway infrastructure is represented by station-to-station edges carrying physical link distance and direction-specific \texttt{uv}/\texttt{vu} angle, traffic-count, and average-delay features (Table~\ref{tab:gold_gnn_station_edge_features}). Finally, each train's observed and scheduled itinerary context is represented by train-to-station edges: past edges encode previous events through planned offset, observed delay, event type, and recency rank (Table~\ref{tab:gold_gnn_past_edge_features}), while future edges carry scheduled future-event inputs and the associated residual-delay target (Table~\ref{tab:gold_gnn_future_edge_features}). This representation attaches network-context features such as distances, train counts, and average delays directly to the station nodes and railway links where they are measured, rather than folding them into a train-level vector. It therefore preserves the distinction between train-specific itinerary context and shared network state within each snapshot. An illustrative example of this representation is given in Figure~\ref{fig:gnn_toy}, and a full-network snapshot produced by the dataset construction is shown in Figure~\ref{fig:gnn_snapshot_country}.

The released artifact stores PyTorch Geometric graph objects for the training and test splits as \texttt{train/graphs\_part\_*.pt} and \texttt{test/graphs\_part\_*.pt}. These graph objects include reverse relations for all edge types. Past and future train-to-station reverse relations reuse the same edge attributes, while station-to-station reverse relations swap the \texttt{uv\_*} and \texttt{vu\_*} attributes so that direction-specific features remain aligned with the direction of the relation. The node and edge feature columns used in these graph objects are recorded in \texttt{feature\_spec.yaml}, and preprocessing statistics are stored in \texttt{normalization.yaml}. As in the tabular and sequential datasets, normalization statistics are fitted on the training split only. The target on each future train-to-station edge is the normalized residual future delay, defined from the future delay minus \texttt{last\_known\_delay}; the horizon rank is retained so that edge predictions can be mapped back to the corresponding core target slots. At evaluation time, predictions are transformed back to delay seconds and compared against the core evaluation table.

\begin{table}[H]
\centering
\small
\renewcommand{\arraystretch}{1.08}
\begin{tabular}{p{3.0cm} p{5.5cm} p{1.4cm} p{2.4cm}}
\toprule
Feature group & Content & Encoding & Normalization \\
\midrule
train operator & indicator equal to one for SNCB/NMBS-operated services & binary & none \\
train relation type & coarse category derived from the journey \texttt{train\_relation} field & one-hot & none \\
snapshot time of day & sine/cosine snapshot-hour features at frequencies 1, 2, and 4 & 6 scalars & none \\
snapshot day of year & sine/cosine day-of-year features at frequencies 1, 2, and 4 & 6 scalars & none \\
snapshot day of week & snapshot weekday category & one-hot & none \\
\bottomrule
\end{tabular}
\caption{Train-node input feature groups in the GNN gold dataset.}
\label{tab:gold_gnn_train_node_features}
\end{table}

\begin{table}[H]
\centering
\small
\renewcommand{\arraystretch}{1.08}
\begin{tabular}{p{3.0cm} p{5.5cm} p{1.4cm} p{2.4cm}}
\toprule
Feature group & Content & Encoding & Normalization \\
\midrule
latitude & station latitude & scalar & min--max \\
longitude & station longitude & scalar & min--max \\
temperature & hourly temperature at the station and snapshot hour & scalar & z-score \\
rain & hourly rain at the station and snapshot hour & scalar & signed sqrt, z-score \\
snowfall & hourly snowfall at the station and snapshot hour & scalar & signed sqrt, z-score \\
relative humidity & hourly relative humidity at the station and snapshot hour & scalar & z-score \\
wind speed & hourly wind speed at the station and snapshot hour & scalar & z-score \\
embedding & normalized-Laplacian operational-point embedding components & 8 scalars & none \\
stopped traffic count & number of active trains currently stopped at the station & scalar & min--max \\
stopped traffic delay & average current delay of active trains currently stopped at the station & scalar & signed sqrt, z-score \\
\bottomrule
\end{tabular}
\caption{Station-node input feature groups in the GNN gold dataset.}
\label{tab:gold_gnn_station_node_features}
\end{table}

\begin{table}[H]
\centering
\small
\renewcommand{\arraystretch}{1.08}
\begin{tabular}{p{3.0cm} p{5.5cm} p{1.4cm} p{2.4cm}}
\toprule
Feature group & Content & Encoding & Normalization \\
\midrule
link distance & physical distance of the railway node link & scalar & min--max \\
\texttt{uv} angle & sine and cosine of the node-link angle from $u$ to $v$ & 2 scalars & none \\
\texttt{vu} angle & sine and cosine of the node-link angle from $v$ to $u$ & 2 scalars & none \\
\texttt{uv} traffic count & number of active trains currently occupying the node link from $u$ to $v$ & scalar & min--max \\
\texttt{vu} traffic count & number of active trains currently occupying the node link from $v$ to $u$ & scalar & min--max \\
\texttt{uv} average delay & average current delay of active trains currently occupying the node link from $u$ to $v$ & scalar & signed sqrt, z-score \\
\texttt{vu} average delay & average current delay of active trains currently occupying the node link from $v$ to $u$ & scalar & signed sqrt, z-score \\
\bottomrule
\end{tabular}
\caption{Station-to-station edge input feature groups in the GNN gold dataset.}
\label{tab:gold_gnn_station_edge_features}
\end{table}

\begin{table}[H]
\centering
\small
\renewcommand{\arraystretch}{1.08}
\begin{tabular}{p{3.0cm} p{5.5cm} p{1.4cm} p{2.4cm}}
\toprule
Feature group & Content & Encoding & Normalization \\
\midrule
planned offset & planned time of the past event relative to the snapshot time & scalar & signed sqrt, z-score \\
observed delay & observed delay at the past event & scalar & signed sqrt, z-score \\
past rank & recency rank of the past event in the train itinerary window & scalar & min--max \\
event type & arrival, departure, or passage indicator & one-hot & none \\
\bottomrule
\end{tabular}
\caption{Train-to-past-station edge input feature groups in the GNN gold dataset.}
\label{tab:gold_gnn_past_edge_features}
\end{table}

\begin{table}[H]
\centering
\small
\renewcommand{\arraystretch}{1.08}
\begin{tabular}{p{3.0cm} p{5.5cm} p{1.4cm} p{2.4cm}}
\toprule
Feature group & Content & Encoding & Normalization \\
\midrule
planned offset & planned time of the future event relative to the snapshot time & scalar & signed sqrt, z-score \\
future rank & horizon rank of the future event in the prediction window & scalar & min--max \\
event type & arrival, departure, or passage indicator & one-hot & none \\
future delay residual & target delay at the future event minus \texttt{last\_known\_delay} & target & signed sqrt, z-score \\
\bottomrule
\end{tabular}
\caption{Train-to-future-station edge input and target groups in the GNN gold dataset.}
\label{tab:gold_gnn_future_edge_features}
\end{table}

\paragraph{Graph-event dataset.}
\label{app:gold_graph_event_dataset}
The graph-event gold dataset provides the model-ready inputs for the deterministic graph-event benchmark described in Appendix~\ref{app:modeldetails}. Unlike the learning-based gold datasets, it does not construct normalized feature tensors. Instead, it exports the event, journey, infrastructure, and travel-time lookup artifacts needed to run the graph-event simulation on the core test prediction instances. The released files are summarized in Table~\ref{tab:gold_graph_event_artifacts}. The \texttt{node\_links.parquet} file is copied from the silver release and keeps the columns \texttt{line\_section\_id}, \texttt{u\_node\_id}, \texttt{v\_node\_id}, \texttt{distance\_m}, and \texttt{link\_id}. The \texttt{test/events.parquet} file contains all event rows for trains appearing in the graph-event test journeys, with columns \texttt{train\_id}, \texttt{service\_date}, \texttt{op\_id}, \texttt{event\_type}, \texttt{planned\_ts}, \texttt{line\_dep}, and \texttt{line\_arr}. The \texttt{test/journeys.parquet} file contains one row per core test \texttt{(snapshot, active train)} pair and stores \texttt{ts}, \texttt{train\_id}, \texttt{service\_date}, \texttt{current\_event\_idx}, \texttt{last\_known\_delay}, \texttt{train\_relation}, and \texttt{path}; here \texttt{current\_event\_idx} identifies the last event before the snapshot, \texttt{train\_relation} is the coarse category derived from the journey \texttt{train\_relation} field, and \texttt{path} contains the journey-level inferred \texttt{deduced\_paths}. Finally, \texttt{travel\_time\_samples.pkl} stores travel-time statistics estimated from the full event sequences of trains active in the sampled training snapshots. These statistics are keyed by origin operational point, origin event type, departure line, destination operational point, destination event type, arrival line, and train-relation category, and store the number of samples, minimum, maximum, mean, median, and lower quantiles from 5\% to 45\%. They are used by the graph-event model to estimate expected travel times between consecutive events without using test-period observations.

\begin{table}[H]
\centering
\small
\begin{tabular}{p{4.2cm} p{8.1cm}}
\toprule
Artifact & Contents \\
\midrule
\texttt{node\_links.parquet} & silver railway node-link table, with link identifiers, endpoint operational points, line-section identifiers, and link distances \\
\texttt{test/events.parquet} & event rows for trains appearing in the graph-event test journeys, including operational point, event type, planned timestamp, and arrival/departure line context \\
\texttt{test/journeys.parquet} & one row per core test \texttt{(snapshot, active train)} pair, including current event index, last known delay, train relation, and inferred path information \\
\texttt{travel\_time\_samples.pkl} & training-snapshot travel-time statistics for event-pair contexts, used as lookup values by the graph-event simulation \\
\bottomrule
\end{tabular}
\caption{Artifacts in the graph-event gold dataset.}
\label{tab:gold_graph_event_artifacts}
\end{table}

\section{Benchmark Protocol, Implementations and Additional Experiments}

This section provides additional details on the benchmark protocol, model implementations, training procedure, and supplementary experiments. We first specify the evaluation setup in more detail, including target masking, aggregate metrics, and the horizon- and delay-delta-based breakdowns used in the main paper. We then describe the benchmark model implementations and training procedure, report the hyperparameter search spaces and selected configurations, and provide additional results on the lite benchmark tier together with the feature-family ablation study.

\subsection{Prediction Task and Metrics Details}
\label{app:predtaskandmetricsdetails}

For a given snapshot $s$ at time $t^{s}$, the goal is to predict, for every active train, the delay at the next $n$ scheduled events, where each event corresponds to a scheduled arrival, departure, or passage at an operational point (see Table~\ref{tab:silver_components}). In this appendix subsection, we provide additional detail on the practical evaluation protocol. Conveniently, the core dataset's \texttt{test\_eval\_table} contains all metadata required to instantiate the full evaluation procedure, so users only need to provide predicted delays in seconds indexed by the core key columns.

Each snapshot contains target slots for the next 15 scheduled events of each active train, but in practice not every train has 15 valid future targets, since some journeys end earlier. We therefore construct an explicit mask indicating which target slots correspond to valid future events. During training, masked targets are excluded from the training objectives of learning-based models. During evaluation, masked targets are automatically excluded from all reported metrics and breakdowns. 

For stochastic learning-based models, all reported metrics are first computed independently for each test-evaluation run and then summarized as mean and standard deviation across the fixed seeds $\{0,1,\dots,9\}$. Deterministic models are evaluated once, so only a single metric value is reported.

\paragraph{Aggregate Metrics.}
We compute aggregate metrics over the explicit set of valid target slots. Let $\mathcal{E}$ denote the set of all valid prediction targets in the evaluation split after masking. Each element $(s,i,j) \in \mathcal{E}$ corresponds to one active train instance $i$ in snapshot $s$ and one future event slot $j$. The aggregate mean absolute error and root mean squared error are then defined as
\[
\mathrm{MAE} = \frac{1}{|\mathcal{E}|}\sum_{(s,i,j) \in \mathcal{E}} \left| \hat{d}_{s,i,j} - d_{s,i,j} \right|,
\qquad
\mathrm{RMSE} = \sqrt{\frac{1}{|\mathcal{E}|}\sum_{(s,i,j) \in \mathcal{E}} \left( \hat{d}_{s,i,j} - d_{s,i,j} \right)^2 }.
\]

\paragraph{Prediction Horizon.}
For each valid target $(s,i,j) \in \mathcal{E}$, let
\[
h_{s,i,j} = \frac{t^{\mathrm{obs}}_{s,i,j} - t^{s}}{60}
\]
denote its prediction horizon in minutes. Horizon-based breakdowns are obtained by restricting evaluation to targets whose horizons fall within a given bin $[a,b)$. For such a bin, let
\[
\mathcal{E}_{[a,b)}^{\mathrm{hor}} = \{\, (s,i,j) \in \mathcal{E} \mid a \leq h_{s,i,j} < b \,\}.
\]
The corresponding bin-specific metrics are then defined as
\[
\begin{aligned}
\mathrm{MAE}_{[a,b)}^{\mathrm{hor}} &=
\frac{1}{|\mathcal{E}_{[a,b)}^{\mathrm{hor}}|}
\sum_{(s,i,j) \in \mathcal{E}_{[a,b)}^{\mathrm{hor}}}
\left| \hat{d}_{s,i,j} - d_{s,i,j} \right|,\\
\mathrm{RMSE}_{[a,b)}^{\mathrm{hor}} &=
\sqrt{
\frac{1}{|\mathcal{E}_{[a,b)}^{\mathrm{hor}}|}
\sum_{(s,i,j) \in \mathcal{E}_{[a,b)}^{\mathrm{hor}}}
\left( \hat{d}_{s,i,j} - d_{s,i,j} \right)^2
}.
\end{aligned}
\]
We use bin edges at $0, 5, 10, 15, 20, 25, 30, 35, 40, 45, \infty$.

\begin{figure}[H]
    \centering
    \includegraphics[width=0.6\linewidth]{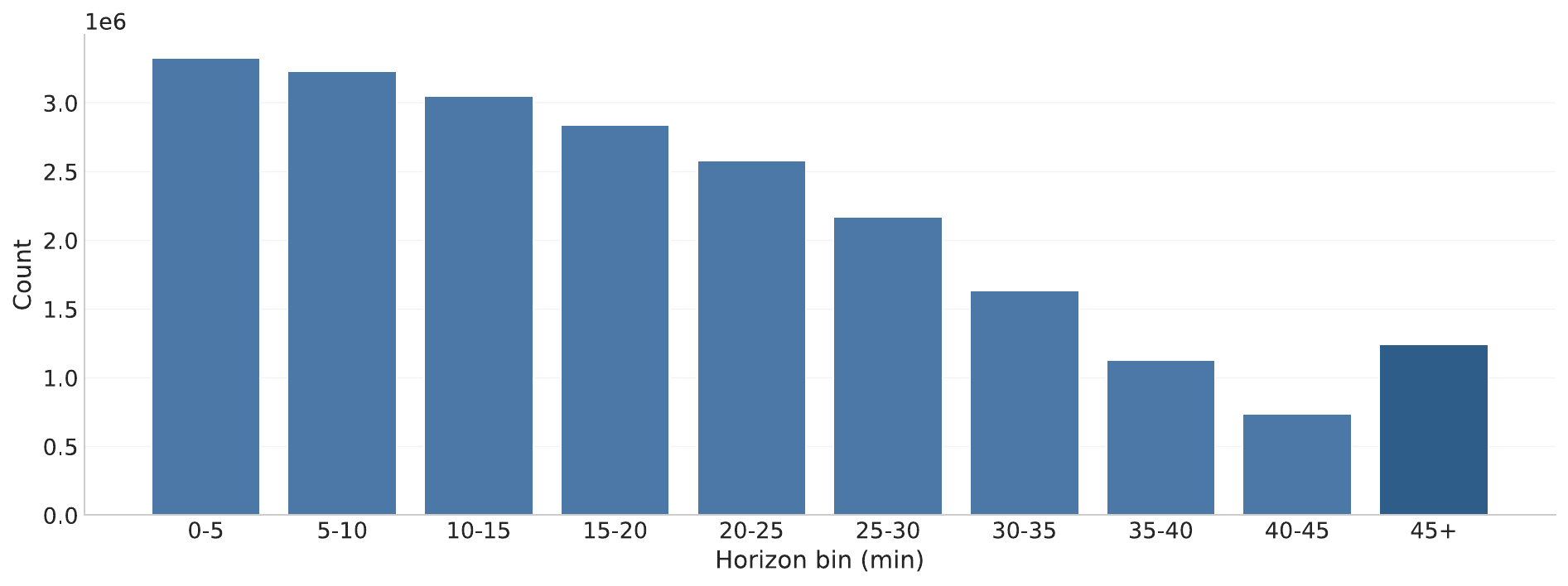}
    \caption{Overall distribution of valid targets across horizon bins on the standard tier.}
    \label{fig:overall_horizon_bin_counts}
\end{figure}

\begin{figure}[H]
    \centering
    \includegraphics[width=\linewidth]{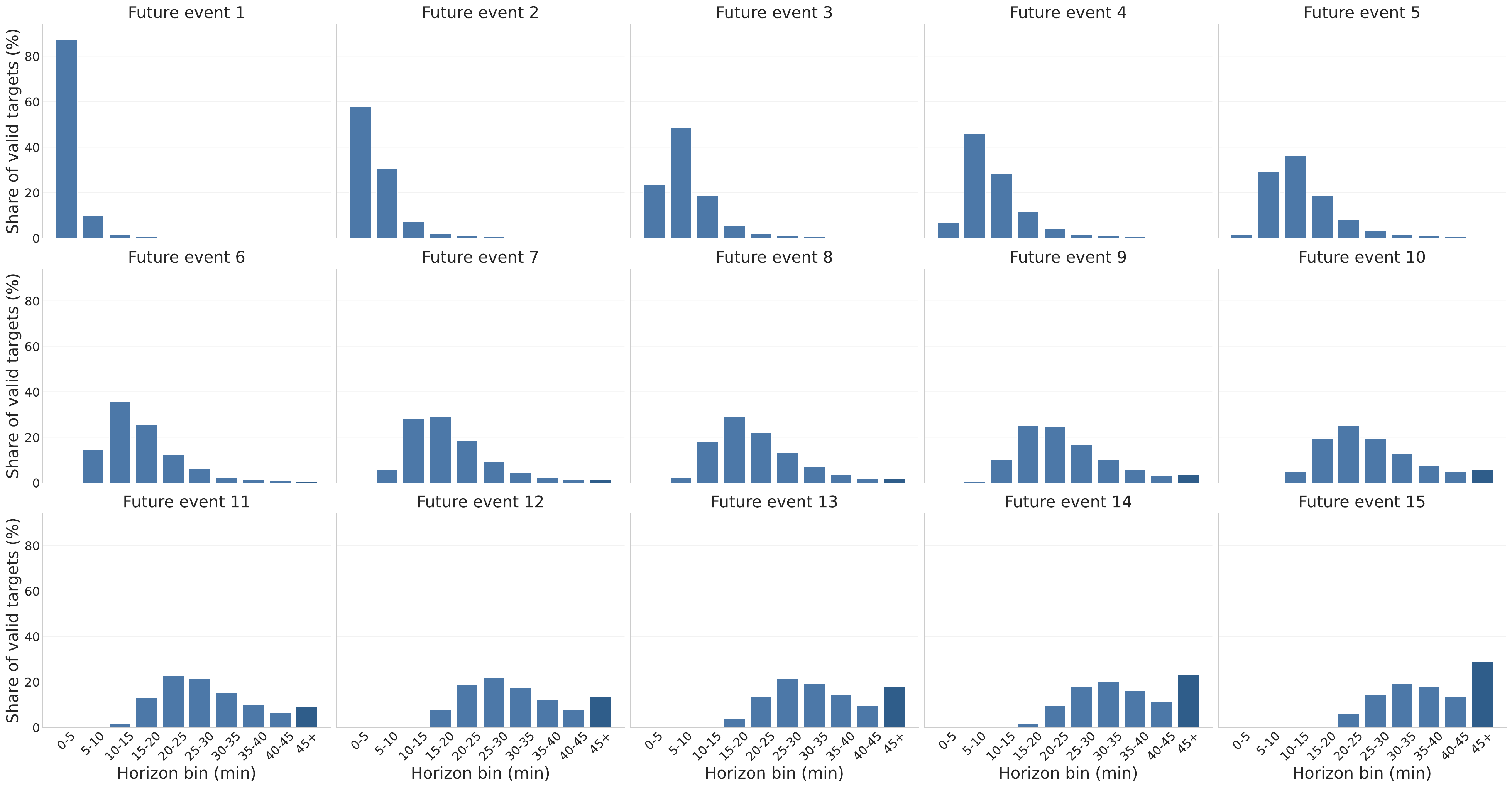}
    \caption{Distribution of valid targets across horizon bins broken down by future-event slot on the standard tier.}
    \label{fig:horizon_bin_counts_by_future_event}
\end{figure}

The final open-ended 45+ bin is used because target events become progressively scarcer at larger horizons under the fixed 15-event prediction window, which corresponds on average to about 40 minutes into the future. This effect is visible in the overall horizon distribution (Figure~\ref{fig:overall_horizon_bin_counts}), while the per-target breakdown further shows how these bins are distributed across the 15 future-event slots (Figure~\ref{fig:horizon_bin_counts_by_future_event}).

\begin{figure}[H]
    \centering
    \includegraphics[width=0.6\linewidth]{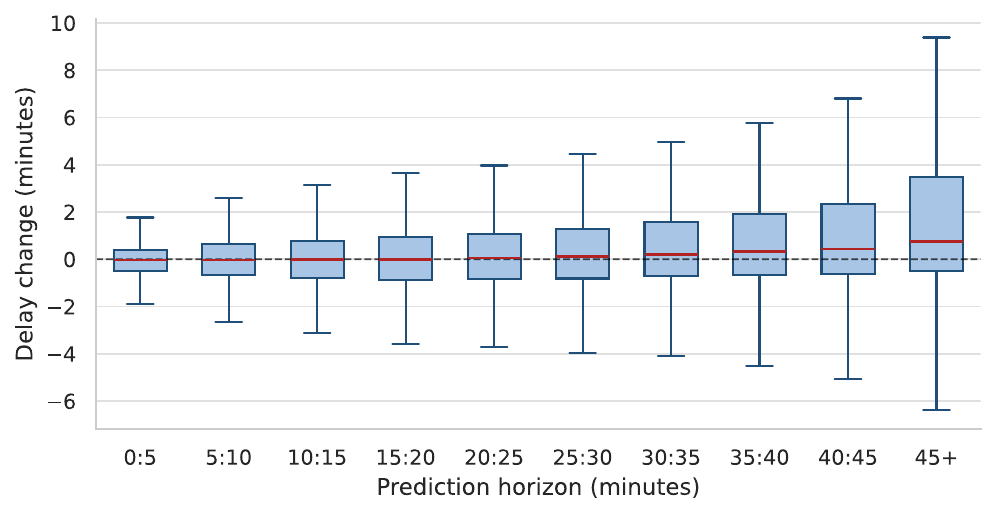}
    \caption{Delay-delta distribution by horizon bin on the standard tier. Boxes show the interquartile range with the median indicated by the horizontal line; whiskers extend to 1.5 times the interquartile range, and outliers are omitted for readability.}
    \label{fig:standard_delay_delta_by_horizon_boxplot}
\end{figure}

An important caveat is that the latest horizon bins are progressively biased toward trains that accumulate delay. This effect is especially pronounced in the final 45+ bin, because under the fixed future-event prediction window the later time bins become thinner by construction, as illustrated in Figure~\ref{fig:standard_delay_delta_by_horizon_boxplot}. As a result, differences across later horizon bins should not be interpreted as reflecting horizon length alone, but rather as a mixture of horizon and delay regime.

\paragraph{Delay Delta.}
For each valid target $(s,i,j) \in \mathcal{E}$, let
\[
\Delta d_{s,i,j} = \frac{d_{s,i,j} - d_{s,i}^{\mathrm{last}}}{60}
\]
denote its delay delta in minutes relative to the last known delay $d_{s,i}^{\mathrm{last}}$ defined in Section~\ref{sec:prediction-task}. Delay-delta-based breakdowns are obtained by restricting evaluation to targets whose delay deltas fall within a given bin $[a,b)$. For such a bin, let
\[
\mathcal{E}_{[a,b)}^{\Delta} = \{\, (s,i,j) \in \mathcal{E} \mid a \leq \Delta d_{s,i,j} < b \,\}.
\]
The corresponding bin-specific metrics are then defined as
\[
\begin{aligned}
\mathrm{MAE}_{[a,b)}^{\Delta} &=
\frac{1}{|\mathcal{E}_{[a,b)}^{\Delta}|}
\sum_{(s,i,j) \in \mathcal{E}_{[a,b)}^{\Delta}}
\left| \hat{d}_{s,i,j} - d_{s,i,j} \right|,\\
\mathrm{RMSE}_{[a,b)}^{\Delta} &=
\sqrt{
\frac{1}{|\mathcal{E}_{[a,b)}^{\Delta}|}
\sum_{(s,i,j) \in \mathcal{E}_{[a,b)}^{\Delta}}
\left( \hat{d}_{s,i,j} - d_{s,i,j} \right)^2
}.
\end{aligned}
\]
We use bin edges at $-\infty, -5, -2, -1, -0.5, 0, 0.5, 1, 2, 5, 10, \infty$.

\begin{figure}[H]
    \centering
    \includegraphics[width=0.6\linewidth]{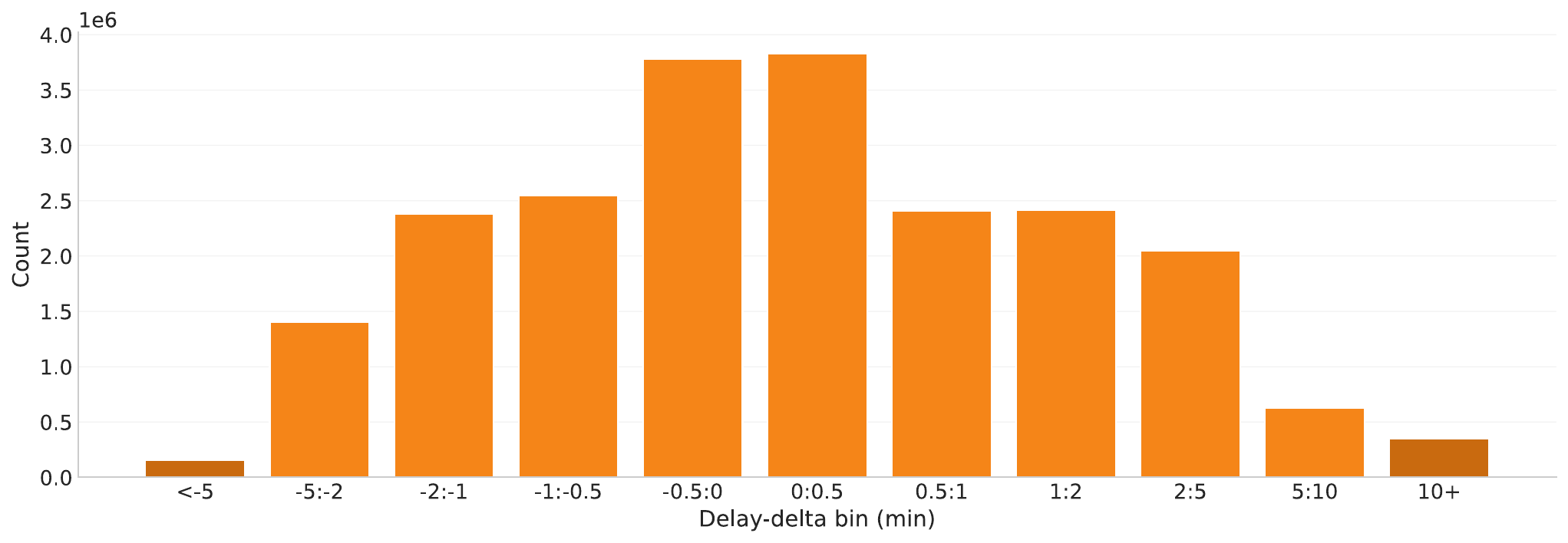}
    \caption{Overall distribution of valid targets across delay-delta bins on the standard tier.}
    \label{fig:overall_delay_delta_bin_counts}
\end{figure}

\begin{figure}[H]
    \centering
    \includegraphics[width=\linewidth]{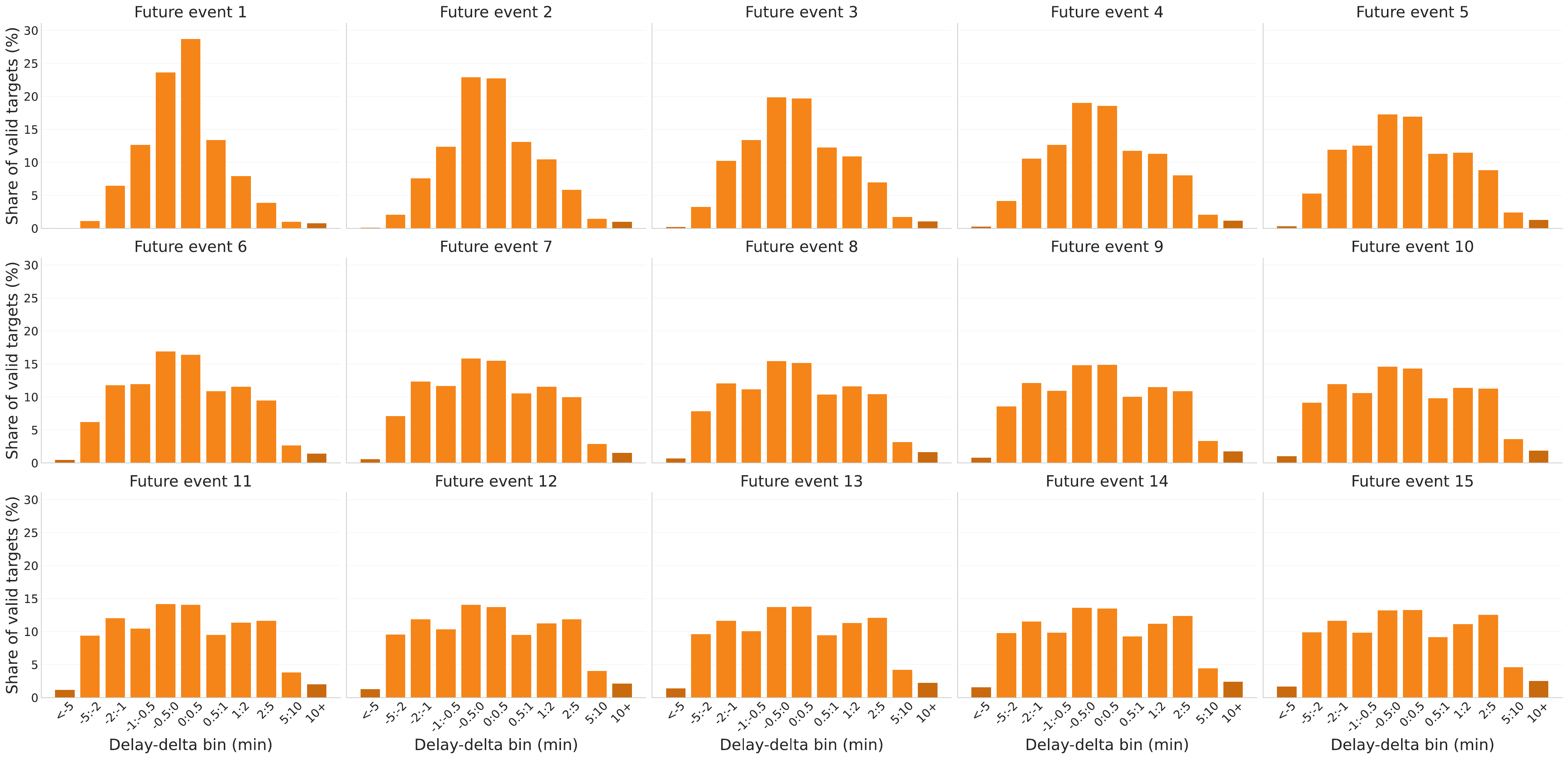}
    \caption{Distribution of valid targets across delay-delta bins broken down by future-event slot on the standard tier.}
    \label{fig:delay_delta_bin_counts_by_future_event}
\end{figure}

The outer bins are wider because extreme delay recovery and accumulation are much less frequent than small local changes around zero, as illustrated by the overall delay-delta bin counts (Figure~\ref{fig:overall_delay_delta_bin_counts}). Since these bins do not have equal width, this plot should not be interpreted as a density over delay delta, but only as a visualization of support across the evaluation bins. The per-target breakdown further shows that this support becomes progressively more spread out across later future-event slots, which is expected since trains have more time to recover or accumulate delay as the prediction target moves further ahead (Figure~\ref{fig:delay_delta_bin_counts_by_future_event}).

\subsection{Model Details}
\label{app:modeldetails}

\subsubsection{Translation}

We include a simple translation baseline, which is used in similar forms in operational railway settings \cite{arthaud2024transformers, wang2015data}. For every future target event $(s,i,j)$, it predicts
\[
\hat{d}_{s,i,j} = d_{s,i}^{\mathrm{last}},
\]
where $d_{s,i}^{\mathrm{last}}$ is defined in Section~\ref{sec:prediction-task}.
Despite its simplicity, this baseline provides a strong reference point for evaluating the added value of more complex models.

\subsubsection{Graph-event}

\begin{figure}[H]
\centering
\resizebox{0.9\textwidth}{!}{%
\begin{tikzpicture}[
  node distance=1.1cm,
  every node/.style={font=\small},
  box/.style={rectangle, rounded corners=4pt, draw, minimum width=5.5cm,
              minimum height=0.9cm, align=center, fill=gray!8},
  decision/.style={diamond, draw, aspect=2.2, align=center, fill=gray!8,
                   inner sep=2pt},
  terminal/.style={rectangle, rounded corners=12pt, draw, minimum width=2.8cm,
                   minimum height=0.8cm, align=center, fill=gray!15},
  highlight/.style={box, fill=blue!8},
  record/.style={box, minimum width=2.8cm, fill=teal!10},
  update/.style={box, fill=orange!10},
  arr/.style={-{Stealth[length=5pt]}, thick},
]

\node[terminal] (init) {Initialise priority queue};

\node[decision, below=of init] (empty) {Queue\\empty?};

\node[terminal, right=2.2cm of empty] (end) {End};

\node[highlight, below=of empty] (dequeue)
  {Dequeue train with earliest state finish time $t$};

\node[highlight, below=of dequeue] (consume)
  {Consume current state};

\node[decision, below=of consume] (reached) {New itinerary\\event reached?};

\node[record, right=2cm of reached] (record)
  {Record delay\\prediction};

\node[update, below=of reached] (update)
  {Construct next state\\
   \footnotesize(with finish time according to the simulation rules)};

\node[highlight, below=of update] (requeue)
  {Push state to priority queue};

\draw[arr] (init)     -- (empty);
\draw[arr] (empty)    -- node[right, xshift=2pt]{\textit{no}}  (dequeue);
\draw[arr] (dequeue)  -- (consume);
\draw[arr] (consume)  -- (reached);
\draw[arr] (reached)  -- node[right, xshift=2pt]{\textit{no}}  (update);
\draw[arr] (update)   -- (requeue);

\draw[arr] (empty) -- node[above]{\textit{yes}} (end);

\draw[arr] (reached) -- node[above]{\textit{yes}} (record);
\draw[arr] (record)  |- (update);

\draw[arr] (requeue.west)
  -- ++(-2.2,0)
  |- (empty.west);

\end{tikzpicture}
}
\caption{Illustration of the flowchart for the graph-event model.}
\label{fig:graph_event_flowchart}
\end{figure}
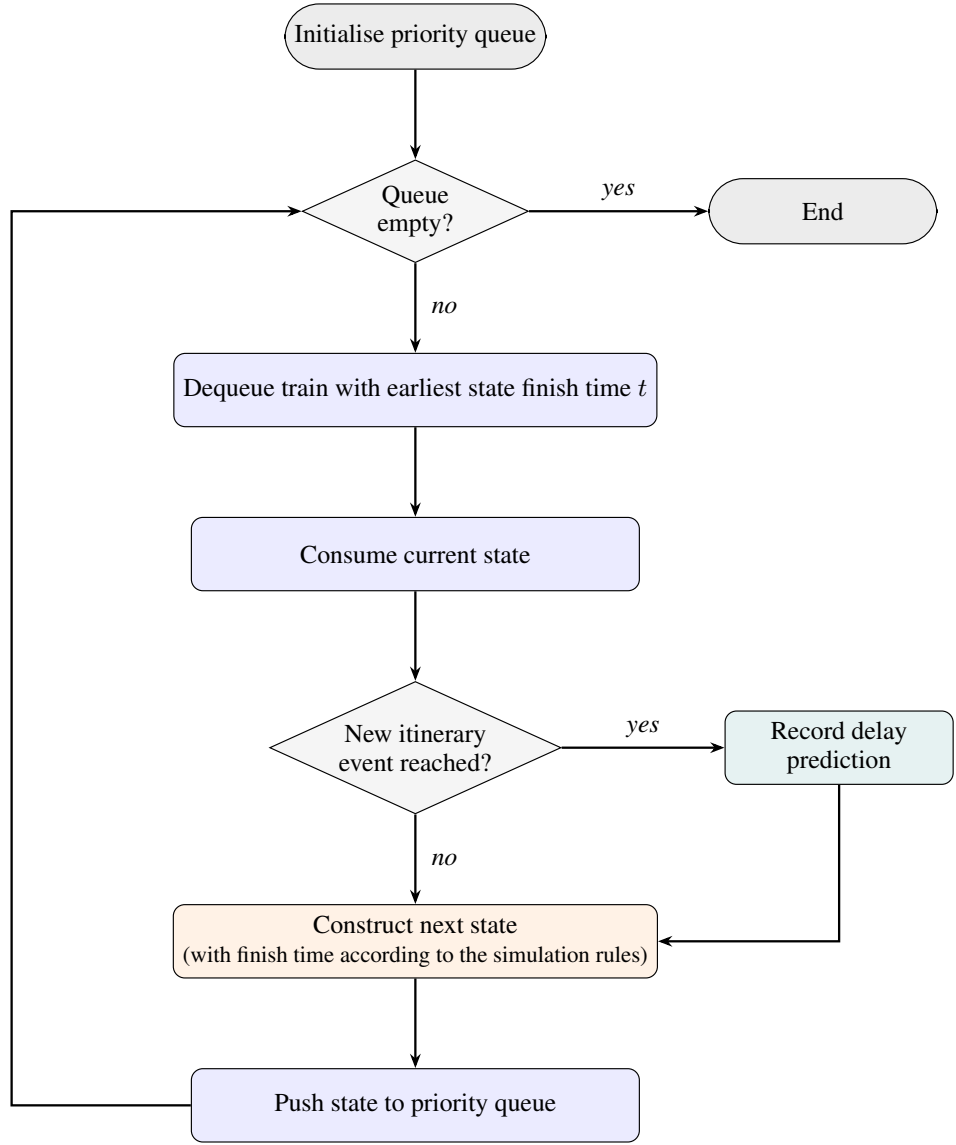

The graph-event model is a deterministic, non-learning reference model inspired by event-based train delay propagation approaches \cite{goverde2010delay, wei2015modeling}. Its input artifacts are described in Appendix~\ref{app:gold_graph_event_dataset}. Its purpose in the benchmark is to provide a graph-based model that explicitly simulates train movements and simple infrastructure interactions without relying on learned representations. Starting from each prediction snapshot, it propagates the currently active trains forward over the railway network using their current states, inferred paths, empirical travel-time statistics estimated from the training data, and a simple precedence constraint for trains sharing infrastructure. Future delays are then read out event by event from the resulting simulated evolution. Figure~\ref{fig:graph_event_flowchart} provides an overview of the full pipeline, including the initialization stage and the subsequent graph event processing.

\paragraph{Railway Network} We represent the railway network as a graph $G = (V, E)$, where each node $v \in V$ corresponds to an operational point and each edge $e \in E$ corresponds to a rail link between two operational points. Each edge $e$ is associated with a physical length $\ell(e) \in \mathbb{R}_{>0}$ measured in meters. For the simulation, we use a directed version of this graph in which each rail link is duplicated in both directions, allowing train movements and link occupancy constraints to be modeled directionally. This graph provides the spatial support on which the graph-event model simulates train movements. Figure~\ref{fig:full-network-visualization} illustrates the resulting network structure on the Belgian railway system. 

\paragraph{Train Itinerary} For a given train, we represent its itinerary as an ordered sequence of events
\[
\bigl((v_0,\tau_0,t^{\mathrm{sched}}_0),\,(v_1,\tau_1,t^{\mathrm{sched}}_1),\,\dots,\,(v_m,\tau_m,t^{\mathrm{sched}}_m)\bigr),
\]
where $v_k \in V$ is the operational point of event $k$, $\tau_k$ its event type (arrival, departure or passage), and $t^{\mathrm{sched}}_k$ its scheduled time. Consecutive itinerary events are not necessarily adjacent in the railway network, and even when two operational points are directly connected they may be linked by more than one rail edge due to potential parallel tracks. The model therefore uses the inferred path information stored in the \texttt{deduced\_paths} field of the \texttt{journeys} table to recover the exact ordered link-level route followed between successive events. More precisely, for each consecutive event pair $(k,k+1)$, we define a subpath
\[
P_k = (e_{k,1}, e_{k,2}, \dots, e_{k,r_k}),
\]
where each $e_{k,r} \in E$ and the ordered sequence $P_k$ connects $v_k$ to $v_{k+1}$ in the railway network. These subpaths form the link-level support on which the event-driven simulation is later carried out.

\paragraph{Snapshot Initialization.}

At a prediction snapshot with time $t^{\mathrm{snap}}$, each active train is initialized from its current itinerary position, while its initial delay state is set to the last known delay $d^{\mathrm{last}}$. More precisely, let $k$ denote the index of the most recent attained event before the snapshot. The initial delay of the train is set to $d^{\mathrm{last}}$. If $k<0$, the train is treated as being before its first event and initialized in a special departing state. If $k \geq m$, it is treated as having completed its route and initialized in a special arrived state. Otherwise, the train is initialized relative to the subpath $P_k$ connecting events $k$ and $k+1$.

The expected total traversal time of this event pair, denoted by $S_k$, is estimated from empirical event-pair travel-time statistics computed on the training split, keyed by the start and end operational points, event types, line information, and train relation. More precisely, the model uses the first finite positive estimate among the 0.40 quantile, the 0.45 quantile, and the median of the matching empirical travel-time statistics, in that order, and falls back to the scheduled travel time if no such estimate is available. To ensure that the train cannot reach event $k+1$ before its scheduled time, the expected total traversal time is lower-bounded by the remaining scheduled travel time measured from the heuristic start time of the event pair, i.e.
\[
S_k \leftarrow \max\!\left(S_k,\; t^{\mathrm{sched}}_{k+1} - \bigl(t^{\mathrm{sched}}_k + d^{\mathrm{last}}\bigr)\right).
\]

If $P_k$ is empty, the train is initialized in a stopped-at-station state, corresponding to a stop at node $v_k$ between the current event and the next one, with expected state duration $S_k$. Otherwise, under a constant-speed assumption, the total time $S_k$ is distributed across the directed links $(e_{k,1},\dots,e_{k,r_k})$ of the subpath proportionally to their lengths, i.e.
\[
S_{k,r} = S_k \frac{\ell(e_{k,r})}{\sum_{s=1}^{r_k} \ell(e_{k,s})},
\qquad r = 1,\dots,r_k.
\]
For the current event pair, we define the elapsed time since the predicted start of the event pair as
\[
\epsilon_k = t^{\mathrm{snap}} - \bigl(t^{\mathrm{sched}}_k + d^{\mathrm{last}}\bigr).
\]

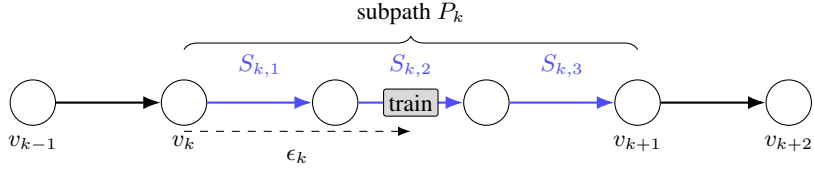
\begin{figure}[H]
\centering
\begin{tikzpicture}[>=Latex, font=\small]
  \foreach \x/\name in {0/n1,2/n2,4/n3,6/n4,8/n5,10/n6} {
    \node[circle, draw, fill=white, minimum size=6mm] (\name) at (\x,0) {};
  }

  \draw[->, thick] (n1) -- (n2);
  \draw[->, thick, blue!70] (n2) -- (n3);
  \draw[->, thick, blue!70] (n3) -- (n4);
  \draw[->, thick, blue!70] (n4) -- (n5);
  \draw[->, thick] (n5) -- (n6);

  \node[below=9pt] at (n1) {$v_{k-1}$};

  \node[below=9pt] at (n2) {$v_{k}$};

  \node[below=9pt] at (n5) {$v_{k+1}$};

  \node[below=9pt] at (n6) {$v_{k+2}$};

  \draw[decorate, decoration={brace, amplitude=5pt}]
    (2,0.7) -- (8,0.7)
    node[midway, above=6pt] {subpath $P_k$};

  \node[draw, fill=black!15, rounded corners=1pt, inner sep=2pt] (train) at (5,0) {train};

  \draw[->, dashed] ($(n2)+(0,-0.38)$) -- ($(n3)!0.5!(n4)+(0,-0.38)$);
  \node[below] at (3.5,-0.5) {$\epsilon_k$};

  \node[above=6pt, blue!70] at (5,0) {$S_{k,2}$};
  \node[above=6pt, blue!70] at (7,0) {$S_{k,3}$};

  \node[above=6pt, blue!70] at (3,0) {$S_{k,1}$};

\end{tikzpicture}
\caption{Illustration of snapshot initialization in the graph-event model. The train is initialized on the directed subpath $P_k$ between itinerary events $k$ and $k+1$. The elapsed time $\epsilon_k$ is used to infer the current position along the subpath, while the expected link traversal times $S_{k,1}, S_{k,2}, S_{k,3}$ distribute the total event-pair travel time over the directed links proportionally to their lengths.}
\label{fig:graph_event_init}
\end{figure}

The train is then initialized on the directed link reached after elapsed time $\epsilon_k$ along the subpath. Equivalently, this corresponds to the index $r$ such that
\[
\sum_{s=1}^{r-1} S_{k,s} \leq \epsilon_k < \sum_{s=1}^{r} S_{k,s},
\]
up to clipping at the ends of the subpath if needed, as illustrated in Figure~\ref{fig:graph_event_init}. In this way, each active train is initialized either at a station stop or on a directed rail link, together with its current delay and the remaining future itinerary needed for simulation. When several trains occupy the same directed link, their entry times determine their ordering on that link and therefore the order in which precedence constraints are applied during simulation.

At the end of this step, each active train is represented by its current itinerary event index, its current occupancy state (directed link, stopped-at-station, departing, or arrived), the expected finish time of that state, and its current delay.

\paragraph{Event-Driven Propagation.}
A key modeling assumption is that precedence constraints are enforced only on directed links, not on station states. In other words, trains are not allowed to overtake one another while traversing the same directed rail link, but they may effectively reorder while stopped at nodes between successive itinerary events. This keeps the interaction model simple and local to shared infrastructure occupancy.

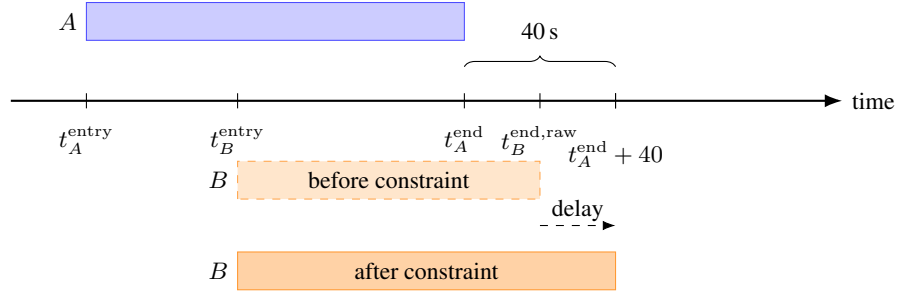
\begin{figure}[H]
\centering
\begin{tikzpicture}[>=Latex, font=\small]
  \def\xEntryA{1}
  \def\xEndA{6}
  \def\xEntryB{3}
  \def\xEndBRaw{7}
  \def\xGap{8}
  \def\xEndB{8}

  \def\yA{1.05}
  \def\yBRaw{-1.05}
  \def\yB{-2.25}
  \def\barHalfHeight{0.25}

  \draw[->, thick] (0,0) -- (11,0) node[right] {time};

  \foreach \x in {\xEntryA,\xEndA,\xEntryB,\xEndBRaw,\xGap,\xEndB} {
    \draw (\x,0.1) -- (\x,-0.1);
  }

  \node[below=6pt] at (\xEntryA,0) {$t^{\mathrm{entry}}_A$};
  \node[below=6pt] at (\xEndA,0) {$t^{\mathrm{end}}_A$};
  \node[below=6pt] at (\xEntryB,0) {$t^{\mathrm{entry}}_B$};
  \node[below=6pt] at (\xEndBRaw,0) {$t^{\mathrm{end,raw}}_B$};
  \node[below=6pt] at (\xGap,-0.2) {$t^{\mathrm{end}}_A + 40$};

  \draw[fill=blue!20, draw=blue!70]
    (\xEntryA,\yA-\barHalfHeight) rectangle (\xEndA,\yA+\barHalfHeight);
  \node[left] at (\xEntryA,\yA) {$A$};

  \draw[fill=orange!20, draw=orange!70, dashed]
    (\xEntryB,\yBRaw-\barHalfHeight) rectangle (\xEndBRaw,\yBRaw+\barHalfHeight);
  \node[left] at (\xEntryB,\yBRaw) {$B$};
  \node at ({(\xEntryB+\xEndBRaw)/2},\yBRaw) {before constraint};

  \draw[fill=orange!35, draw=orange!80]
    (\xEntryB,\yB-\barHalfHeight) rectangle (\xEndB,\yB+\barHalfHeight);
  \node[left] at (\xEntryB,\yB) {$B$};
  \node at ({(\xEntryB+\xEndB)/2},\yB) {after constraint};

  \draw[->, dashed] (\xEndBRaw,-1.65) -- (\xEndB,-1.65);
  \node[above=-2pt] at ({(\xEndBRaw+\xEndB)/2},-1.65) {delay};

  \draw[decorate, decoration={brace, amplitude=4pt}] (\xEndA,0.45) -- (\xGap,0.45);
  \node[above=7pt] at ({(\xEndA+\xGap)/2},0.45) {40\,s};
\end{tikzpicture}
\caption{Illustration of the precedence rule in the graph-event model. Train $A$ enters a directed link before train $B$. If the unconstrained finish time of $B$ occurs less than 40 seconds after the finish time of $A$, it is shifted forward so that $t^{\mathrm{end}}_B \geq t^{\mathrm{end}}_A + 40$.}
\label{fig:graph_event_precedence}
\end{figure}

For an event pair $(k,k+1)$, if the corresponding subpath $P_k$ is empty, the train occupies a stopped-at-station state with expected finish time
\[
t^{\mathrm{end}}_k = t^{\mathrm{entry}}_k + S_k.
\]
If instead the train occupies the directed link $e_{k,r}$ of a non-empty subpath, its expected finish time is
\[
t^{\mathrm{end}}_{k,r} = t^{\mathrm{entry}}_{k,r} + S_{k,r}.
\]
As illustrated in Figure~\ref{fig:graph_event_precedence}, when several trains occupy the same directed link, they are ordered by their entry times, and precedence is enforced in that order by replacing each predicted finish time with
\[
t^{\mathrm{end}}_{k,r} \leftarrow \max\!\bigl(t^{\mathrm{end}}_{k,r},\; t^{\mathrm{end}}_{\mathrm{prev}} + 40\bigr),
\]
where $t^{\mathrm{end}}_{\mathrm{prev}}$ denotes the finish time of the preceding train on the same directed link, thereby enforcing a fixed 40-second separation between successive trains on that link. The choice of a 40-second separation is based on empirical testing on the training data. These same rules are used both to build the initial ordered queues from the snapshot state and to insert new train states later during simulation.

The global evolution is implemented with a priority queue over predicted state-finish times. At each step, the train with smallest current state finish time $t$ is removed from its current occupancy state and its state is consumed. If the consumed state is the directed link $e_{k,r}$ with $r<r_k$, a new state is created for the next directed link $e_{k,r+1}$ of the same subpath, with entry time $t$. If the consumed state is the final directed link of the subpath, or a stopped-at-station state, the train reaches event $k+1$, its delay is updated to
\[
d_{k+1} = t - t^{\mathrm{sched}}_{k+1},
\]
and a new state is constructed toward the next itinerary event. In all cases, the resulting state is inserted into its corresponding occupancy queue, the precedence rule is applied if needed, and its predicted finish time is pushed back into the global priority queue. Whenever this transition increases the itinerary event index, the updated delay is recorded as the prediction for the corresponding future target event.

When a train reaches the end of its itinerary, it enters a terminal arrived state and is no longer reinserted into the priority queue. The simulation ends once this queue becomes empty.

\subsubsection{XGBoost}

\begin{figure}[H]
\centering
\resizebox{0.9\textwidth}{!}{%
\begin{tikzpicture}[
  every node/.style={font=\small},
  feat/.style={rectangle, draw=blue!28, line width=0.55pt, fill=blue!10,
               minimum width=5cm, minimum height=1.0cm, align=center},
  xgb/.style={rectangle, rounded corners=4pt, draw, fill=orange!10,
              minimum width=2.2cm, minimum height=1.4cm, align=center},
  outbox/.style={rectangle, draw=green!32!black, line width=0.55pt, fill=green!10,
                 minimum width=2.2cm, minimum height=0.95cm, align=center},
  arr/.style={-{Stealth[length=5pt]}, gray!70, thick},
  plain/.style={gray!60, thick},
]

\node[feat] (F) {tabular features\\$x$};

\node[xgb, below=1.4cm of F, xshift=-4.5cm] (X1)  {\textbf{XGBoost}\\model 1};
\node[xgb, right=0.5cm of X1]               (X2)  {\textbf{XGBoost}\\model 2};
\node[xgb, right=1.4cm of X2]               (Xn1) {\textbf{XGBoost}\\model $n{-}1$};
\node[xgb, right=0.5cm of Xn1]              (Xn)  {\textbf{XGBoost}\\model $n$};

\node at ($(X2)!0.5!(Xn1)$) {$\cdots$};

\coordinate (branch_y) at ($(X1.north)+(0,0.35)$);
\coordinate (branch_top) at ($(F.south |- branch_y)$);
\coordinate (branch_left) at ($(X1.north)+(0,0.35)$);
\coordinate (branch_right) at ($(Xn.north)+(0,0.35)$);

\draw[plain] (F.south) -- (branch_top);
\draw[plain] (branch_left) -- (branch_right);
\draw[plain] (branch_left) -- (X1.north);
\draw[plain] ($(X2.north)+(0,0.35)$) -- (X2.north);
\draw[plain] ($(Xn1.north)+(0,0.35)$) -- (Xn1.north);
\draw[plain] (branch_right) -- (Xn.north);

\node[outbox, below=0.8cm of X1]  (O1)  {event 1 delay\\$\hat y_1$};
\node[outbox, below=0.8cm of X2]  (O2)  {event 2 delay\\$\hat y_2$};
\node[outbox, below=0.8cm of Xn1] (On1) {event $n{-}1$ delay\\$\hat y_{n-1}$};
\node[outbox, below=0.8cm of Xn]  (On)  {event $n$ delay\\$\hat y_n$};

\draw[arr] (X1.south)  -- (O1.north);
\draw[arr] (X2.south)  -- (O2.north);
\draw[arr] (Xn1.south) -- (On1.north);
\draw[arr] (Xn.south)  -- (On.north);
\end{tikzpicture}
}
\caption{Illustration of the XGBoost architecture.}
\label{fig:xgboost_architecture}
\end{figure}
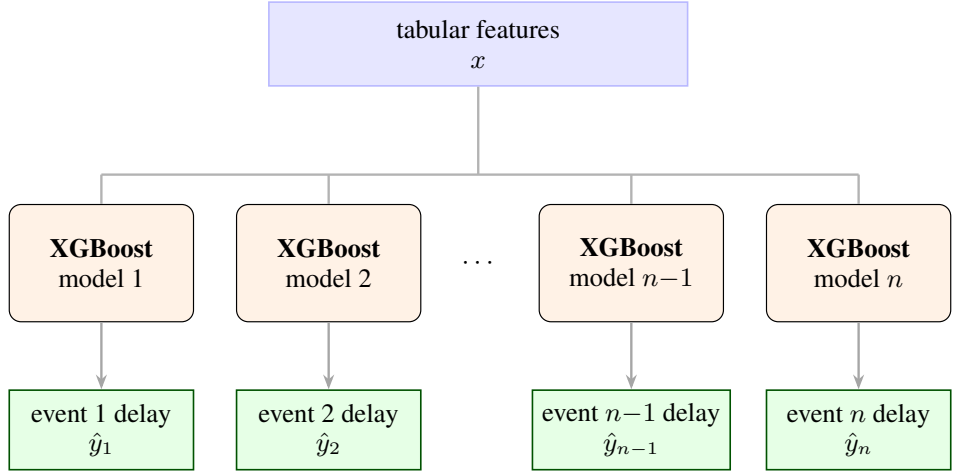

The XGBoost model uses the same tabular representation as the MLP, namely a single fixed-dimensional feature vector for each train-snapshot pair; this input data format is described in Appendix~\ref{app:gold_tabular_dataset}. As illustrated in Figure~\ref{fig:xgboost_architecture}, it trains one independent gradient-boosted tree regressor per prediction rather than predicting all targets jointly. Each regressor therefore maps the shared tabular input to the normalized residual future delay at one specific event horizon, and the final prediction vector is obtained by concatenating the outputs of the $n$ target-specific models.

\subsubsection{MLP}

\begin{figure}[H]
\centering
\resizebox{0.9\textwidth}{!}{%
\begin{tikzpicture}[every node/.style={font=\small}]
  \def\layersep{2.4}
  \def\vstep{1.15}
  \def\ydots{-3.45}
  \tikzset{
    indata/.style={circle, draw=blue!28, line width=0.55pt, fill=blue!10, minimum size=0.7cm, inner sep=0pt},
    hiddennode/.style={circle, draw, minimum size=0.7cm, inner sep=0pt, fill=violet!15},
    outdata/.style={circle, draw=green!32!black, line width=0.55pt, fill=green!10, minimum size=0.7cm, inner sep=0pt}
  }

  \node[indata] (I1) at (0,\ydots+\vstep) {$x_1$};
  \node (Id) at (0,\ydots) {$\vdots$};
  \node[indata] (Ik) at (0,\ydots-\vstep) {$x_d$};
  \node[align=center] at (0,-1.1)
    {\textbf{Input}\\[2pt]tabular features};

  \node[hiddennode] (H1a) at (\layersep,\ydots+2*\vstep) {};
  \node[hiddennode] (H1b) at (\layersep,\ydots+\vstep) {};
  \node (H1d) at (\layersep,\ydots) {$\vdots$};
  \node[hiddennode] (H1c) at (\layersep,\ydots-\vstep) {};
  \node[hiddennode] (H1d) at (\layersep,\ydots-2*\vstep) {};

  \node[hiddennode] (H2a) at (2*\layersep,\ydots+3*\vstep) {};
  \node[hiddennode] (H2b) at (2*\layersep,\ydots+2*\vstep) {};
  \node[hiddennode] (H2c) at (2*\layersep,\ydots+\vstep) {};
  \node (H2d) at (2*\layersep,\ydots) {$\vdots$};
  \node[hiddennode] (H2d) at (2*\layersep,\ydots-\vstep) {};
  \node[hiddennode] (H2e) at (2*\layersep,\ydots-2*\vstep) {};
  \node[hiddennode] (H2f) at (2*\layersep,\ydots-3*\vstep) {};
  \node[align=center] at (2*\layersep,1.1) {\textbf{Hidden layers}};

  \node[hiddennode] (H3a) at (3*\layersep,\ydots+2*\vstep) {};
  \node[hiddennode] (H3b) at (3*\layersep,\ydots+\vstep) {};
  \node (H3d) at (3*\layersep,\ydots) {$\vdots$};
  \node[hiddennode] (H3c) at (3*\layersep,\ydots-\vstep) {};
  \node[hiddennode] (H3d) at (3*\layersep,\ydots-2*\vstep) {};

  \node[outdata] (O1) at (4*\layersep,\ydots+\vstep) {$\hat y_1$};
  \node (Od) at (4*\layersep,\ydots) {$\vdots$};
  \node[outdata] (On) at (4*\layersep,\ydots-\vstep) {$\hat y_n$};
  \node[align=center] at (4*\layersep,-1.1) {\textbf{Output}};

  \node[anchor=west] at (10.0,\ydots+\vstep) {event 1 delay};
  \node[anchor=west] at (10.0,\ydots-\vstep) {event $n$ delay};

  \draw[gray!40, thin] (I1) -- (H1a);
  \draw[gray!40, thin] (I1) -- (H1b);
  \draw[gray!40, thin] (I1) -- (H1c);
  \draw[gray!40, thin] (I1) -- (H1d);

  \draw[gray!40, thin] (Ik) -- (H1a);
  \draw[gray!40, thin] (Ik) -- (H1b);
  \draw[gray!40, thin] (Ik) -- (H1c);
  \draw[gray!40, thin] (Ik) -- (H1d);

  \draw[gray!40, thin] (H1a) -- (H2a);
  \draw[gray!40, thin] (H1a) -- (H2b);
  \draw[gray!40, thin] (H1a) -- (H2c);
  \draw[gray!40, thin] (H1a) -- (H2d);
  \draw[gray!40, thin] (H1a) -- (H2e);
  \draw[gray!40, thin] (H1a) -- (H2f);

  \draw[gray!40, thin] (H1b) -- (H2a);
  \draw[gray!40, thin] (H1b) -- (H2b);
  \draw[gray!40, thin] (H1b) -- (H2c);
  \draw[gray!40, thin] (H1b) -- (H2d);
  \draw[gray!40, thin] (H1b) -- (H2e);
  \draw[gray!40, thin] (H1b) -- (H2f);

  \draw[gray!40, thin] (H1c) -- (H2a);
  \draw[gray!40, thin] (H1c) -- (H2b);
  \draw[gray!40, thin] (H1c) -- (H2c);
  \draw[gray!40, thin] (H1c) -- (H2d);
  \draw[gray!40, thin] (H1c) -- (H2e);
  \draw[gray!40, thin] (H1c) -- (H2f);

  \draw[gray!40, thin] (H1d) -- (H2a);
  \draw[gray!40, thin] (H1d) -- (H2b);
  \draw[gray!40, thin] (H1d) -- (H2c);
  \draw[gray!40, thin] (H1d) -- (H2d);
  \draw[gray!40, thin] (H1d) -- (H2e);
  \draw[gray!40, thin] (H1d) -- (H2f);

  \draw[gray!40, thin] (H2a) -- (H3a);
  \draw[gray!40, thin] (H2a) -- (H3b);
  \draw[gray!40, thin] (H2a) -- (H3c);
  \draw[gray!40, thin] (H2a) -- (H3d);

  \draw[gray!40, thin] (H2b) -- (H3a);
  \draw[gray!40, thin] (H2b) -- (H3b);
  \draw[gray!40, thin] (H2b) -- (H3c);
  \draw[gray!40, thin] (H2b) -- (H3d);  

  \draw[gray!40, thin] (H2c) -- (H3a);
  \draw[gray!40, thin] (H2c) -- (H3b);
  \draw[gray!40, thin] (H2c) -- (H3c);
  \draw[gray!40, thin] (H2c) -- (H3d);

  \draw[gray!40, thin] (H2d) -- (H3a);
  \draw[gray!40, thin] (H2d) -- (H3b);
  \draw[gray!40, thin] (H2d) -- (H3c);
  \draw[gray!40, thin] (H2d) -- (H3d);

  \draw[gray!40, thin] (H2e) -- (H3a);
  \draw[gray!40, thin] (H2e) -- (H3b);
  \draw[gray!40, thin] (H2e) -- (H3c);
  \draw[gray!40, thin] (H2e) -- (H3d);

  \draw[gray!40, thin] (H2f) -- (H3a);
  \draw[gray!40, thin] (H2f) -- (H3b);
  \draw[gray!40, thin] (H2f) -- (H3c);
  \draw[gray!40, thin] (H2f) -- (H3d);

  \draw[gray!40, thin] (H3a) -- (O1);
  \draw[gray!40, thin] (H3b) -- (O1);
  \draw[gray!40, thin] (H3c) -- (O1);
  \draw[gray!40, thin] (H3d) -- (O1);
  \draw[gray!40, thin] (H3a) -- (On);
  \draw[gray!40, thin] (H3b) -- (On);
  \draw[gray!40, thin] (H3c) -- (On);
  \draw[gray!40, thin] (H3d) -- (On);
\end{tikzpicture}
}
\caption{Illustration of the MLP architecture.}
\label{fig:mlp_architecture}
\end{figure}
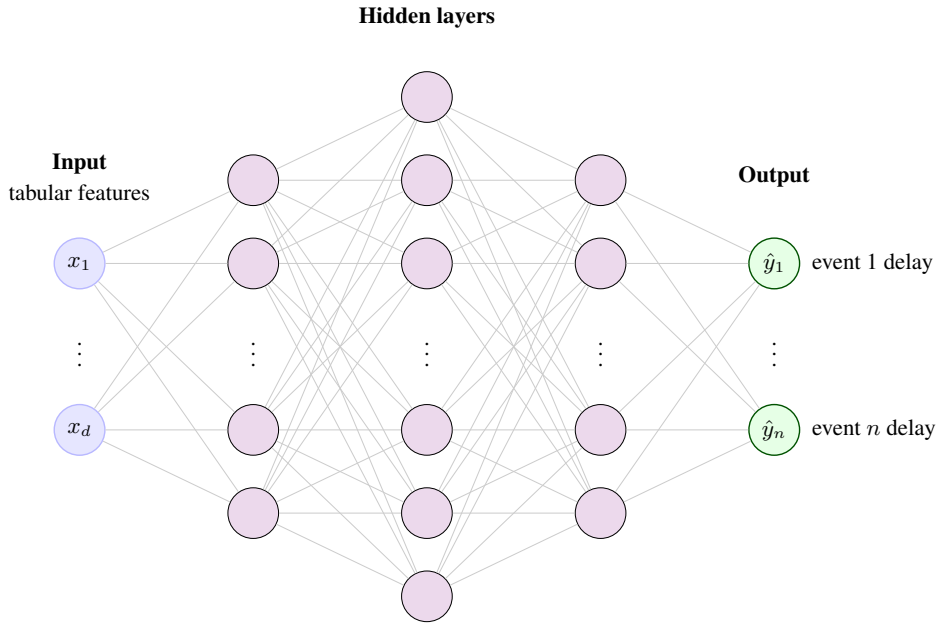

The MLP model operates on the tabular representation, where all input features for a train at a given snapshot are concatenated into a single fixed-dimensional vector; this input data format is described in Appendix~\ref{app:gold_tabular_dataset}. As illustrated in Figure~\ref{fig:mlp_architecture}, the model is a standard feed-forward multilayer perceptron composed of a stack of fully connected hidden layers with ReLU activations and dropout regularization. Its output layer has one scalar neuron per prediction target, so that the model directly predicts one normalized residual future-delay target for each of the next $n$ events in a single forward pass.

\subsubsection{LSTM}

\definecolor{tokenblue}{RGB}{230,241,251}
\definecolor{layerpurple}{RGB}{238,237,254}
\definecolor{headcoral}{RGB}{250,236,231}

\begin{figure}[H]
\centering
\resizebox{0.8\textwidth}{!}{%
\begin{tikzpicture}[
  node distance=0.95cm and 1.4cm,
  every node/.style={font=\small},
  inputbox/.style={rectangle, draw=blue!28, line width=0.55pt,
                   fill=tokenblue!85, minimum width=2.75cm, minimum height=1.0cm,
                   align=center},
  widebox/.style={rectangle, rounded corners=4pt, draw=purple!48, line width=0.9pt,
                  fill=layerpurple, minimum width=3.1cm, minimum height=0.95cm,
                  align=center},
  smallbox/.style={rectangle, rounded corners=4pt, draw=orange!42, line width=0.9pt,
                   fill=headcoral, minimum width=1.3cm, minimum height=0.72cm,
                   align=center},
  thinbox/.style={rectangle, rounded corners=4pt, draw=orange!42, line width=0.9pt,
                  fill=headcoral, minimum width=2.3cm, minimum height=0.64cm,
                  align=center},
  mlpbox/.style={rectangle, rounded corners=4pt, draw=red!55!black, line width=0.95pt,
                 fill=red!10, minimum width=2.7cm, minimum height=0.95cm,
                 align=center},
  mergebox/.style={rectangle, rounded corners=4pt, draw=yellow!50!brown,
                   dashed, line width=0.85pt, fill=yellow!18, minimum width=1.95cm, minimum height=0.60cm,
                   align=center},
  outbox/.style={rectangle, draw=green!32!black, line width=0.55pt,
                 fill=green!10, minimum width=2.8cm, minimum height=1.0cm,
                 align=center},
  arr/.style={-{Stealth[length=4pt]}, gray!60, thin},
  lab/.style={font=\small, text=black, align=center}
]

\node[mlpbox] (mlp) {MLP};
\node[widebox, right=2.8cm of mlp] (enc) {LSTM encoder};

\node[thinbox, below=0.15cm of enc.south west, anchor=north west, xshift=-2.5cm] (att) {Attention pooling};

\node[widebox, below=3.1cm of enc] (dec) {LSTM decoder};
\node[smallbox] (hproj) at ($(enc.south)!0.5!(dec.north) + (-0.85cm,0)$) {Linear\\proj.};
\node[smallbox, right=0.45cm of hproj] (cproj) {Linear\\proj.};
\node[mlpbox, below=0.85cm of dec] (head) {shared MLP\\per step};

\node[inputbox, above=0.7cm of mlp] (staticlab) {static features\\$x^s$};
\node[inputbox] (pastlab) at (enc |- staticlab) {past events\\sequence\\$x_1^p, \ldots, x_k^p$};

\node[outbox, below=0.75cm of head] (outlab) {delays\\$\hat y_1, \ldots, \hat y_n$};

\coordinate (encLeftMid) at (enc.west);
\coordinate (encToAttX) at (att.north |- encLeftMid);

\node[mergebox] (expandbox) at ($(mlp)!0.5!(att) + (0,-1.65cm)$) {concat + expand};
\node[mergebox] (concatbox) at (expandbox |- dec.west) {concat};
\node[inputbox, left=0.55cm of concatbox] (futurelab) {future event sequence\\$x_1^f, \ldots, x_n^f$};
\coordinate (ctxdrop) at (expandbox.south |- concatbox.north);

\draw[arr] (staticlab.south) -- (mlp.north);
\draw[arr] (mlp.south) -- ++(0,-0.35)
  node[pos=0.5, left, font=\scriptsize, text=black] {$c^s$}
  |- (expandbox.west);

\draw[arr] (pastlab.south) -- (enc.north);
\draw[arr] (encLeftMid) -- (encToAttX)
  node[pos=0.62, above, font=\scriptsize, text=black] {$h_1^p, \ldots, h_k^p$}
  -- (att.north);

\coordinate (hstart) at (hproj.north |- enc.south);
\coordinate (cstart) at (cproj.north |- enc.south);

\draw[arr] (hstart) -- node[midway, right, font=\scriptsize, text=black] {$h_k^p$} (hproj.north);
\draw[arr] (cstart) -- node[midway, right, font=\scriptsize, text=black] {$c_k^p$} (cproj.north);

\draw[arr] (att.south) -- ++(0,-0.35)
  node[pos=0.5, right, font=\scriptsize, text=black] {$c^p$}
  |- (expandbox.east);

\draw[arr] (expandbox.south) -- node[midway, right, font=\scriptsize, text=black] {$c_1, \ldots, c_n$} (concatbox.north);

\draw[arr] (futurelab.east) -- (concatbox.west);

\draw[arr] (concatbox.east) -- node[midway, above, font=\scriptsize, text=black] {$c'_1, \ldots, c'_n$} (dec.west);

\coordinate (hdecin) at (hproj.south |- dec.north);
\coordinate (cdecin) at (cproj.south |- dec.north);

\draw[arr] (hproj.south) -- node[midway, right, font=\scriptsize, text=black] {$h_0^f$} (hdecin);
\draw[arr] (cproj.south) -- node[midway, right, font=\scriptsize, text=black] {$c_0^f$} (cdecin);

\draw[arr] (dec.south) -- node[right, font=\scriptsize, text=black] {$h_1^f, \ldots, h_n^f$} (head.north);
\draw[arr] (head.south) -- (outlab.north);

\end{tikzpicture}
}
\caption{Illustration of the LSTM sequence-to-sequence architecture.}
\label{fig:lstm_architecture}
\end{figure}
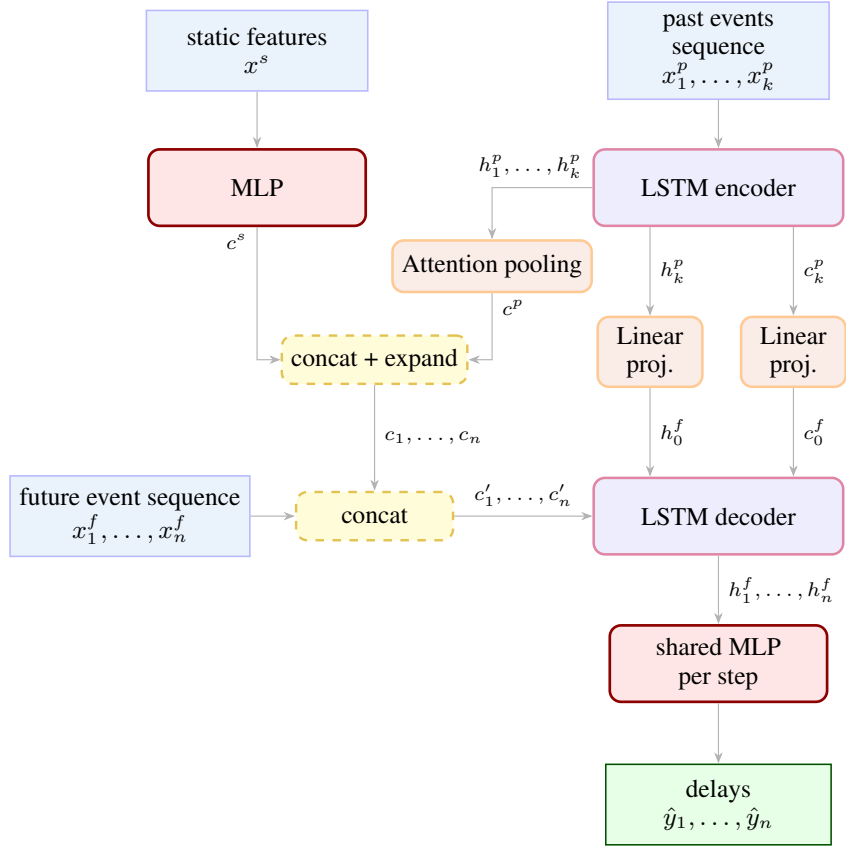

Our prediction setting naturally has a sequential structure: each instance contains a past event sequence describing the recent observed progression of a train up to the prediction snapshot, together with a future event sequence corresponding to the downstream events of the same train for which delays must be predicted. The sequential input data format used by this model is described in Appendix~\ref{app:gold_sequential_dataset}. This makes recurrent architectures a relevant family to consider, in line with prior work on train delay prediction using LSTMs \cite{huang2020modeling, yu2023delay}. However, existing LSTM-based formulations do not directly match our snapshot-based benchmark setting, which requires predicting the next sequence of scheduled events for each active train. We therefore adopt a tailored encoder-decoder LSTM sequence-to-sequence architecture for this model, allowing the model to encode the recent history of the train and decode it into successive delay predictions along its future itinerary.

The input is organized into three feature families: static features, past event features, and future event features. Static features encode train- and snapshot-level context together with local rail-context and weather information. Past event features encode the observed train history through schedule offsets, observed delays, event types, and station embeddings. Future event features encode the known scheduled structure of the upcoming itinerary through schedule offsets, event types, and station embeddings. In the benchmark, these inputs are instantiated over a fixed context of past and future events and a fixed look-ahead over upcoming rail links.

As illustrated in Figure~\ref{fig:lstm_architecture}, the model processes the three feature families through dedicated components. First, the past event sequence $(x_1^p, \ldots, x_k^p)$ is fed to an LSTM encoder, producing hidden states $(h_1^p, \ldots, h_k^p)$ together with final hidden and cell states $(h_k^p, c_k^p)$. A learned attention-pooling module then scores the encoder hidden states and forms a context vector $c^p$ as a weighted sum over the past sequence, allowing the decoder to focus on the most relevant parts of the observed history rather than relying only on the encoder's last state. Second, the static features $x^s$ are passed through a small MLP to obtain a static embedding $c^s$. The two context sources are concatenated and then expanded across the prediction horizon to form a repeated context sequence $(c_1, \ldots, c_n)$, so that each future decoding step has access to the same global information about the train and its recent history. Third, each future target step contributes its future known features $(x_1^f, \ldots, x_n^f)$, and the model appends the normalized position of the step within the prediction horizon. These are concatenated with the repeated context sequence to form decoder inputs $(c'_1, \ldots, c'_n)$.

The decoder is itself an LSTM initialized from transformed versions of the encoder final states: linear projection layers map $(h_k^p, c_k^p)$ to $(h_0^f, c_0^f)$ before decoding begins. This bridge lets the decoder start from a representation shaped by the observed past while still allowing a learned adaptation between encoder and decoder dynamics. The decoder then processes the full future input sequence and returns hidden states $(h_1^f, \ldots, h_n^f)$. Finally, a shared step-wise MLP is applied independently to each decoder state $h_i^f$ to predict the corresponding delay $\hat y_i$ for future event $i$. In this way, the model combines sequential memory from the past, static train-level context, and known future covariates to produce a coherent sequence of downstream delay predictions.

\subsubsection{Transformer}

\definecolor{attnpurple}{RGB}{127,119,221}
\definecolor{divcolor}{RGB}{153,60,29}

\begin{figure}[H]
\centering
\resizebox{\textwidth}{!}{%
\begin{tikzpicture}[
  every node/.style={font=\small},
  token/.style={rectangle, draw=blue!40,
                fill=tokenblue, minimum width=1.65cm, minimum height=1cm, align=center},
  layer/.style={rectangle, rounded corners=4pt, draw=purple!40,
                fill=layerpurple, minimum width=11.5cm, minimum height=1.1cm, align=center},
  head/.style={rectangle, rounded corners=4pt, draw=orange!35,
               fill=headcoral, minimum width=11.5cm, minimum height=0.62cm, align=center},
  projbox/.style={rectangle, rounded corners=4pt, draw=orange!35,
                  fill=headcoral, minimum width=1.65cm, minimum height=0.72cm, align=center},
  predbox/.style={rectangle, rounded corners=4pt, draw=orange!35,
                  fill=headcoral, minimum width=1.65cm, minimum height=0.72cm, align=center},
  outgreen/.style={rectangle, draw=green!32!black, line width=0.55pt,
                   fill=green!10, minimum width=1.9cm, minimum height=0.82cm, align=center},
  arr/.style={-{Stealth[length=4pt]}, gray!60, thin},
  stem/.style={gray!45, thin},
]

\node[token] (T1)  {\textbf{train 1}\\{\scriptsize tab features}\\{\scriptsize $x^{(1)}$}};
\node[token, right=0.95cm of T1]  (T2)  {\textbf{train 2}\\{\scriptsize tab features}\\{\scriptsize $x^{(2)}$}};
\node[right=0.6cm of T2]          (Td)  {$\cdots$};
\node[token, right=0.6cm of Td]   (Tn1) {\textbf{train $m{-}1$}\\{\scriptsize tab features}\\{\scriptsize $x^{(m-1)}$}};
\node[token, right=0.95cm of Tn1] (Tn)  {\textbf{train $m$}\\{\scriptsize tab features}\\{\scriptsize $x^{(m)}$}};

\node[layer, below=2.15cm of $(T2)!0.5!(Tn1)$] (L1) {\textbf{Encoder layer 1}};

\coordinate (l1b1)  at (T1  |- L1.south);  \coordinate (l1t1)  at (T1  |- L1.north);
\coordinate (l1b2)  at (T2  |- L1.south);  \coordinate (l1t2)  at (T2  |- L1.north);
\coordinate (l1bn1) at (Tn1 |- L1.south);  \coordinate (l1tn1) at (Tn1 |- L1.north);
\coordinate (l1bn)  at (Tn  |- L1.south);  \coordinate (l1tn)  at (Tn  |- L1.north);

\node[projbox] (IP1)  at ($(T1.south)!0.5!(l1t1)$)  {\scriptsize Linear proj. $f_\theta$};
\node[projbox] (IP2)  at ($(T2.south)!0.5!(l1t2)$)  {\scriptsize Linear proj. $f_\theta$};
\node[projbox] (IPn1) at ($(Tn1.south)!0.5!(l1tn1)$) {\scriptsize Linear proj. $f_\theta$};
\node[projbox] (IPn)  at ($(Tn.south)!0.5!(l1tn)$)  {\scriptsize Linear proj. $f_\theta$};
\node (IPd) at ($(IP2)!0.5!(IPn1)$) {$\cdots$};

\draw[stem] (T1.south)  -- (IP1.north);
\draw[stem] (IP1.south) -- (l1t1);

\draw[stem] (T2.south)  -- (IP2.north);
\draw[stem] (IP2.south) -- (l1t2);

\draw[stem] (Tn1.south)  -- (IPn1.north);
\draw[stem] (IPn1.south) -- (l1tn1);

\draw[stem] (Tn.south)  -- (IPn.north);
\draw[stem] (IPn.south) -- (l1tn);

\node[below=0.28cm of L1] (vd) {$\vdots$};

\node[layer, below=0.28cm of vd] (LL) {\textbf{Encoder layer L}};

\coordinate (llb1)  at (T1  |- LL.south);  \coordinate (llt1)  at (T1  |- LL.north);
\coordinate (llb2)  at (T2  |- LL.south);  \coordinate (llt2)  at (T2  |- LL.north);
\coordinate (llbn1) at (Tn1 |- LL.south);  \coordinate (lltn1) at (Tn1 |- LL.north);
\coordinate (llbn)  at (Tn  |- LL.south);  \coordinate (lltn)  at (Tn  |- LL.north);

\foreach \a/\b in {l1b1/llt1, l1b2/llt2, l1bn1/lltn1, l1bn/lltn}
  \draw[stem] (\a) -- (\b);

\node[predbox] (PH1)  at ($(llb1)+(0,-0.95)$)  {\scriptsize Linear proj. $g_\phi$};
\node[predbox] (PH2)  at ($(llb2)+(0,-0.95)$)  {\scriptsize Linear proj. $g_\phi$};
\node[predbox] (PHn1) at ($(llbn1)+(0,-0.95)$) {\scriptsize Linear proj. $g_\phi$};
\node[predbox] (PHn)  at ($(llbn)+(0,-0.95)$)  {\scriptsize Linear proj. $g_\phi$};
\node (PHd)  at ($(PH2)!0.5!(PHn1)$) {$\cdots$};

\draw[stem] (llb1)  -- (PH1.north);
\draw[stem] (llb2)  -- (PH2.north);
\draw[stem] (llbn1) -- (PHn1.north);
\draw[stem] (llbn)  -- (PHn.north);

\node[outgreen] (OUT1)  at ($(PH1.south)+(0,-0.78)$)  {\scriptsize train 1 delays\\{\scriptsize $\hat y_1^{(1)}, \ldots, \hat y_n^{(1)}$}};
\node[outgreen] (OUT2)  at ($(PH2.south)+(0,-0.78)$)  {\scriptsize train 2 delays\\{\scriptsize $\hat y_1^{(2)}, \ldots, \hat y_n^{(2)}$}};
\node[outgreen] (OUTn1) at ($(PHn1.south)+(0,-0.78)$) {\scriptsize train $m{-}1$ delays\\{\scriptsize $\hat y_1^{(m-1)}, \ldots, \hat y_n^{(m-1)}$}};
\node[outgreen] (OUTn)  at ($(PHn.south)+(0,-0.78)$)  {\scriptsize train $m$ delays\\{\scriptsize $\hat y_1^{(m)}, \ldots, \hat y_n^{(m)}$}};
\node at ($(OUT2)!0.5!(OUTn1)$) {$\cdots$};

\draw[stem] (PH1.south)  -- (OUT1.north);
\draw[stem] (PH2.south)  -- (OUT2.north);
\draw[stem] (PHn1.south) -- (OUTn1.north);
\draw[stem] (PHn.south)  -- (OUTn.north);

\end{tikzpicture}
}
\caption{Illustration of the Transformer architecture.}
\label{fig:transformer_architecture}
\end{figure}
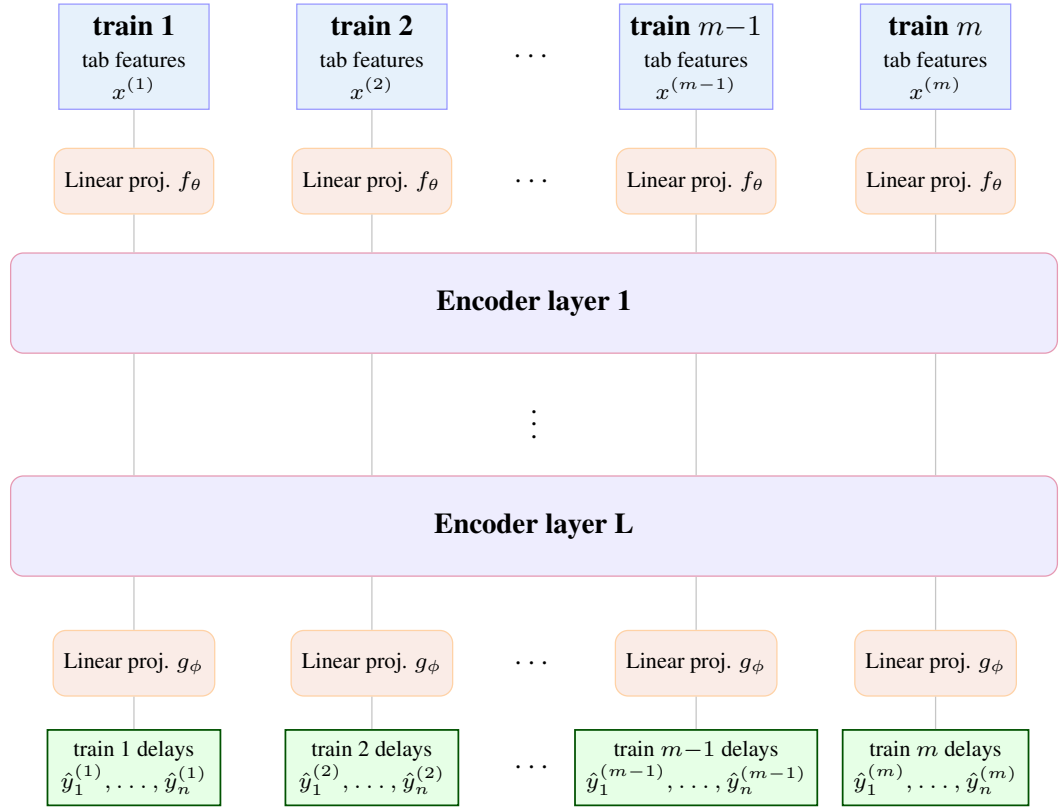

The Transformer model from \cite{arthaud2024transformers} operates on the set of active trains present in a prediction snapshot. It uses the same tabular feature representation as the MLP and XGBoost models, whose input data format is described in Appendix~\ref{app:gold_tabular_dataset}, but instead of processing each train independently, it groups all train feature vectors from a given snapshot into a variable-length sequence. As illustrated in Figure~\ref{fig:transformer_architecture}, each train vector is first projected into a shared embedding space and then processed by a stack of Transformer encoder layers with self-attention, ReLU feed-forward blocks, and dropout regularization, allowing every train representation to attend to all others in the snapshot. A shared linear projection head is applied independently to the contextualized representation of each train, producing one normalized residual future-delay target per future event.

\subsubsection{GNN}

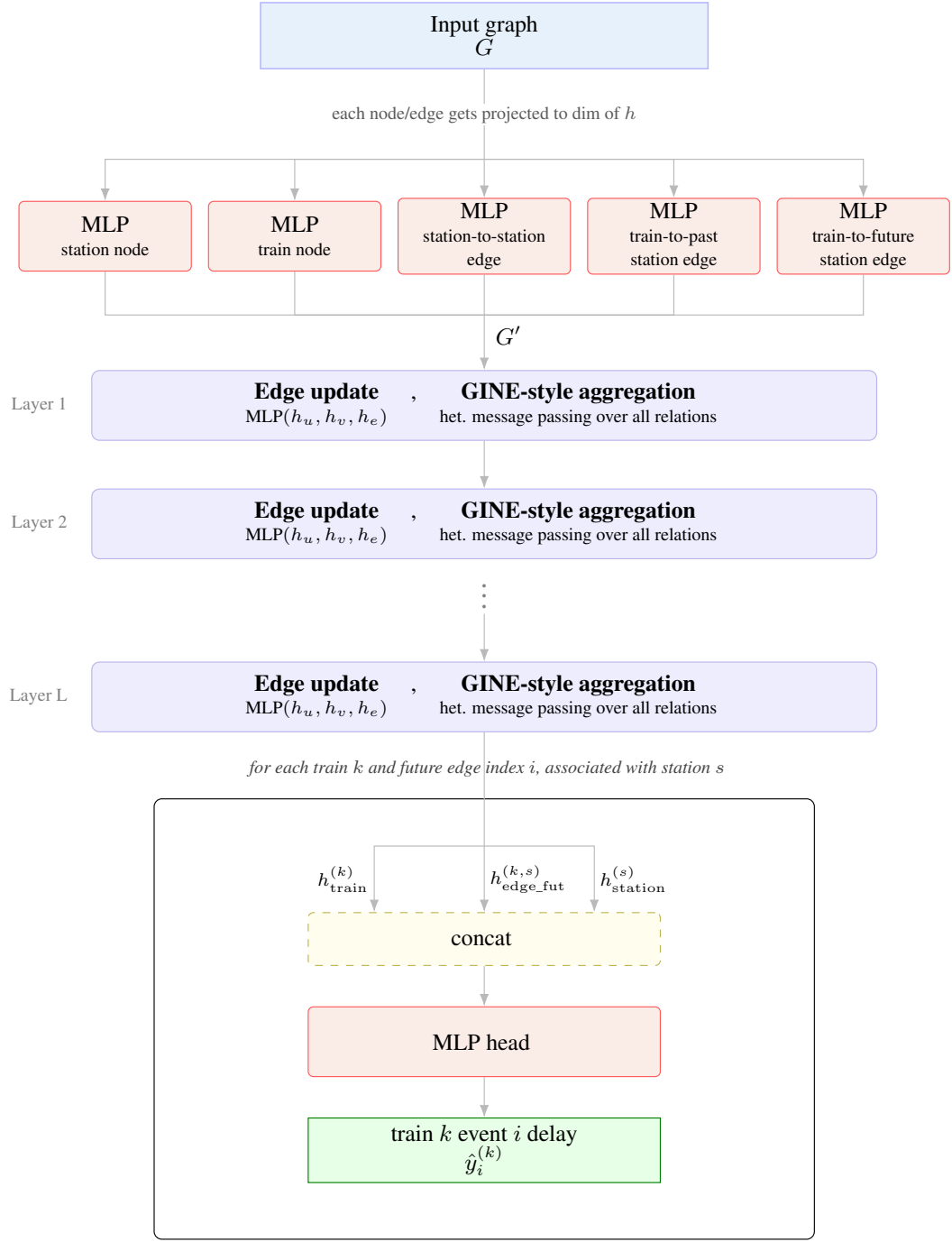
\begin{figure}[H]
\centering
\resizebox{\textwidth}{!}{%
\begin{tikzpicture}[
    font=\small,
    >=Latex,
    line/.style={draw=gray!55, -Latex},
    inputbox/.style={rectangle, draw=blue!28, line width=0.55pt,
                     fill=tokenblue, minimum width=6.1cm, minimum height=0.9cm,
                     align=center},
    mlp/.style={rectangle, rounded corners=2pt, draw=red!65, fill=headcoral,
                minimum width=3.1cm, minimum height=0.95cm, align=center},
    projmlp/.style={rectangle, rounded corners=2pt, draw=red!65, fill=headcoral,
                    minimum width=2.35cm, minimum height=0.95cm, align=center},
    edgeupd/.style={rectangle, rounded corners=2pt, draw=orange!70, fill=headcoral,
                    minimum width=4.1cm, minimum height=0.78cm, align=center},
    agg/.style={rectangle, rounded corners=2pt, draw=attnpurple!60, fill=layerpurple,
                minimum width=5.3cm, minimum height=0.78cm, align=center},
    layerblock/.style={rectangle, rounded corners=3pt, draw=attnpurple!55,
                       fill=layerpurple, minimum width=10.7cm, minimum height=0.95cm,
                       align=center},
    rep/.style={rectangle, rounded corners=2pt, draw=orange!65, fill=headcoral,
                minimum width=3.1cm, minimum height=0.9cm, align=center},
    repblue/.style={rectangle, rounded corners=2pt, draw=blue!35, fill=tokenblue,
                    minimum width=3.1cm, minimum height=0.9cm, align=center},
    outbox/.style={rectangle, draw=green!50!black, fill=green!10,
                   minimum width=4.8cm, minimum height=0.88cm, align=center},
    note/.style={draw=gray!25, fill=gray!8, rounded corners=2pt, align=center},
    every node/.style={inner sep=2pt}
]

\node[inputbox] (G) at (0,0) {
    Input graph\\[-1pt] $G$
};

\node[projmlp, below=1.75cm of G] (m3) {
    MLP\\[-1pt] \scriptsize station-to-station\\[-1pt] \scriptsize edge
};
\node[projmlp, left=0.22cm of m3] (m2) {
    MLP\\[-1pt] \scriptsize train node
};
\node[projmlp, left=0.22cm of m2] (m1) {
    MLP\\[-1pt] \scriptsize station node
};
\node[projmlp, right=0.22cm of m3] (m4) {
    MLP\\[-1pt] \scriptsize train-to-past\\[-1pt] \scriptsize station edge
};
\node[projmlp, right=0.22cm of m4] (m5) {
    MLP\\[-1pt] \scriptsize train-to-future\\[-1pt] \scriptsize station edge
};

\node[inner sep=0pt, below=1.2cm of G] (inputsplit) {};
\draw[gray!55] (G.south) -- node[midway, fill=white, inner sep=1pt, font=\scriptsize, text=black!65, align=center] {each node/edge gets projected to dim of $h$} (inputsplit.center);
\draw[line] (inputsplit.center) -| (m1.north);
\draw[line] (inputsplit.center) -| (m2.north);
\draw[line] (inputsplit.center) -- (m3.north);
\draw[line] (inputsplit.center) -| (m4.north);
\draw[line] (inputsplit.center) -| (m5.north);

\coordinate (gmerge) at ($(m3.south)+(0,-0.55cm)$);
\draw[gray!55] (m1.south) |- (gmerge);
\draw[gray!55] (m2.south) |- (gmerge);
\draw[gray!55] (m3.south) -- (gmerge);
\draw[gray!55] (m4.south) |- (gmerge);
\draw[gray!55] (m5.south) |- (gmerge);
\node[anchor=west] (gprime) at ($(gmerge)+(0.08cm,-0.28cm)$) {$G'$};

\node[layerblock, below=0.75cm of gmerge] (l1) {
    \begin{tabular}{c@{\quad}c@{\quad}c}
    \textbf{Edge update} & , & \textbf{GINE-style aggregation}\\[-1pt]
    \scriptsize MLP$(h_u,h_v,h_e)$ & & \scriptsize het. message passing over all relations
    \end{tabular}
};
\node[font=\scriptsize, gray, left=0.25cm of l1.west, anchor=east] {Layer 1};
\draw[line] (gmerge) -- (l1.north);

\node[layerblock, below=0.65cm of l1] (l2) {
    \begin{tabular}{c@{\quad}c@{\quad}c}
    \textbf{Edge update} & , & \textbf{GINE-style aggregation}\\[-1pt]
    \scriptsize MLP$(h_u,h_v,h_e)$ & & \scriptsize het. message passing over all relations
    \end{tabular}
};
\node[font=\scriptsize, gray, left=0.25cm of l2.west, anchor=east] {Layer 2};
\draw[line] (l1.south) -- (l2.north);

\node[font=\large, gray, below=0.05cm of l2] (dots) {$\vdots$};

\node[layerblock, below=0.65cm of dots] (lL) {
    \begin{tabular}{c@{\quad}c@{\quad}c}
    \textbf{Edge update} & , & \textbf{GINE-style aggregation}\\[-1pt]
    \scriptsize MLP$(h_u,h_v,h_e)$ & & \scriptsize het. message passing over all relations
    \end{tabular}
};
\node[font=\scriptsize, gray, left=0.25cm of lL.west, anchor=east] {Layer L};
\draw[line] (dots) -- (lL.north);

\def\catdrop{2.45cm}
\def\boxdrop{0.90cm}
\def\boxwidth{9.0cm}
\def\boxheight{6.0cm}
\def\splitDrop{1.55cm}
\def\labelgap{0.15cm}

\node[draw=yellow!70!black, dashed, fill=yellow!8, rounded corners=2pt,
      minimum width=4.8cm, minimum height=0.72cm,
        below=\catdrop of lL, align=center] (cat) {
    concat
};

\node[mlp, minimum width=4.8cm, below=0.55cm of cat] (head) {
    MLP head
};
\node[outbox, below=0.55cm of head] (delay) {
    train $k$ event $i$ delay\\[-1pt]
    $\hat y_i^{(k)}$
};

\draw[line] (cat) -- (head);
\draw[line] (head) -- (delay);

\node[draw=black, rounded corners=3pt, minimum width=\boxwidth,
    minimum height=\boxheight, anchor=north] (readoutbox) at ($(lL.south)+(0,-\boxdrop)$) {};
\node[above=\labelgap of readoutbox.north, font=\scriptsize\itshape, black!75, align=center, xshift=0.2em, yshift=0.2em] (for) {
    for each train $k$ and future edge index $i$, associated with station $s$
};

\coordinate (readsplit) at ($(lL.south)+(0,-\splitDrop)$);
\coordinate (trainend) at ($(cat.north)+(-1.5cm,0)$);
\coordinate (stationend) at (cat.north);
\coordinate (edgeend) at ($(cat.north)+(1.5cm,0)$);
\draw[draw=gray!55] (lL.south) -- (readsplit);
\draw[line] (readsplit) -| node[pos=0.75, left, font=\scriptsize, text=black] {$h_{\rm train}^{(k)}$} (trainend);
\draw[line] (readsplit) -- node[pos=0.5, right, font=\scriptsize, text=black] {$h_{\rm edge\_fut}^{(k,s)}$} (stationend);
\draw[line] (readsplit) -| node[pos=0.75, right, font=\scriptsize, text=black] {$h_{\rm station}^{(s)}$} (edgeend);

\end{tikzpicture}
}
\caption{Illustration of the GNN architecture.}
\label{fig:gnn_architecture}
\end{figure}

In our benchmark, each prediction snapshot can be viewed not only as a collection of train-level features, but as a structured system of interactions between active trains, stations, and railway links. This makes graph neural networks a natural model family for RIDE, as they can operate directly on relational structure rather than relying only on flattened or sequentialized representations. Following recent graph-based approaches to train delay prediction \cite{li2024railway, huang2024explainable}, we design a heterogeneous GNN adapted to our snapshot-based forecasting setting. In particular, existing graph formulations do not directly match our task, which requires predicting delays for the next sequence of scheduled events of each active train within a shared network snapshot. We therefore introduce a graph construction and prediction setup tailored to this benchmark.

The input graph representation is described in Appendix~\ref{app:gold_gnn_dataset} and illustrated in Figures~\ref{fig:gnn_toy} and~\ref{fig:gnn_snapshot_country}. The GNN consumes this heterogeneous snapshot graph with train and station nodes, station-to-station edges, train-to-station edges for past and future events, and the reverse relations provided by the dataset. Future delay residuals are not included in the future-edge input attributes; they are stored as edge targets and used only to compute the training loss. The normalized future-event rank remains an input feature, while the restored rank metadata is used to align edge predictions with evaluation horizons.

The GNN architecture is illustrated in Figure~\ref{fig:gnn_architecture}. All node and edge features are first projected to a common hidden dimension. The model then applies several layers of heterogeneous GINE-style message passing over the station-to-station, train-to-past-station, and train-to-future-station relations, together with their reverse relations. At each layer, each edge embedding is updated from the states of its incident nodes together with its previous edge state, node states are updated through heterogeneous message aggregation, and residual connections with optional layer normalization are applied. Finally, each future train-to-station edge defines a prediction target. For each such edge, the corresponding train representation, station representation, and future-edge embedding are concatenated and passed through an MLP head to predict the normalized residual future delay. This design allows the model to combine network topology, local infrastructure context, and train-specific event history within a unified graph representation.

For illustration, Figure~\ref{fig:gnn_snapshot_country} shows a full-network GNN snapshot. To keep the visualization readable, we display at most the 7 closest past and 7 closest future train-to-station edges for each train, and we select a snapshot containing only 53 active trains to avoid overloading the figure.

\subsection{Training Procedure}
\label{app:training_procedure}

All learning-based models were trained on the training split, with hyperparameter selection based on validation MAE computed on the last 10\% of training snapshots in temporal order. After hyperparameter selection, final benchmark results were obtained by retraining each model on the full training split and evaluating on the common 2025 test split through the shared \texttt{test\_eval\_table}. For stochastic learning-based models, final test results were aggregated over seeds $\{0,1,2,3,4,5,6,7,8,9\}$.

Table~\ref{tab:training_procedure} summarizes the main model-specific training choices.

\begin{table}[H]
\centering
\begin{tabular}{lll}
\toprule
Aspect & XGBoost & Neural models \\
\midrule
Objective & \texttt{reg:absoluteerror} & L1 loss \\
Optimizer & native XGBoost boosting & AdamW \\
Max trial length & determined by \texttt{n\_estimators} & 75 epochs \\
Early stopping & 25 rounds & 6 epochs \\
\bottomrule
\end{tabular}
\caption{Training settings used during hyperparameter-search trials for learning-based models.}
\label{tab:training_procedure}
\end{table}

All XGBoost and MLP runs use fp32 precision, while LSTM, Transformer, and GNN use fp32 on the lite tier and bf16 on the standard tier.

\subsection{Hyperparameter Search and Best Configurations}
\label{app:hparamandconfigs}

Hyperparameter search was performed independently for each model family and benchmark tier using Optuna \cite{akiba2019optuna}, optimizing validation MAE on the last 10\% of training snapshots in temporal order. Search spaces were chosen to reflect each model family's architectural requirements while remaining as comparable as possible across tiers; shared settings are merged across tiers when identical. The best configuration per model family and tier retained for final retraining and test evaluation is reported alongside each search space for reproducibility.

\subsubsection{XGBoost}

\begin{table}[H]
\centering
\begin{tabular}{llcc}
\toprule
Hyperparameter & Type & Standard & Lite \\
\midrule
\texttt{n\_estimators} & int & \multicolumn{2}{c}{$\{200, 400, 800, 1200\}$} \\
\texttt{max\_depth} & int & \multicolumn{2}{c}{$\{6, 8, 10, 12\}$} \\
\texttt{learning\_rate} & float & \multicolumn{2}{c}{log-uniform $[5\times10^{-3}, 10^{-1}]$} \\
\texttt{subsample} & float & \multicolumn{2}{c}{$\{0.6, 0.7, 0.8, 0.9, 1.0\}$} \\
\texttt{colsample\_bytree} & float & \multicolumn{2}{c}{$\{0.5, 0.6, 0.7, 0.8, 0.9, 1.0\}$} \\
\texttt{min\_child\_weight} & int & \multicolumn{2}{c}{$\{1, 5, 10, 20\}$} \\
\texttt{reg\_alpha} & float & \multicolumn{2}{c}{log-uniform $[10^{-3}, 10^{1}]$} \\
\texttt{reg\_lambda} & float & \multicolumn{2}{c}{log-uniform $[10^{-3}, 10^{1}]$} \\
\bottomrule
\end{tabular}
\caption{XGBoost hyperparameter search space for the lite and standard benchmark tiers.}
\label{tab:xgboost_search_space}
\end{table}

\begin{table}[H]
\centering
\begin{tabular}{lcc}
\toprule
Hyperparameter & Standard & Lite \\
\midrule
\texttt{n\_estimators} & 1200 & 1200 \\
\texttt{max\_depth} & 12 & 12 \\
\texttt{learning\_rate} & 8.13e-2 & 6.62e-2 \\
\texttt{subsample} & 1.0 & 1.0 \\
\texttt{colsample\_bytree} & 0.6 & 0.7 \\
\texttt{min\_child\_weight} & 1 & 20 \\
\texttt{reg\_alpha} & 4.09e-2 & 9.70e-2 \\
\texttt{reg\_lambda} & 9.05e-1 & 1.31e-1 \\
\bottomrule
\end{tabular}
\caption{Best XGBoost configurations selected for the lite and standard benchmark tiers.}
\label{tab:xgboost_best_configs}
\end{table}

\subsubsection{MLP}

\begin{table}[H]
\centering
\begin{tabular}{llcc}
\toprule
Hyperparameter & Type & Standard & Lite \\
\midrule
\texttt{hidden\_dims} & list[int] & \multicolumn{2}{c}{3/5/7/9-layer pyramidal templates, widths from 256 up to 4096\footnotemark} \\
\texttt{dropout} & float & \multicolumn{2}{c}{$\{0.05, 0.10, 0.15, 0.20, 0.25\}$} \\
\texttt{batch\_size} & int & \multicolumn{2}{c}{$\{512, 1024, 2048, 4096\}$} \\
\texttt{lr} & float & \multicolumn{2}{c}{log-uniform $[10^{-4}, 3\times10^{-3}]$} \\
\texttt{weight\_decay} & float & \multicolumn{2}{c}{log-uniform $[10^{-6}, 10^{-3}]$} \\
\bottomrule
\end{tabular}
\caption{MLP hyperparameter search space for the lite and standard benchmark tiers.}
\label{tab:mlp_search_space}
\end{table}
\footnotetext{The exact templates are [256, 512, 256], [256, 512, 1024, 512, 256], [256, 512, 1024, 2048, 1024, 512, 256], and [256, 512, 1024, 2048, 4096, 2048, 1024, 512, 256].}

\begin{table}[H]
\centering
\begin{tabular}{lcc}
\toprule
Hyperparameter & Standard & Lite \\
\midrule
\texttt{hidden\_dims} & pyramidal, peak 2048 & pyramidal, peak 4096 \\
\texttt{dropout} & 0.10 & 0.15 \\
\texttt{epochs} & 64 & 72 \\
\texttt{batch\_size} & 4096 & 1024 \\
\texttt{lr} & 1.31e-4 & 1.09e-4 \\
\texttt{weight\_decay} & 1.68e-6 & 1.09e-5 \\
\bottomrule
\end{tabular}
\caption{Best MLP configurations selected for the lite and standard benchmark tiers.}
\label{tab:mlp_best_configs}
\end{table}

\subsubsection{LSTM}

\begin{table}[H]
\centering
\begin{tabular}{llcc}
\toprule
Hyperparameter & Type & Standard & Lite \\
\midrule
\texttt{batch\_size} & int & \multicolumn{2}{c}{$\{512, 1024, 2048\}$} \\
\texttt{lr} & float & \multicolumn{2}{c}{log-uniform $[5\times10^{-5}, 2\times10^{-3}]$} \\
\texttt{weight\_decay} & float & \multicolumn{2}{c}{log-uniform $[10^{-6}, 10^{-4}]$} \\
\texttt{hidden\_dim} & int & \multicolumn{2}{c}{$\{128, 256, 512\}$} \\
\texttt{num\_layers} & int & \multicolumn{2}{c}{$\{1, 2, 3\}$} \\
\texttt{dropout} & float & \multicolumn{2}{c}{$\{0.05, 0.10, 0.15, 0.20, 0.25\}$} \\
\texttt{static\_hidden\_dim} & int & \multicolumn{2}{c}{set to \texttt{hidden\_dim}} \\
\texttt{static\_out\_dim} & int & \multicolumn{2}{c}{set to $\lfloor\texttt{hidden\_dim}/2\rfloor$} \\
\texttt{head\_hidden\_dim} & int & \multicolumn{2}{c}{set to \texttt{hidden\_dim}} \\
\bottomrule
\end{tabular}
\caption{LSTM hyperparameter search space for the lite and standard benchmark tiers.}
\label{tab:lstm_search_space}
\end{table}

\begin{table}[H]
\centering
\begin{tabular}{lcc}
\toprule
Hyperparameter & Standard & Lite \\
\midrule
\texttt{batch\_size} & 2048 & 512 \\
\texttt{lr} & 1.99e-3 & 1.32e-3 \\
\texttt{weight\_decay} & 3.41e-5 & 1.72e-5 \\
\texttt{hidden\_dim} & 512 & 512 \\
\texttt{num\_layers} & 3 & 3 \\
\texttt{dropout} & 0.20 & 0.15 \\
\texttt{static\_hidden\_dim} & 512 & 512 \\
\texttt{static\_out\_dim} & 256 & 256 \\
\texttt{head\_hidden\_dim} & 512 & 512 \\
\texttt{epochs} & 13 & 12 \\
\bottomrule
\end{tabular}
\caption{Best LSTM configurations selected for the lite and standard benchmark tiers.}
\label{tab:lstm_best_configs}
\end{table}

\subsubsection{Transformer}

\begin{table}[H]
\centering
\begin{tabular}{llcc}
\toprule
Hyperparameter & Type & Standard & Lite \\
\midrule
\texttt{d\_model} & int & \{128, 256, 512, 1024\} & \{128, 256, 512\} \\
\texttt{nhead} & int & \multicolumn{2}{c}{$\{1, 2, 4, 8\}$} \\
\texttt{num\_layers} & int & \multicolumn{2}{c}{$\{2, 3, 4, 5, 6\}$} \\
\texttt{dim\_feedforward} & int & \multicolumn{2}{c}{set to $4\times\texttt{d\_model}$} \\
\texttt{dropout} & float & \multicolumn{2}{c}{$\{0.05, 0.10, 0.15, 0.20, 0.25, 0.30, 0.35\}$} \\
\texttt{batch\_size} & int & \{32, 64, 128, 256\} & \{16, 32, 64\} \\
\texttt{lr} & float & \multicolumn{2}{c}{log-uniform $[5\times10^{-5}, 5\times10^{-3}]$} \\
\texttt{weight\_decay} & float & \multicolumn{2}{c}{log-uniform $[10^{-6}, 10^{-4}]$} \\
\bottomrule
\end{tabular}
\caption{Transformer hyperparameter search space for the lite and standard benchmark tiers.}
\label{tab:transformer_search_space}
\end{table}

\begin{table}[H]
\centering
\begin{tabular}{lcc}
\toprule
Hyperparameter & Standard & Lite \\
\midrule
\texttt{d\_model} & 512 & 256 \\
\texttt{nhead} & 1 & 1 \\
\texttt{num\_layers} & 6 & 6 \\
\texttt{dim\_feedforward} & 2048 & 1024 \\
\texttt{dropout} & 0.35 & 0.30 \\
\texttt{batch\_size} & 128 & 32 \\
\texttt{lr} & 6.58e-5 & 1.27e-4 \\
\texttt{weight\_decay} & 2.44e-5 & 1.85e-6 \\
\texttt{epochs} & 60 & 58 \\
\bottomrule
\end{tabular}
\caption{Best Transformer configurations selected for the lite and standard benchmark tiers.}
\label{tab:transformer_best_configs}
\end{table}

\subsubsection{GNN}

\begin{table}[H]
\centering
\begin{tabular}{llcc}
\toprule
Hyperparameter & Type & Standard & Lite \\
\midrule
\texttt{batch\_size} & int & \multicolumn{2}{c}{$\{16, 32, 64\}$} \\
\texttt{lr} & float & \multicolumn{2}{c}{log-uniform $[5\times10^{-5}, 2\times10^{-3}]$} \\
\texttt{weight\_decay} & float & \multicolumn{2}{c}{log-uniform $[10^{-6}, 10^{-4}]$} \\
\texttt{hidden\_dim} & int & \multicolumn{2}{c}{$\{64, 128, 256\}$} \\
\texttt{num\_layers} & int & \multicolumn{2}{c}{$\{2, 3, 4, 5, 6\}$} \\
\texttt{hetero\_aggr} & string & \multicolumn{2}{c}{$\{\text{sum}, \text{mean}\}$} \\
\texttt{use\_layer\_norm} & bool & \multicolumn{2}{c}{$\{\text{true}, \text{false}\}$} \\
\texttt{gnn\_dropout} & float & \multicolumn{2}{c}{$\{0.00, 0.01, 0.05, 0.10, 0.15\}$} \\
\texttt{head\_dropout} & float & \multicolumn{2}{c}{$\{0.00, 0.01, 0.05, 0.10, 0.15\}$} \\
\texttt{edge\_head\_hidden\_dim} & int & \multicolumn{2}{c}{$\{128, 256, 512\}$} \\
\bottomrule
\end{tabular}
\caption{GNN hyperparameter search space for the lite and standard benchmark tiers.}
\label{tab:gnn_search_space}
\end{table}

\begin{table}[H]
\centering
\begin{tabular}{lcc}
\toprule
Hyperparameter & Standard & Lite \\
\midrule
\texttt{batch\_size} & 16 & 32 \\
\texttt{lr} & 1.17e-4 & 7.90e-4 \\
\texttt{weight\_decay} & 9.35e-6 & 6.86e-5 \\
\texttt{hidden\_dim} & 128 & 128 \\
\texttt{num\_layers} & 5 & 6 \\
\texttt{hetero\_aggr} & sum & mean \\
\texttt{use\_layer\_norm} & false & false \\
\texttt{gnn\_dropout} & 0.15 & 0.15 \\
\texttt{head\_dropout} & 0.00 & 0.10 \\
\texttt{edge\_head\_hidden\_dim} & 512 & 256 \\
\texttt{epochs} & 49 & 34 \\
\bottomrule
\end{tabular}
\caption{Best GNN configurations selected for the lite and standard benchmark tiers.}
\label{tab:gnn_best_configs}
\end{table}

\subsection{Computational Resources and Search Budgets}
\label{app:resourcesandbudget}

The experimental setup tables report per-worker resources and budgets; each Optuna job launches 8 workers in parallel, so the aggregate worker-hour budget is 8 times the reported per-worker wall-clock budget.

\begin{table}[H]
\centering
\begin{tabular}{ll}
\toprule
Setting & Value \\
\midrule
GPU per worker & NVIDIA V100-32GB \\
CPU resources per worker & 10 $\times$ Intel(R) Xeon(R) Gold 6248 @ 2.50GHz \\
RAM & 40\,GB \\
Optuna workers per job & 8 \\
Max trials per worker & 50 \\
Optuna budget per worker & 10h for MLP and XGBoost; 19h for other models \\
\bottomrule
\end{tabular}
\caption{Experimental setup for the lite tier.}
\label{tab:experimental_setup_lite}
\end{table}

\begin{table}[H]
\centering
\begin{tabular}{ll}
\toprule
Setting & Value \\
\midrule
GPU per worker & NVIDIA A100 \\
CPU resources per worker & 8 $\times$ EPYC 7543 Milan AMD processors \\
RAM & 64\,GB \\
Optuna workers per job & 8 \\
Max trials per worker & 50 \\
Optuna budget per worker & 10h for MLP and XGBoost; 19h for other models \\
\bottomrule
\end{tabular}
\caption{Experimental setup for the standard tier.}
\label{tab:experimental_setup_standard}
\end{table}

\subsection{Additional Experiments}

\subsubsection{Additional Standard-Tier Result Figures}
\label{app:standard-tier-additional-figures}

\begin{figure}[H]
    \centering
    \includegraphics[width=\linewidth]{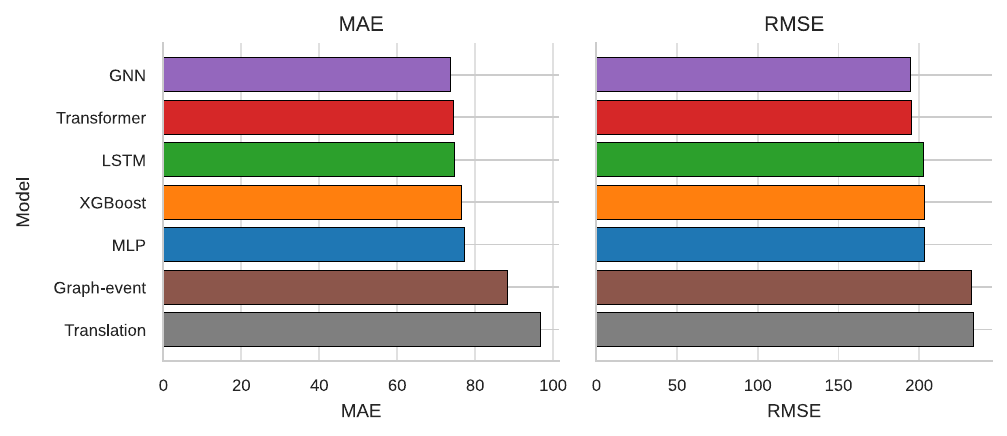}
    \caption{Aggregate standard-tier benchmark performance across models.}
    \label{fig:standard-main-results-bar}
\end{figure}

\begin{figure}[H]
    \centering
    \includegraphics[width=\linewidth]{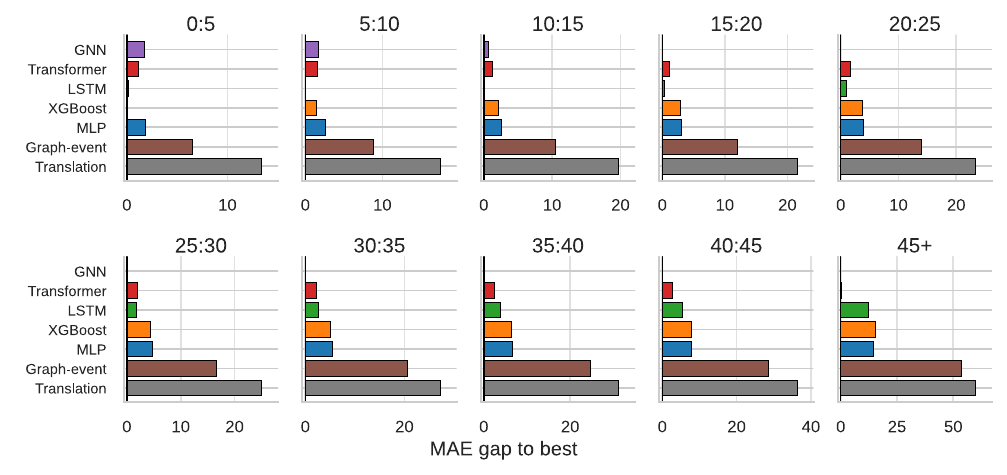}
    \caption{Standard-tier prediction-horizon breakdown, shown as MAE gaps relative to the best model in each horizon bin.}
    \label{fig:standard-horizon-mae-plot}
\end{figure}

\begin{figure}[H]
    \centering
    \includegraphics[width=\linewidth]{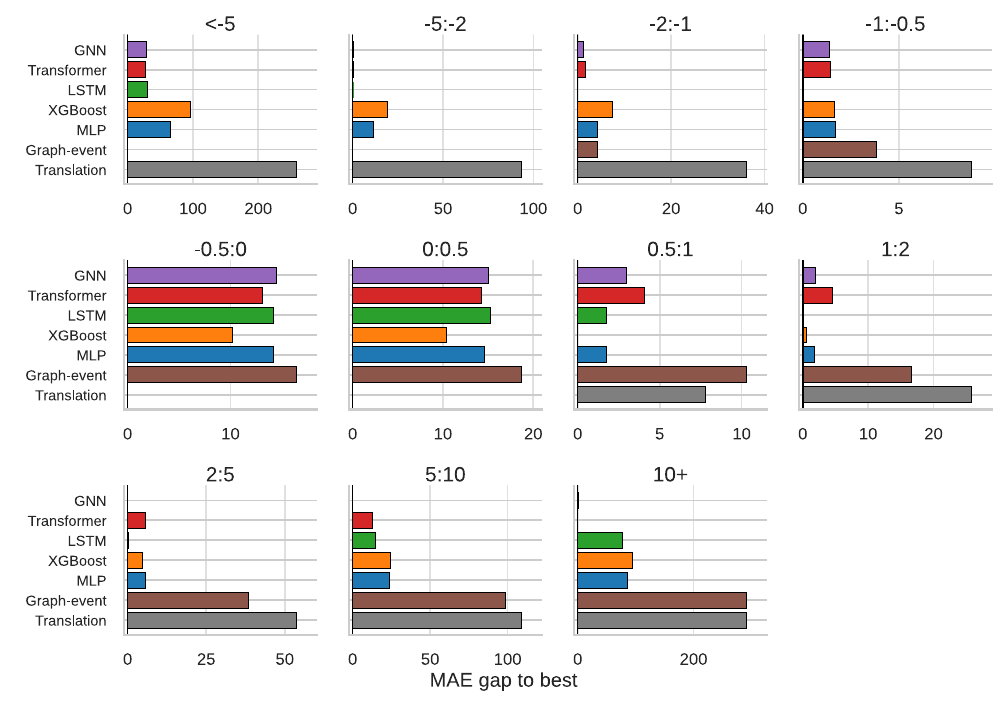}
    \caption{Standard-tier delay-delta breakdown, shown as MAE gaps relative to the best model in each delay-delta bin.}
    \label{fig:standard-delay-delta-mae-plot}
\end{figure}

This subsection provides complementary visualizations of the standard-tier benchmark results. Figure~\ref{fig:standard-main-results-bar} summarizes aggregate performance, while Figures~\ref{fig:standard-horizon-mae-plot} and~\ref{fig:standard-delay-delta-mae-plot} show regime-wise MAE gaps across prediction horizon and delay-delta bins.

\subsubsection{Benchmark on the Lite Dataset}

\begin{figure}[H]
    \centering
    \includegraphics[width=\linewidth]{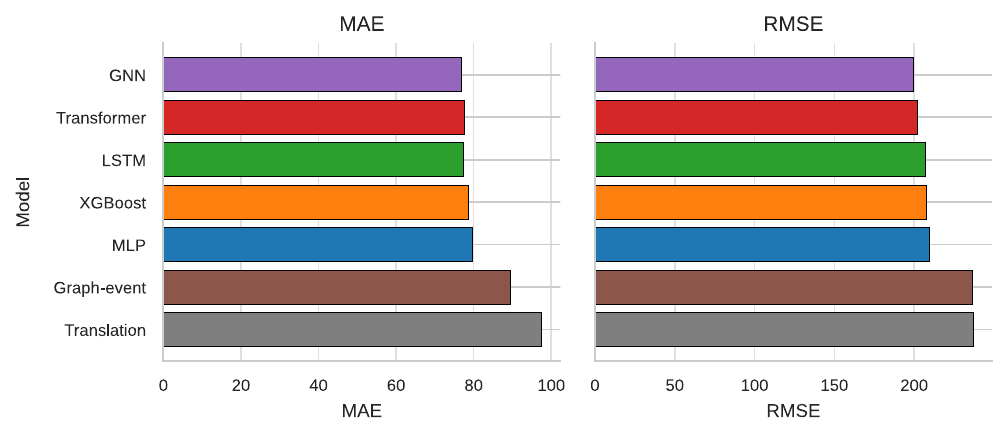}
    \caption{Aggregate lite-tier benchmark performance across models.}
    \label{fig:lite-main-results-bar}
\end{figure}

\paragraph{Aggregate Results.} Table~\ref{tab:lite_main_results} summarizes results on the lite benchmark tier, which is intentionally smaller and uses 15{,}000 training snapshots and 3{,}000 test snapshots. The GNN again attains the best mean performance (76.78 MAE, 199.84 RMSE), followed closely by the LSTM (77.26 MAE) and Transformer (77.72 MAE). Among tabular models, XGBoost outperforms the MLP (78.59 vs.\ 79.73 MAE). The graph-event model improves clearly over the simple translation rule (89.45 vs.\ 97.48 MAE), yet remains well behind all learning-based approaches, so the gap between learning and non-learning methods persists in this lower-data regime. Compared to the standard tier, the learning-based models are more gradually separated, and one shift in the relative ordering is worth noting. The Transformer drops below the LSTM, consistent with the expectation that attention-based architectures are less sample-efficient. The GNN's advantage over the second-best model has also narrowed to a smaller margin (76.78 vs.\ 77.26 MAE), while its variance across runs is higher than any other learning-based model. That the LSTM remains this competitive despite operating on one train at a time and lacking any explicit network-level delay propagation beyond local network context features is a notable result. As on the standard tier, no single architecture dominates convincingly, suggesting that further progress is as likely to come from stronger feature design and problem-specific modeling choices as from architectural changes alone. Figure~\ref{fig:lite-main-results-bar} provides a compact visual summary of these aggregate lite-tier comparisons.

\begin{table}[H]
\centering
\begin{tabular}{lll}
\toprule
Model & MAE & RMSE \\
\midrule
Translation & 97.48 & 236.87 \\
Graph-event & 89.45 & 236.30 \\
MLP & 79.73 $\pm$ 0.10 & 209.44 $\pm$ 1.06 \\
XGBoost & 78.59 $\pm$ 0.02 & 207.97 $\pm$ 0.05 \\
LSTM & 77.26 $\pm$ 0.26 & 207.07 $\pm$ 0.68 \\
Transformer & 77.72 $\pm$ 0.18 & 201.81 $\pm$ 0.74 \\
GNN & \textbf{76.78 $\pm$ 0.68} & \textbf{199.84 $\pm$ 1.06} \\
\bottomrule
\end{tabular}
\caption{Lite-tier test performance. MAE/RMSE in seconds; $\pm$: std. over 10 seeds.}
\label{tab:lite_main_results}
\end{table}

\begin{figure}[H]
    \centering
    \includegraphics[width=\linewidth]{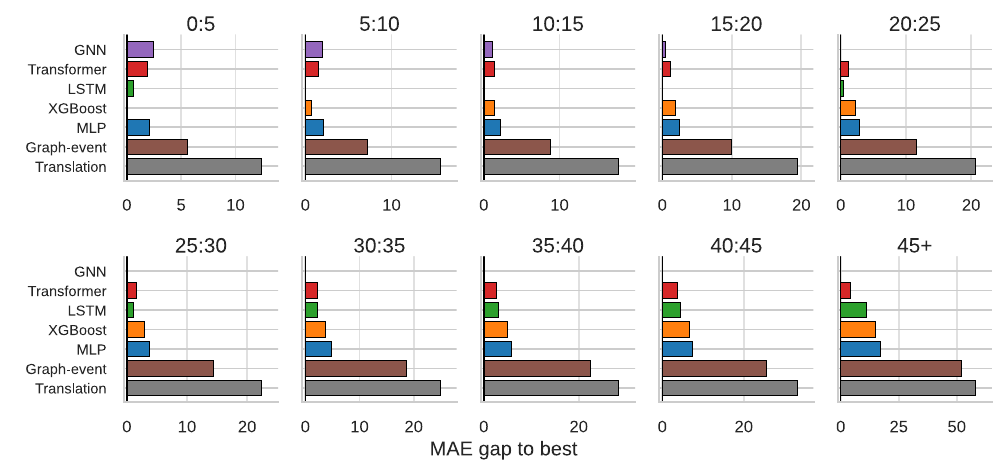}
    \caption{Lite-tier prediction-horizon breakdown, shown as MAE gaps relative to the best model in each horizon bin.}
    \label{fig:lite-horizon-mae-plot}
\end{figure}

\paragraph{Results by Prediction Horizon.} Table~\ref{tab:lite_horizon_mae} breaks down performance by prediction horizon. A broadly similar pattern emerges on the lite tier, although with a few notable differences. In the very shortest 0:5 minute bin, XGBoost performs best, consistent with its strong performance in the dominant regime of small delay changes, as explained in the delay change paragraph below. Beyond that, the LSTM is the strongest model from 5 to 20 minutes, so its short-horizon advantage extends further than on the standard tier, where it is strongest from 5 to 15 minutes after XGBoost takes the very shortest 0:5 minute bin. The GNN then becomes the best-performing approach from 20 minutes onward. The Transformer again remains competitive across all horizons and, as on the standard tier, tends to outperform the LSTM at longer horizons, but here this separation emerges more gradually and only becomes clear later in the horizon range. As on the standard tier, one important caveat is that the latest horizon bins are progressively biased toward trains that accumulate delay, with this effect being particularly pronounced in the final 45+ bin, since these bins become thinner by construction under the fixed future-event prediction window. As a result, performance differences across horizon bins should not be interpreted as a pure horizon effect alone, but rather as reflecting a mixture of horizon length and delay regime. Figure~\ref{fig:lite-horizon-mae-plot} visualizes the corresponding per-bin MAE gaps relative to the best model in each horizon regime.

\begin{table}[H]
\centering
\setlength{\tabcolsep}{3pt}
\begin{tabular}{lllllllllll}
\toprule
Model & 0:5 & 5:10 & 10:15 & 15:20 & 20:25 & 25:30 & 30:35 & 35:40 & 40:45 & 45+ \\
\midrule
Translation & 37.4 & 57.2 & 72.3 & 85.0 & 95.6 & 107.5 & 122.4 & 142.0 & 166.2 & 324.5 \\
Graph-event & 30.6 & 48.7 & 63.2 & 75.6 & 86.6 & 99.4 & 116.1 & 136.0 & 158.6 & 318.1 \\
MLP & 27.1 & 43.6 & 56.6 & 68.0 & 77.7 & 88.8 & 102.3 & 119.4 & 140.6 & 283.5 \\
XGBoost & \textbf{25.0} & 42.2 & 55.8 & 67.4 & 77.2 & 88.0 & 101.3 & 118.5 & 139.6 & 281.1 \\
LSTM & 25.6 & \textbf{41.5} & \textbf{54.4} & \textbf{65.6} & 75.3 & 86.2 & 99.7 & 116.7 & 137.6 & 277.5 \\
Transformer & 26.9 & 43.0 & 55.8 & 66.7 & 76.1 & 86.7 & 99.8 & 116.3 & 136.8 & 270.4 \\
GNN & 27.5 & 43.5 & 55.6 & 66.0 & \textbf{74.9} & \textbf{85.0} & \textbf{97.5} & \textbf{113.5} & \textbf{133.1} & \textbf{266.3} \\
\bottomrule
\end{tabular}
\caption{Mean test-set MAE by prediction horizon on the lite tier. Bins are expressed in minutes.}
\label{tab:lite_horizon_mae}
\end{table}

\begin{figure}[H]
    \centering
    \includegraphics[width=\linewidth]{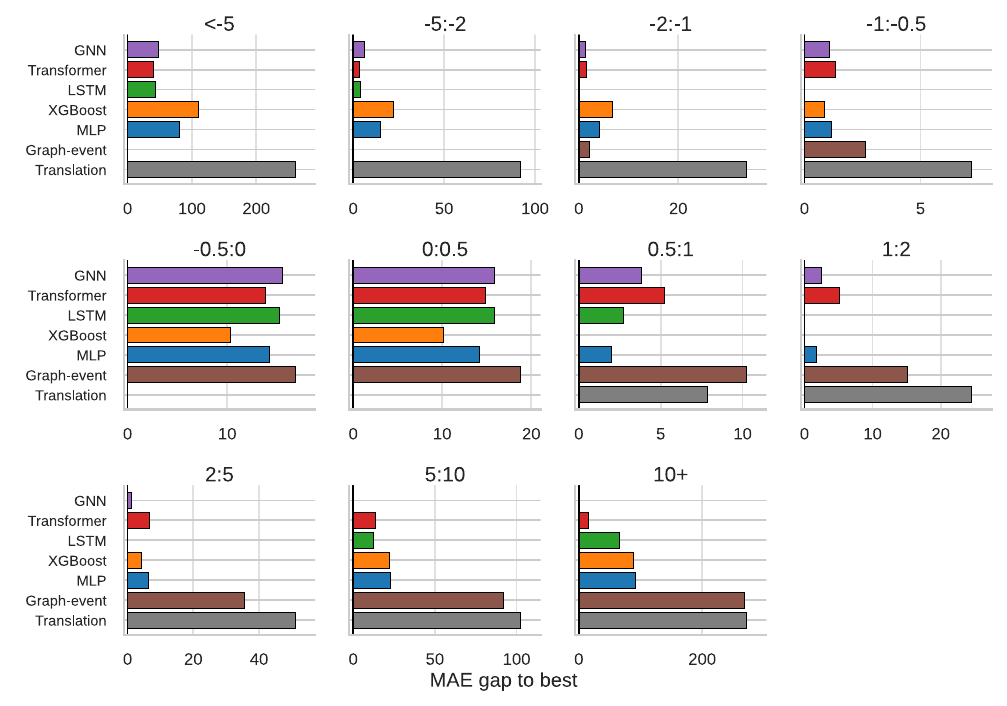}
    \caption{Lite-tier delay-delta breakdown, shown as MAE gaps relative to the best model in each delay-delta bin.}
    \label{fig:lite-delay-delta-mae-plot}
\end{figure}

\paragraph{Results by Delay Change.} Table~\ref{tab:lite_delay_delta_mae} reports performance by delay-delta bin. As on the standard tier, the translation baseline is strongest around zero delay change, where simply carrying forward the current delay provides a strong approximation. Excluding this trivial baseline, XGBoost is notably strongest in the central delay-delta regime around zero, roughly from $-0.5$ to $1$ minute, which corresponds to the most typical delay-change regime in the data. This suggests, again, that XGBoost has a stronger tendency than the other learning-based models to stay close to the dominant regime of small delay changes. The LSTM again performs particularly well on moderate negative delay-delta bins, indicating that sequential structure alone already provides a strong signal for modeling local delay recovery. For moderate delay accumulation, between 0.5 and 5 minutes, the strongest models are again tabular or sequential rather than graph-based, with the LSTM now performing best in the 2 to 5 minute range. At the opposite end, the graph-event model performs best on large negative delay deltas, i.e.\ strong delay recovery, which is consistent with its tendency to predict travel times that are on average faster than the scheduled ones. In contrast, the GNN performs best across the large positive delay-delta bins and is now also clearly strongest in the most extreme 5+ minute accumulation regime. As on the standard tier, this suggests that models with learned interaction patterns between trains are better able to capture substantial delay increases, likely because these are tied to propagation effects in the network. Figure~\ref{fig:lite-delay-delta-mae-plot} visualizes these regime-wise differences as MAE gaps relative to the best model in each delay-delta bin.

\begin{table}[H]
\centering
\setlength{\tabcolsep}{3pt}
\begin{tabular}{llllllllllll}
\toprule
Model & <-5 & -5:-2 & -2:-1 & -1:-0.5 & -0.5:0 & 0:0.5 & 0.5:1 & 1:2 & 2:5 & 5:10 & 10+ \\
\midrule
Translation & 431.4 & 174.4 & 85.2 & 44.3 & \textbf{14.7} & \textbf{13.5} & 43.3 & 85.0 & 184.8 & 409.1 & 1262.7 \\
Graph-event & \textbf{171.2} & \textbf{82.3} & 53.7 & 39.7 & 31.5 & 32.3 & 45.6 & 75.6 & 169.3 & 398.5 & 1259.1 \\
MLP & 251.0 & 97.3 & 55.6 & 38.2 & 28.9 & 27.7 & 37.4 & 62.2 & 140.0 & 329.2 & 1083.1 \\
XGBoost & 281.5 & 104.3 & 58.2 & 37.9 & 25.0 & 23.6 & \textbf{35.4} & 60.5 & 138.0 & 328.7 & 1079.0 \\
LSTM & 214.0 & 86.3 & \textbf{51.6} & \textbf{37.0} & 30.0 & 29.4 & 38.1 & \textbf{60.5} & \textbf{133.8} & 318.8 & 1056.9 \\
Transformer & 212.0 & 85.5 & 53.0 & 38.4 & 28.6 & 28.4 & 40.6 & 65.6 & 140.4 & 320.4 & 1006.5 \\
GNN & 218.5 & 88.4 & 52.9 & 38.1 & 30.3 & 29.4 & 39.2 & 62.9 & 134.9 & \textbf{306.6} & \textbf{991.3} \\
\bottomrule
\end{tabular}
\caption{Mean test-set MAE by delay-delta bin on the lite tier. Bins are expressed in minutes; negative values correspond to delay recovery and positive values to delay accumulation.}
\label{tab:lite_delay_delta_mae}
\end{table}

Together, these two evaluation breakdowns again provide a more informative view of model behavior than aggregate MAE and RMSE alone. They make it possible to identify which models perform best in specific forecasting regimes, such as short- versus long-horizon prediction or delay recovery versus delay accumulation, and thereby support broader insights into the dynamics of train delay prediction, even in the smaller-data regime of the lite tier.

\subsubsection{Ablation Study on Feature Families}

To better understand which information sources drive performance in the tabular setting, we perform a feature-family ablation study on the lite-tier MLP model. Starting from the best-performing full-feature lite MLP configuration, we remove one feature family at a time, retrain the model, and compare the resulting test-set performance against the full-feature model. Hyperparameters are kept fixed to the selected full-feature configuration, while the number of training epochs for each ablation is chosen in a preliminary validation phase using the last 10\% of training snapshots in temporal order. Final results are then obtained on the lite test split using the selected epoch count and aggregating over three fixed seeds, $\{0,1,2\}$.

\begin{figure}[H]
    \centering
    \includegraphics[width=\linewidth]{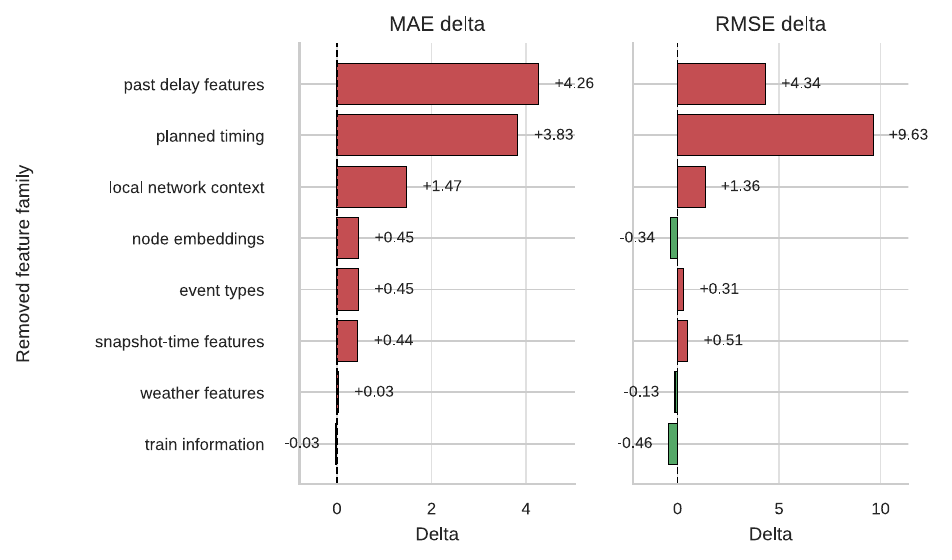}
    \caption{Lite-tier MLP feature-family ablation results, shown as MAE and RMSE changes relative to the full-feature model.}
    \label{fig:lite-mlp-ablation-deltas}
\end{figure}

Table~\ref{tab:mlp_feature_ablation} shows that the most important feature families for the lite-tier MLP are past delay features and planned timing features. Removing past delay features produces the largest MAE degradation (+4.26), while removing planned timing features is nearly as harmful (+3.83) and leads to the largest RMSE increase (+9.63). Unsurprisingly, this indicates that the MLP depends strongly on past itinerary delay information and schedule-relative information. Interestingly, the larger RMSE increase when removing planned timing features suggests that schedule-relative information may be more important for limiting larger prediction errors tied to more extreme cases. Local network context also contributes meaningfully, though to a lesser extent (+1.47 MAE). This is notable in light of the broader benchmark results: the overall gap between the GNN and the MLP on the lite tier is only about 3 MAE, so a 1.47 MAE degradation from removing local network context highlights just how informative these engineered features already are for summarizing nearby structural effects. Snapshot-time features have a smaller but still visible effect (+0.44 MAE), suggesting that coarse temporal context beyond the immediate event sequence remains useful. Event-type indicators and node embeddings also have a modest but measurable contribution (+0.45 MAE in both cases). For the MLP, node embeddings can plausibly act mainly as a way to memorize operational-point-specific patterns associated with particular parts of the network, whereas more expressive architectures may be better able to exploit such representations for modeling interactions between trains and events, for instance through attention mechanisms. In contrast, removing train information or weather features has almost no effect, with changes close to zero. For weather, this weak effect may reflect redundancy with other inputs: temporal features already capture strong seasonal and time-of-day patterns, including some of the variation associated with temperature, while delay history may already absorb part of the operational impact of weather conditions. Interestingly, RMSE actually improves slightly when removing train information ($-0.46$), node embeddings ($-0.34$), or weather features ($-0.13$). Figure~\ref{fig:lite-mlp-ablation-deltas} visualizes these ablation effects relative to the full-feature MLP.

A general caveat in feature ablation studies is that removing a feature family may change not only the information available to the model, but also the effective input dimensionality, model capacity, and regularization regime when the rest of the hyperparameters are kept fixed. In our setup, we reselect only the number of training epochs for each ablation, while keeping all other hyperparameters fixed to the full-feature configuration. This issue is therefore most relevant for node embeddings and, to a lesser extent, event-type indicators, since these are among the highest-dimensional components of the tabular representation. As a result, their ablations should not be interpreted purely as information removal.

Overall, the ablation results indicate that, for the MLP, predictive performance depends most strongly on past itinerary delay information, schedule-relative timing, and local network context, while the remaining feature families appear either secondary or partly redundant in the lite setting.

\begin{table}[H]
\centering
\begin{tabular}{lcccc}
\toprule
Setting & MAE & $\Delta$MAE & RMSE & $\Delta$RMSE \\
\midrule
MLP (all features) & 79.73 $\pm$ 0.10 & --- & 209.44 $\pm$ 1.06 & --- \\
\midrule
w/o train information & 79.70 $\pm$ 0.04 & -0.03 & 208.98 $\pm$ 0.37 & -0.46 \\
w/o snapshot-time features & 80.17 $\pm$ 0.12 & +0.44 & 209.95 $\pm$ 0.67 & +0.51 \\
w/o planned timing & 83.56 $\pm$ 0.21 & +3.83 & 219.07 $\pm$ 0.96 & +9.63 \\
w/o past delay features & 83.99 $\pm$ 0.04 & +4.26 & 213.78 $\pm$ 0.30 & +4.34 \\
w/o event types & 80.18 $\pm$ 0.10 & +0.45 & 209.75 $\pm$ 0.60 & +0.31 \\
w/o node embeddings & 80.18 $\pm$ 0.07 & +0.45 & 209.10 $\pm$ 1.06 & -0.34 \\
w/o local network context & 81.20 $\pm$ 0.07 & +1.47 & 210.80 $\pm$ 0.36 & +1.36 \\
w/o weather features & 79.76 $\pm$ 0.05 & +0.03 & 209.32 $\pm$ 0.25 & -0.13 \\
\bottomrule
\end{tabular}
\caption{Main results of the lite-tier MLP feature-family ablation study. The first row shows the full-feature model, and each subsequent row removes one feature family. MAE and RMSE are reported as mean $\pm$ standard deviation across runs, with $\Delta$ values measured relative to the base model.}
\label{tab:mlp_feature_ablation}
\end{table}

\newpage

\end{document}